\documentclass[10pt,journal]{IEEEtran}

\usepackage{amsfonts}
\usepackage{amsmath}
\usepackage{amsthm}
\usepackage{graphicx}
\usepackage{subfigure}
\usepackage{multirow}
\usepackage{multicol}
\usepackage{graphicx}
\usepackage{epstopdf}
\usepackage[justification=centering]{caption}
\usepackage{color}

\newtheorem{assumption}{Assumption}
\newtheorem{theorem}{Theorem}
\newtheorem{corollary}{Corollary}
\newtheorem{lemma}{Lemma}

\newtheorem{definition}{Definition}

\begin{document}

\title{ Depth Selection for Deep ReLU Nets in Feature Extraction and Generalization}

\author{Zhi Han, Siquan Yu,
        Shao-Bo Lin, ~and Ding-Xuan Zhou
\IEEEcompsocitemizethanks{
\IEEEcompsocthanksitem Z. Han and Yu are with the State Key Laboratory of Robotics, Shenyang Institute of Automation, Chinese Academy of Sciences, Shenyang, China and Institutes for Robotics and Intelligent Manufacturing, Chinese Academy of Sciences, Shenyang, China.
S. Yu is also with  the School of Information Science and Engineering, Northeastern University, Shenyang, China.
S.B. Lin is with the  Center of Intelligent Decision-Making and Machine Learning, School of Management, Xi'an Jiaotong University, Xi'an, China.
D.X. Zhou is with the School of Data Science and Department of Mathematics, City University of Hong Kong, Hong Kong, China.

Corresponding author: S. B. Lin (sblin1983@gmail.com)}}

%\markboth{IEEE TRANSACTIONS ON Pattern Analysis and Machine Intelligence}%
%{Shell \MakeLowercase{\textit{et al.}}: Bare Demo of IEEEtran.cls for Computer Society Journals}

\IEEEtitleabstractindextext{%
\begin{abstract}
Deep learning  is recognized to be capable of discovering deep features for representation learning and pattern recognition without requiring elegant feature engineering techniques by
taking
advantage of human ingenuity and prior knowledge. Thus it has triggered enormous research activities in machine learning and pattern recognition.  One of the most important  challenge of deep learning is  to figure out   relations between a feature and the depth of deep neural networks (deep nets for short) to reflect the necessity of depth. Our purpose is to quantify this feature-depth correspondence   in feature extraction and generalization. We present the adaptivity of features to depths and vice-verse via showing a depth-parameter trade-off in extracting both single feature and composite features. Based on these results, we prove that implementing the classical empirical risk minimization on deep nets can achieve the optimal generalization performance for numerous learning tasks. Our theoretical results are verified by a series of numerical experiments including  toy simulations and a real application of earthquake seismic intensity prediction.
\end{abstract}


}

\maketitle

\IEEEdisplaynontitleabstractindextext

\IEEEpeerreviewmaketitle

\section{Introduction}\label{Sec.Introduction}

\IEEEPARstart{A} systemic machine learning process   frequently
comes down to    two steps: feature extraction and target-driven
learning. The former   focuses  on designing preprocessing pipelines
and data transformations that result in a tractable representation
of data, while the latter  utilizes learning algorithms related to
specific targets, such as regression, classification and clustering  on the   data representation to finish the learning task.
Studies in the second step abound  in machine learning \cite{Bishop}
and numerous learning schemes such as kernel methods
\cite{Evgeniou2000}, neural networks \cite{Hagan1996} and boosting
\cite{Hastie2001} have been proposed. However,
feature extraction  in the first step is usually labor intensive,
which requires elegant feature engineering techniques by taking
advantages of human ingenuity and prior knowledge.

\begin{figure}[h]
    \centering
    \includegraphics[scale=0.5]{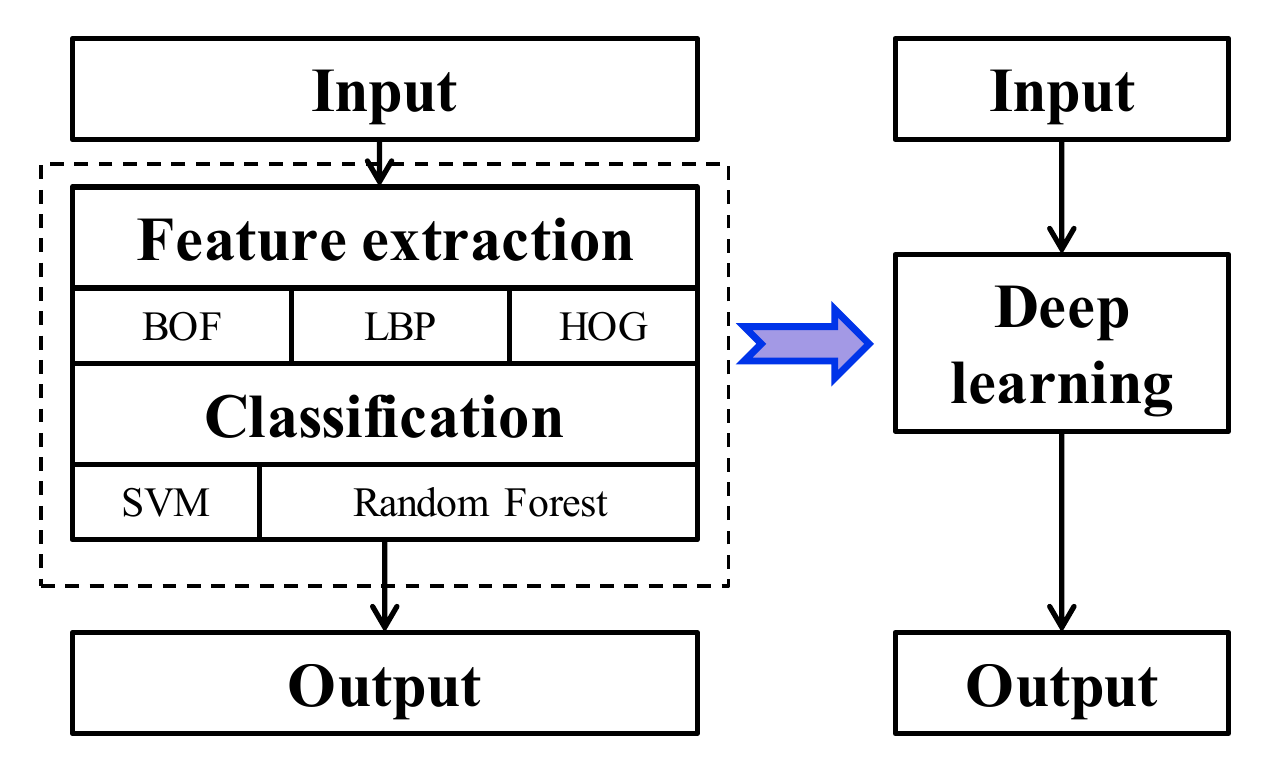}
    \caption{Magic behind deep learning}
    \label{Fig:flow}
\end{figure}

To extend the applicability of machine learning, it is crucial  to
make learning algorithms be less dependent of human factors. Deep
learning \cite{Hinton2006,Goodfellow2016},  which has been
successfully used in image classification, natural language
processing and game theory, provides a promising technique in machine learning. The heart of deep learning is to adopt deep neural networks
(deep nets for short) with certain structures  to extract data features
and design target-driven  algorithms, simultaneously. As shown in Figure \ref{Fig:flow}, deep learning embodies the utilities of feature extraction algorithms such as bag of feature (BOF), local binary pattern (LBP), histogram of oriented gradient (HOG) and classification algorithms like support vector machine (SVM), random forest,    via tuning parameters in a unified deep nets model.
%
%\begin{figure}[!th]
%\centering
%\includegraphics*[scale=0.3]{g142.pdf}
%%  \vspace{-.5cm}
%\centerline{{\small flows for deep learning}}
% \label{Fig:abc}
%\end{figure}

 The great  success of deep
learning in applications {  demonstrates the feasibility of deep nets in specific learning tasks}. However, whether {  deep learning is   generalizable  to other learning tasks relies on rigorous
theoretical verifications}, which is unfortunately at its infancy. In
particular, it is highly desired to clarify the following three important problems: 1) which data
features\footnote{  Data feature in this paper  means  priors for presentation learning according to the terminology in the nice review paper \cite{Bengio2013}. It includes both the a-priori information of target functions and structures of the input space. } can be extracted by deep nets; 2) how to  set the depth of deep nets in special learning tasks;
 3)  How about the generalization ability of  deep learning
algorithms. The first problem {  refers to} the representation performance of deep nets, needing tools from information theory like coding theory \cite{Petersen2017} and entropy theory \cite{Guo2019}. The second one concerns  approximation abilities of deep nets with different depth, requiring approximation theory techniques such as {   local polynomial approximations \cite{Yarotsky2017},}  covering number estimates \cite{Kohler2017} and  wavelets analysis \cite{Zhou2018a} to quantify  powers and limitations of deep nets. The last one focuses on the generalization capability of deep learning algorithms in machine learning, for which statistical learning theory as well as empirical processing \cite{Cucker2007} should be utilized.

Although lagging heavily behind   applications,  recent
developments of deep learning theory provided  some exciting
theoretical results on these problems. For example, \cite{Shaham2015}
proved that deep nets succeed  in extracting some geometric structures
of data, which has been  adopted in \cite{Chui2018} to design deep
learning algorithm for regression problems with data generated on
manifolds;
\cite{Chui1994} found that deep nets can extract  local position
information of data, which was recently employed in \cite{Lin2018}
to construct deep nets in handling  sparsely located data;
\cite{Petersen2017} proved that deep nets can extract   piecewise
features of data, which was  utilized in \cite{Imaizumi2018} to
develop
 learning algorithms to learn  non-smooth functions efficiently. All these interesting
 studies presented  theoretical verifications on the power of deep learning  in the sense
that deep nets succeed in extracting data features and
deep learning significantly improves the
generalization capabilities of  learning schemes in-hand.

The problem is, however,  there are strongly exclusive correspondences between data features and network depth for these theoretical studies in the sense that each data feature  requires a unique network depth and vice-verse.  To be detailed, a {  hierarchal  structure} corresponds to a hierarchal deep net with the same depth \cite{Mhaskar2016a};  smoothness {  a-priori information} \cite{Yarotsky2017} is related to a deep net with accuracy-dependent depth;  a {  translation-invariance  property} requires a convolutional neural network with accuracy-dependent  layers \cite{Bruna2013}; and a {  rotation-invariance property}  is associated with a deep net  with tree structures and four layers \cite{Chui2018a}.
Such exclusive correspondences hinder heavily  the use of deep nets in feature extraction, since   {  data features such as the smoothness information, hierarchal structure, transformation-invariance} are practically difficult to be specified before the learning process. Furthermore, it is questionable to determine the network depth for simultaneously extracting   multiple data features {  like the translation-invariance and rotation-invariance}, which is pretty common   in practice.
Our first purpose is to break through the feature-depth correspondences by means of proving  that deep nets with certain depth   can extract several data features and vice-verse.

We consider extracting both single  data features such as smoothness, rotation-invariance, sparseness and composite data features combining smoothness, rotation-invariance and sparseness to demonstrate the adaptivity of network depth to features and vice-verse. Intuitively, it is difficult for deep ReLU nets to extract smoothness features due to the non-smooth property of the ReLU function $\sigma(t)=\max\{t,0\}$. A natural remedy for this, as shown in \cite{Yarotsky2017}, is to deepen the network with an accuracy-dependent depth to eliminate the negative effect of non-smoothness.  Since under some
specified capacity measurements such as the number of linear regions
\cite{Montufar2013}, Betti numbers \cite{Bianchini2014}, number of
monomials \cite{Delalleau2011} and covering numbers \cite{Guo2019},
the capacity of deep nets increases exponentially with respect to
the depth, large depth usually means large capacity costs for feature extraction. Furthermore, from an optimization   viewpoint,  large depth requires to solve a   highly nonconvex optimization problem {   \cite[Sec. 8.2]{Goodfellow2016}} involving the
ill-conditioning of the Hessian, the existence of many local minima,
saddle points, plateau and even some flat regions, making it difficult to design optimization algorithms for such deep nets with convergence guarantees. Based on these,  we provide a theoretical guidance for  depth selection to extract   data features by showing that
deep nets with various depths,   larger than a specified value, are capable of  extracting   the smoothness and other data  features.  This shows an adaptivity of the depth to data features  in the sense that any data features from a rich family can be extracted by deep nets with various depths. Conversely, we also provide theoretical guarantees on the  adaptivity of the data feature to depths by showing that deep ReLU nets with some specific depth succeed in extracting the smoothness, spareness and  composite features. All these remove the feature-depth correspondences in feature extraction for deep ReLU nets.

From feature extraction to machine learning,
the tug of war between
bias and variance \cite{Cucker2007} indicates that the prominent performance  of deep
nets in feature extraction is insufficient to demonstrate its success. The good
generalization ability is frequently
built upon the balance between the accuracy of feature extraction and  capacity costs to achieve such an accuracy. This exhibits a bias-variance dilemma in selecting the capacity of deep nets. Different from shallow learning such as kernel methods and boosting, recent studies \cite{Harvey2017,Guo2019} presented  a  depth-parameter dilemma in  controlling the capacity of deep nets in the sense that different depth-parameter pairs   may yield the same capacity.  These two dilemmas as well as the optimization difficulty {  \cite[Sec. 8.2]{Goodfellow2016}} pose an urgent issue for deep learning theory on selecting the depth to guarantee the good generalization ability of deep learning algorithms.
Our second purpose is not only to pursue the  optimal generalization error for  learning schemes based on deep nets, but also to demonstrate the depth selection strategy to realize this optimality.

We study the generalization ability of deep nets with different depths {  via empirical risk minimization (ERM)}. Based on the established  adaptivity  of  the depth to data features in feature extraction, we establish almost optimal generalization error bounds for deep nets with numerous depth-parameter pairs. Our results show  that the feature extraction step
is necessary  when the learning task is somewhat sophisticated and
deep nets succeed in extracting deep data features of the data
distribution, which illustrates the necessity of depth in deep learning. However, we also prove that the depth for realizing the optimal learning performance of deep nets is not unique. In fact, with depth larger than some specified value, all deep nets theoretically perform  similarly and can achieve the optimal generalization error bounds. The only difference is that deeper nets involve {  less free parameters}.

In a nutshell, our analysis  implies three interesting
findings in understanding the success of deep learning.
 The first is  the  flexibility  on
automatically selecting  the accuracy in extracting data  features via
tuning the network parameters, which is different from the classical two-step learning scheme presented in Figure \ref{Fig:flow} that usually involves  extremely high
capacity costs to fully extract data features. The second is the
versatility of deep nets with fixed depth in the sense that they can extract various data features. The third one is that if the depth is larger than a specified value, we can always get a deep net estimator with almost optimal theoretical guarantee. The problem is, however, the difficulty in solving ERM on deep nets  increases with respect to the depth  {  \cite[Sec. 8.2]{Goodfellow2016} and \cite{Allen-Zhu2018}}\footnote{  Here, the difficulty means that larger depth requires more free parameters under an over-parameterization setting to guarantee the convergence of SGD to a local optimal of ERM of high quality and larger depth results in more local minima, saddle points and flat regions.}. Thus, it is numerically difficult {  and time-consuming} to get a deep net estimator with large depth and there is practically an   optimal depth to realize the established optimal generalization error bounds, just as our experimental results exhibit.

The rest of the paper is organized as follows. In the next section,
we will introduce deep nets and show some recent developments of deep nets in feature extraction.
Section 3 focuses on the depth selection for deep ReLU nets  in extracting  single data features, while Section 4 devotes to   the depth selection  in extracting composite data features.  In Section 5, we are interested in the
generalization error analysis for implementing ERM   on
deep nets. Section 6 exhibits some numerical results to verify our
theoretical assertions. In the last section, we draw a simple
conclusion and present some further discussions.

\section{Necessity of Depth in  Feature Extraction}\label{Sec.necessity}

Let $d\in\mathbb N$ be the dimension of input space.
Denote
$x=(x^{(1)},\dots,x^{(d)})\in\mathbb I^d:=[-1,1]^d$. Let
$L\in\mathbb N$ and $d_0,d_1,\dots,d_L\in\mathbb N$ with $d_0=d$.
For
$\vec{h}=(h^{(1)},\dots,h^{(d_k)})^T\in\mathbb R^{d_k}$, define
$\vec{\sigma}(\vec{h})=(\sigma(h^{(1)}),\dots,\sigma(h^{(d_k)}))^T$.
Deep ReLU nets with depth $L$ and width $d_j$ in the $j$-th hidden layer can be mathematically
represented as
\begin{equation}\label{Def:DFCN}
     h_{\{d_0,\dots,d_L,\sigma\}}(x)=\vec{a}\cdot
     \vec{h}_L(x),
\end{equation}
where
\begin{equation}\label{Def:layer vector}
    \vec{h}_k(x)=\vec{\sigma}(W_k\cdot
    \vec{h}_{k-1}(x)+\vec{b}_k),\qquad k=1,2,\dots,L,
\end{equation}
$\vec{h}_{0}(x)=x,$ $\vec{a}\in\mathbb R^{d_L}$,
$\vec{b}_k\in\mathbb R^{d_k},$
 and $W_k=(W_k^{i,j})_{i=1,j=1}^{d_{k},d_{k-1}}$
is a $d_{k}\times
 d_{k-1}$ matrix. Denote by $\mathcal H_{\{d_0,\dots,d_L,\sigma\}}$ the set of
all these deep nets. When $L=1$,  the function defined by
(\ref{Def:DFCN}) is the classical shallow net.

{ 
The structure of  deep nets is reflected by    structures of
  weight matrices $W_k$ and threshold vectors $\vec{b_k}$ and
$\vec{a}$, $k=1,2,\dots,L$.
Besides the deep fully connected networks \cite{Yarotsky2017} that counts the number of free parameters  in the $k$-th layer to be $d_kd_{k-1}+d_k$\footnote{  If k=L, the number is $d_Ld_{L-1}+2d_L$ by taking the outer weights into accounts},   we say that there are $n_k$ free parameters in the $k$-th layer,
if the weight matrix $W_k$ and thresholds $\vec{b_k}$ are generated through the following three ways. The first way is that there are totally $n_k$ tunable entries in $W_k$ and  $\vec{b_k}$, while the
remainder $d_{k}d_{k-1}+d_k-n_k$ entries are fixed. An example is    deep sparsely connected neural networks.
 The second way
 is that  $W_k$ and $\vec{b_k}$ are exactly generated by $n_k$ free parameters including weight-sharing.
  The third way is that the weight matrix is generated jointly by both the above ways.   Like the  most widely used  deep convolutional neural networks, we count  the number of free parameters according to the third way by considering both sparse connections and weight-sharing \cite{Zhou2018,Zhou2018a,Zhou2020b}. It should be mentioned that such a way to count  free parameter is different from \cite{Yarotsky2017} which  considers deep fully connected neural networks. The different way to count  free parameters is consistent with the structure of deep nets, which is the main reason  why we can improve the approximation result of \cite{Yarotsky2017}.}

 \subsection{Capacity measurements of deep nets}
 It is meaningless to pursue  the outperformance  of deep nets over shallow nets
without considering the capacity costs, since the universality of shallow nets
\cite{Cybenko1989,Leshno1993} demonstrates  that shallow nets can extract an
arbitrary data feature as long as  the network is sufficiently wide. We adopt the concept of covering number \cite{Zhou2003}  which is widely
used in statistical learning   and information theory to measure the capacity   to
  cast the comparison into a unified framework.

\begin{figure}[h]
    \centering
    \includegraphics[scale=0.42]{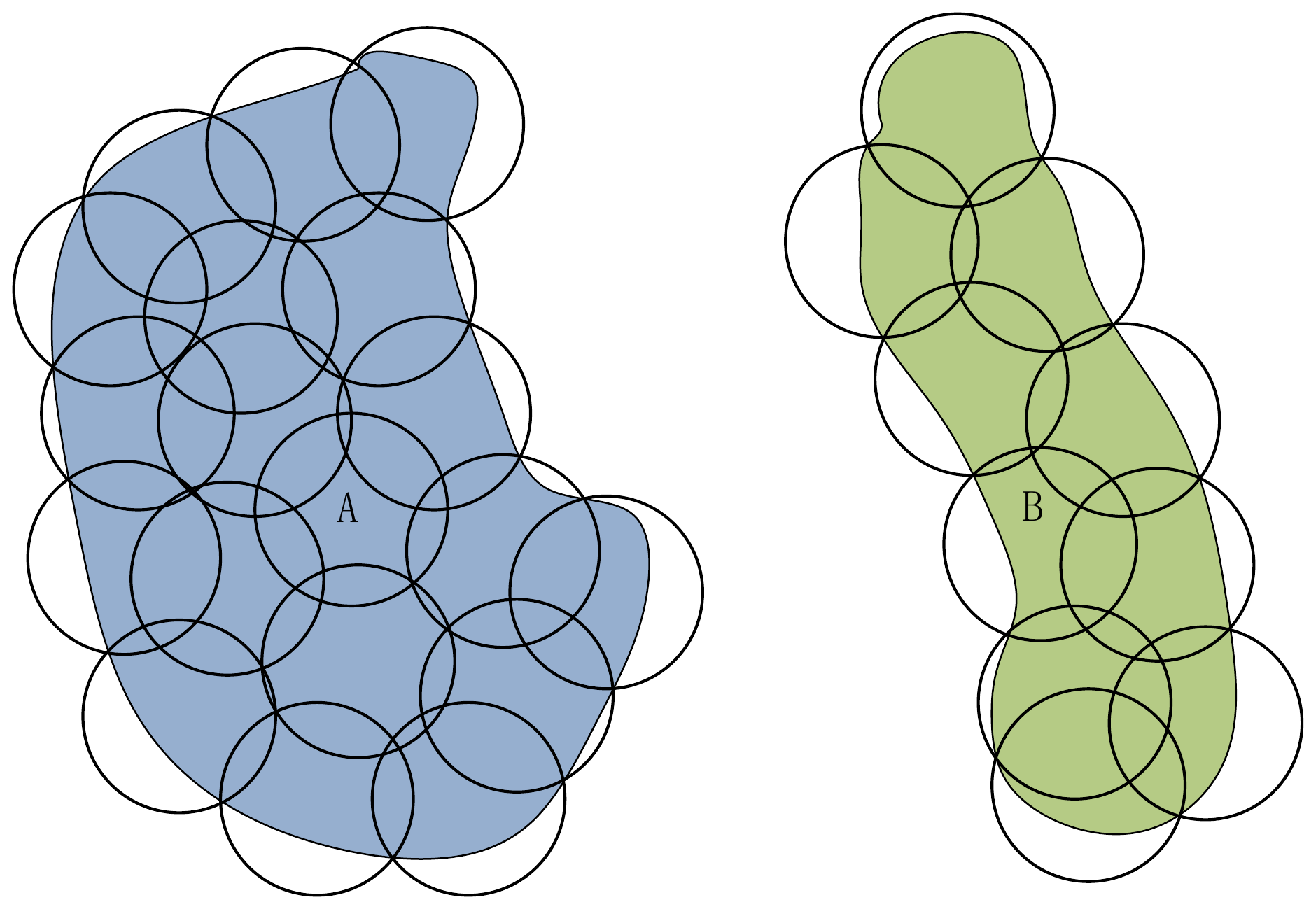}
    \caption{Covering numbers of different sets}
    \label{Fig:covering number}
\end{figure}

Let $\mathbb B$ be a Banach space and $V$ be  a subset of $\mathbb B$. Denote by $\mathcal N(\varepsilon,V,\mathbb B)$ the $\varepsilon$-covering number
of $V$ under the metric of $\mathbb B$, which is
the minimal number of elements in an $\varepsilon$-net of $V$. Intuitively, the $\varepsilon$-covering number measures the capacity of $V$ via counting the minimal number of balls in $\mathbb B$ with radius $\varepsilon$ covering $V$. Figure
\ref{Fig:covering number} showes that the $0.1$-covering number of $A$ is 19 while that of $B$ is 10, coinciding with the intuitive observation that $A$ is larger than $B$.

The quantity $H(\varepsilon, V,
\mathbb B)=\log_2\mathcal N (\varepsilon,  V, \mathbb B)$   is called the
$\varepsilon$-entropy of $  V$ in
$\mathbb B$ which is close to the coding length in information theory according to the  encode-decode theory \cite{Donoho1993}. Thus, it is a powerful capacity measurement to show the expressivity of $V$ in $\mathbb B$.  Furthermore, the $\varepsilon$-covering number determines the limitation  of approximation ability of $V$ \cite{Guo2019} and also the stability of  learning algorithms defined on $V$ \cite{Cucker2007}. All these demonstrate the rationality of adopting the covering number to measure the capacity of deep nets.

Denote by $\mathcal H_{n,L}$ the set of
all  deep nets with $L$ hidden layers, $n$
free parameters and by
\begin{eqnarray}\label{Hypothesis space}
       &&\mathcal H_{n,L,\mathcal R}:=
       \{h_{n,L}
      \in \mathcal H_{ n,L }:
       |w_{k}^{i,j}|,|b_k^{i}|,
       |a_i|\leq \mathcal R,\nonumber\\
      &&1\leq i\leq d_{k},1\leq j\leq d_{k-1},
        1\leq k\leq
      L\}
\end{eqnarray}
  the set of deep nets whose weights and thresholds are uniformly
bounded by $\mathcal R$, where $\mathcal R$ is some positive number
that may depend on $n$, $d_k$,  and $L$.  The boundedness assumption
  is  necessary since
it can be found in \cite{Maiorov1999b,Guo2019} that
there exists some deep nets with two hidden layers and
finitely many neurons possessing an infinite  covering number.

The following lemma    proved in \cite{Guo2019} presents a  tight estimate for the covering number of deep ReLU nets.

\begin{lemma}\label{Lemma:covering number}
Let $\mathcal H_{n,L, \mathcal R}$ be defined by
(\ref{Hypothesis space}). Then
\begin{eqnarray}\label{covering1}
  \mathcal N\left( \varepsilon,\mathcal H_{n,L,\mathcal
       R},L^\infty(\mathbb I^d)\right)
     \leq
   \left(C\mathcal RD_{\max}\right)^{3(L+1)^2n}\varepsilon^{-
   n},
\end{eqnarray}
where $D_{\max}:=\max_{0\leq\ell\leq L}d_\ell$ and $C$ is a
constant depending only on $d$.
\end{lemma}

It
was deduced in {  \cite[Chap. 16]{Gyorfi2002}} that
\begin{equation}\label{covering for shallow}
        \log\mathcal N(\varepsilon,\mathcal H_{n,1, \mathcal
     R}, L^1(\mathbb I^d))=\mathcal O\left(n
        \log\frac{\mathcal R}{\varepsilon}\right).
\end{equation}
   Comparing Lemma \ref{Lemma:covering number} with
(\ref{covering for shallow}), we find that, up to a logarithmic
factor, deep nets with controllable magnitudes of weights do not essentially enlarge the capacity of shallow
nets, provided that they have the same number of free parameters and the
depth of deep nets is at most $\log n$. Furthermore, Lemma \ref{Lemma:covering number} implies that the depth plays a similar role as the number of parameters in controlling the capacity of deep nets, when $\varepsilon$ is not extremely small. This shows a novel depth-parameter dilemma in controlling the capacity and is totally different from shallow nets.

%
%\begin{figure}[!t]
%\centering
%\includegraphics*[scale=0.32]{LnC.pdf}
%\hfill \caption{Depth-parameter dilemma}
%\label{Fig:Depth-parameter}
%\end{figure}

 \subsection{Limitations of shallow nets in extracting features}

The study of approximation capability of shallow nets is a classical
topic in neural networks. We refer the readers to a fruitful review paper \cite{Pinkus1999} for details on this topic. Compared with the classical linear approaches like polynomials, shallow nets with sigmoidal activation function possess  better approximation ability \cite{Mhaskar1996} and  are capable of conducting  dimension-independent error estimates {  under certain restrictions on the target functions \cite{Barron1993}}. More importantly,
   the universality
\cite{Cybenko1989,Leshno1993} showed that shallow nets can extract
any data feature as long as  the network is sufficiently wide. However,
with fixed width, they have limitations in
feature extraction, in terms of saturation \cite{Lin2019}, non-localization  \cite{Chui1994,Safran2016}, non-sparse  approximation \cite{LinH2017,Lin2018} and bottleneck in extracting the smoothness feature \cite{Maiorov1999b,Lin2017a}.
In particular, it was shown in
\cite{Maiorov1999b} that shallow nets whose capacity satisfies (\ref{covering for shallow})
  cannot extract the
smoothness features within accuracy $\mathcal O(n^{-r/(d-1)})$ with
high probability, where $r$ denotes the degree of smoothness.

For  shallow nets with   ReLU  (shallow ReLU nets), the limitation is even stricter. It was shown in \cite{Eldan2015} that  there exist some analytic univariate functions which cannot be expressible for shallow ReLU nets.
Recently,    \cite[Theorem 6]{Yarotsky2017}  proved that any
twice-differentiable nonlinear function defined on $\mathbb I^d$
cannot be $\varepsilon$-approximated by ReLU networks of fixed depth
$L$ with the number of free parameters less than
$c\varepsilon^{-1/(2(L-2))}$, where $c$ is a  positive
constant depending only on $d$. A direct consequence is that a  ReLU network  with depth $L=3$
and $n$ free parameters cannot extract the simple ``square-feature'',
i.e., $t^2$, within  accuracy $n^{-2-\tau}$ for an arbitrary $\tau>0$.
By noting $t^2$ is an infinitely differentiable  function, it is well known
that there exist  linear tools to approximate $t^2$  within accuracy of    order
$n^{-\Gamma}$ \cite{Pinkus1985} for an arbitrarily large $\Gamma<\infty$. All these
results showed that shallow nets, especially shallow ReLU nets, are
difficult to extract data  features and thus have  bottlenecks
in   complex learning  tasks.

\subsection{Necessity of  the depth for ReLU nets}
Advantages of deep nets over shallow nets were firstly revealed by \cite{Chui1994} in the sense  that
deep nets can provide localized approximation but shallow nets fail.  Since then, a great number of data features including those for sparseness, manifold structures, piecewise smoothness and rotation-invariance  are proved to be unrealizable by shallow nets but can be easily extracted by deep nets.
Under the capacity constraint  (\ref{covering1}) that is similar to (\ref{covering for shallow}) for shallow nets, the summary of advantages of deep ReLU nets in feature extraction are listed in the following Table \ref{ReLU_fea}.

\begin{table}[!h]
\caption{Deep  nets in feature extraction (within accuracy
$\varepsilon$,  $r$-smooth function and $d_m$-dimensional manifold)}\label{ReLU_fea}
\begin{center}
\begin{tabular}{|l|l|l|l|l|}
\hline Ref. & Features & Parameters & Depth\\
\hline \cite{Chui1994} & Localized approximation  & $2d+1$ & $2$\\
\hline \cite{Lin2018} & $k$-spatially sparse & $k(2d+1)$ & $2$  \\
\hline \cite{Shaham2015} & Smooth+Manifold & $\varepsilon^{-d_{m}/r}$ & $4$ \\
\hline \cite{Petersen2017} &Piecewise smooth & $\varepsilon^{-d/r}$ & Finite \\
\hline \cite{Safran2016} & $\ell_1$ radial+smooth & $\varepsilon^{-1/r}$ & $\log(\varepsilon^{-1})$   \\
\hline \cite{Schwab2018} & $k$-sparse (frequency) & $k\log(\varepsilon^{-1})$ & $\log(\varepsilon^{-1})$   \\
\hline
\end{tabular}%
\end{center}
\end{table}

To extract the ``square-feature'', the following lemma has been shown in \cite[Proposition 2]{Yarotsky2017}  to verify  that deep ReLU nets can overcome the bottleneck of shallow ReLU nets.
\begin{lemma}\label{Lemma:square gate for infinite}
The function $f(t)=t^2$ on the segment $[0,1]$ can be approximated
with any error $\varepsilon>0$ by a ReLU network having the depth
and free parameters of order $\mathcal O(\log(1/\varepsilon))$.
\end{lemma}

\begin{figure}[h]
\begin{minipage}[b]{0.49\linewidth}
\centering
\includegraphics*[scale=0.22]{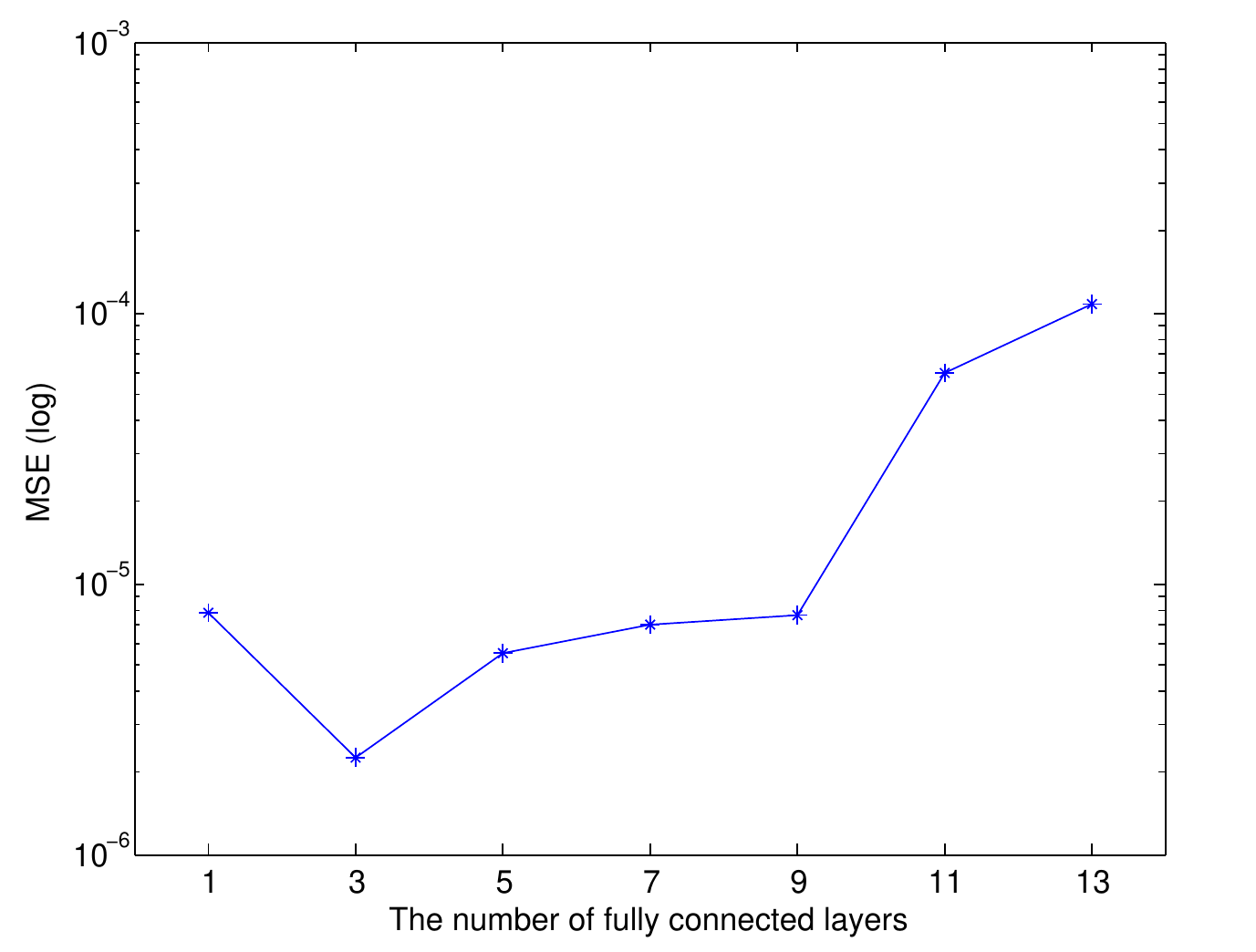}
\centerline{{\small (a) Accuracy and depth}}
\end{minipage}
\begin{minipage}[b]{0.49\linewidth}
\centering
\includegraphics*[scale=0.22]{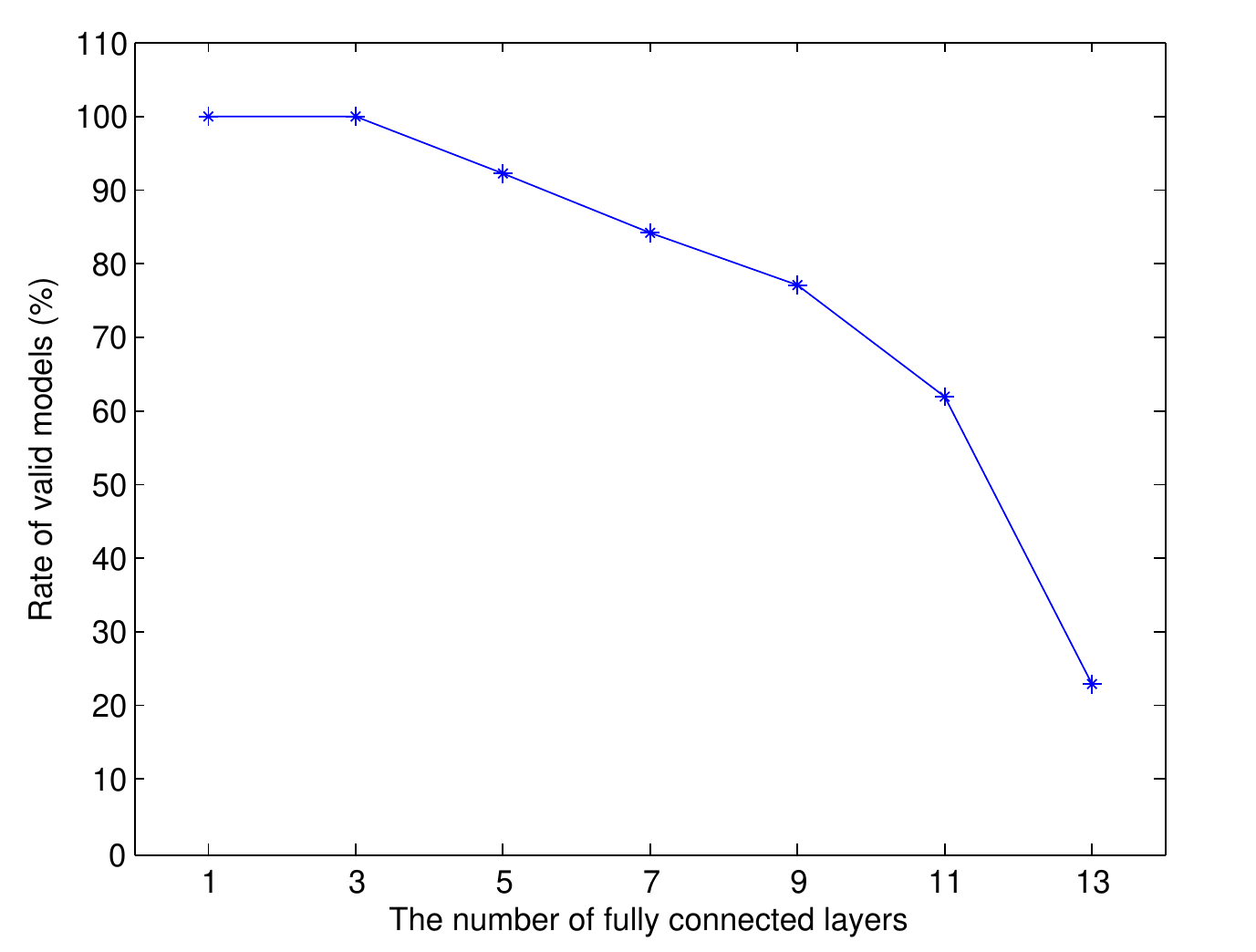}
%  \vspace{-.5cm}
\centerline{{\small (b)   {Valid model} and depth}}
\end{minipage}
\hfill
\caption{The role of depth for approximating $t^2$ using SGD }
\label{fig:algorithm-issue}
\end{figure}

Due to the  non-smoothness  of ReLU, it is { difficult} for a shallow ReLU net
with fixed width to extract smooth  features within an arbitrary
accuracy $\varepsilon$. However, by deepening the network, Lemma
\ref{Lemma:square gate for infinite} shows that deep ReLU nets
succeed in finishing such a task with only $\mathcal
O(\log(1/\varepsilon))$ free parameters. With Lemma
\ref{Lemma:square gate for infinite} and the relation $t_1\cdot
t_2=[(t_1+t_2)^2-t_1^2-t_2^2]/2$, deep ReLU nets can be used as a
``product-gate'' \cite[Proposition 3]{Yarotsky2017} to extract the
``product'' relation between   variables. Then, deep  ReLU nets
with $\mathcal O(\log(1/\varepsilon))$ hidden layers and free
parameters can approximate arbitrary polynomials defined on
$\mathbb I^d$ \cite{Schwab2018}. Therefore, even for some simple data features, deep ReLU nets
theoretically beat shallow ReLU nets, showing the necessity of the
depth in feature extraction.
However,  as shown in \cite[Sec. 8.2]{Goodfellow2016} and \cite{Allen-Zhu2018}, both the convergence issue of  the stochastic gradient descent algorithm   and the gradient vanishing phenomenon make it difficult to practically derive a deep ReLU  net estimator, which hinders the usefulness and efficiency of Lemma \ref{Lemma:square gate for infinite}.  {Figure \ref{fig:algorithm-issue} presents the difficulty for deep ReLU nets in extracting a 2-dimensional ``square-feature'' defined as $f(t)={t_1^2}+{t_2^2}$. For each depth, the network of the best performance is chosen and shown in the figure as a representation for the depth, by searching various (tens of) combinations of different widths and step sizes. The statistics of each depth are made from 100 trials. The relation between accuracy and depth is recorded in Figure \ref{fig:algorithm-issue} (a) and that between the frequencies of valid models and depth is recorded in Figure \ref{fig:algorithm-issue} (b). As shown, the network performs less robust when it gets deeper.}

\section{ Depth Selection for Extracting a Single
 Feature} \label{Sec.Trade-off1}

In this section, we introduce several   data features   and study the role of depth in extracting these data features to
  break  through the feature-depth correspondences. {  Since there are numerous symbols involved in different data features, we provide a table of notations as follows.
\begin{table}[!h]
\caption{Notations}\label{notations}
\begin{center}
\begin{tabular}{|l|l|}
\hline $L$: number of layers & $n$: number of  parameters \\
\hline $d_j$: width in the $j$-th layer & $d$: input dimension \\
\hline $r$: smoothness index &  $\mu$ sparseness index \\
\hline $\mathcal R$: bound of parameters &  $B$:  bound of polynomials  \\
\hline $d^*$:  group structure  dimension&  $\jmath$: group structure degree \\
\hline
\end{tabular}%
\end{center}
\end{table}
}

\subsection{ Data features}
In a seminal review  paper \cite{Bengio2013}, Bengio et al. presented a fruitful review on intuitive and experimental explanations for the success of deep learning in feature extraction. From a numerical viewpoint, deep nets can extract numerous data features including those for  smoothness, hierarchical organization, shared factors, manifold structures and sparsity, part of which were theoretically verified in  the recent paper \cite{Guo2019}. In particular, \cite{Guo2019} rigorously  proved that with the similar capacity costs measured by the covering number, deep nets beat  shallow nets in extracting data features listed in Table I while performing  not essentially better than  shallow nets in extracting the smoothness feature.

 %
%
%
%
%The power of deep learning in feature extraction has
%
% In supervised learning, input-output pairs $\{(x_i,y_i)\}_{i=1}^m$
%are gathered, where $x_i\in\mathbb I^d$ and
%$y_i=f^*(x_i)+\varepsilon_i$ with $\varepsilon_i$ some data noise
%and $f^*:\mathbb I^d\rightarrow\mathbb R$ the function to model the
%relation between input and output.  The aim of deep learning is to
%find an approximation of $f^*$ by deep ReLU nets by the help of the
%data. Thus, the prerequisites of deep learning in supervised
%learning is that $f^*$ can  be approximated well by deep ReLU nets.
%
%As shown in \cite[Theorem 3.1]{Gyorfi2002},  it is impossible to get
%a nontrivial generalization error bound   of a learning algorithm
%without knowing any information for the data. Thus, some features(or
%a-priori information) should be assumed at first. We introduce the
%following two definitions to describes the data features concerning
%the smoothness of  $f^*$ and structure of $x$.

Let $f^*:\mathbb I^d\rightarrow\mathbb R$ be a function to model the potential
relation between input and output, i.e., $y\approx f^*(x)$ with $x\in\mathbb I^d$ and $y\in\mathbb R$ the input variable and output variable respectively. Both structures of $x$ and properties of $f^*$ are regarded as  data features. In the following, we introduce the smoothness feature of $f^*$.
\begin{definition}\label{definition:lip for d} Let   $c_0>0$ and
$r=s+v$ with $s\in\mathbb N_0:=\{0\}\cup\mathbb N$ and $0<v\leq 1$.
We say a   function $f:\mathbb I^d\rightarrow\mathbb R$ is
$(r,c_0)$-smooth if $f$ is $s$-times differentiable and for every
$\alpha_j\in \mathbb N_0$, $j=1,\dots,d$ with
$\alpha_1+\dots+\alpha_d=s$, its $s$-th partial derivative satisfies
the Lipschitz condition
\begin{equation}\label{lip}
          \left|\frac{\partial^sf}{\partial x_1^{\alpha_1}\dots\partial
          x_d^{\alpha_d}}
          (x)-\frac{\partial^sf}{\partial x_1^{\alpha_1}\dots\partial
          x_d^{\alpha_d}}
          (x')\right|\leq c_0\|x-x'\|_2^v,
\end{equation}
where $x,x'\in\mathbb I^d$ and $\|x\|_2$ denotes the Euclidean norm of
  $x$.  Denote by $Lip^{(r,c_0)}$ the set of all
$(r,c_0)$-smooth functions defined on $\mathbb I^d$.
\end{definition}

The smoothness feature of $f^*$ illustrates that $x\approx x'$ implies $f^*(x)\approx f^*(x')$.  It is a standard feature to describe $f^*$ and has been used in a vast literature  \cite{Chui1994,Kohler2017,Guo2017,Petersen2017,Yarotsky2017,Ying2017,Lin2018CA}.
However,  it remains open  whether deep ReLU nets can achieve the optimal performance of algebraic polynomials for realizing the smoothness feature, though encouraging developments  have been made in \cite{Yarotsky2017,Petersen2017}.
Furthermore, as pointed out in \cite{Bengio2013}, the smoothness feature of $f^*$ is insufficient to get around the curse of dimensionality, which requires additional  structure features of $x$. To this end, we introduce the following group structure feature for the input.

\begin{definition}\label{definition:structure for x}
  Let $\jmath,d^*\in\mathbb N$ and $D_1,\dots,D_{d^*}\in\mathbb N$ satisfy $d=D_1+\dots+D_{d^*}$. We say $x$ possesses a $(D_1,\dots,D_{d^*})$-group structure of order $\jmath$ with respect to $f^*$, if there exists some polynomials $P_{k,\jmath}$, $k=1,\dots,d^*$   defined on $\mathbb I^{D_k}$
and of degree at most $\jmath$ and a function $g:\mathbb R^{d^*}\rightarrow\mathbb R$ such that { 
\begin{eqnarray}\label{Structure111}
       f^*(x) &=&
         g\big[P_{1,\jmath}(x^{(1)},\dots,x^{(D_1)}),\dots, \nonumber\\
      &&P_{d^*,\jmath}(x^{(d-D_{d^*}+1)},\dots,x^{(d)})\big].
\end{eqnarray}}
\end{definition}

The group structure depicts the relation between different input variables.
The case $d^*=d$ and {   $P_{k,\jmath}(t)=t$} for $k=1,\dots,d$ denotes that all variables in $x$ are independent. The rotation-invariance assumption \cite{Chui2018a} is included in the case $d^*=1$ implies that   variables in $x$ possess strong dependence. The group structure assumption is more general than the manifold assumption \cite{Chui2018}, rotation-invariance assumption \cite{Chui2018a} and sparseness assumption \cite{Schwab2018} via imposing different restrictions on $P_{k,\jmath}$.

To show the outperformance of deep nets, we impose both the smoothness assumption on $f^*$ and group structure assumption on the input. Such a  smooth-structure assumption  abounds in applications.  For   $d^*=1$ and $P_{1,\jmath}(x)=(x^{(1)})^2+\dots+(x^{(d)})^2$, the  smooth-structure assumption refers to a radial function that plays an important role in designing  earthquake early warning systems \cite{Satriano2011}. For $d^*<d$ and $P_{1,\jmath}(x^{(1)},\dots,x^{(D_1)})=(x^{(1)})^2+\dots+(x^{(D_1)})^2$, the feature assumption is related to a partially radial function that is important in predicting  the magnitude of earthquake \cite{Vikraman2016}. For  $d^*=d$ and
$$
    {  g(P_{1,\jmath}(x^{(1)}), \dots,P_{d,\jmath}(x^{(d)})=P_{1,\jmath}(x^{(1)})+\dots+P_{d,\jmath}(x^{(d)}),}
$$
the smooth-structure assumption  corresponds to the well known additive model \cite{Kohler2017} with polynomial kernels in statistics.  If there exists some $P_{k,\jmath}(\cdot)=0$, the assumption then implies   sparseness which is  standard in computer vision \cite{LinT2008}.
%Figure \ref{fig:feature-Structure} exhibits some exmaples for the mentioned features.
%\begin{figure}[!t]
%\begin{minipage}[b]{0.49\linewidth}
%\centering
%\includegraphics*[scale=0.22]{Fig4d_1.pdf}
%%  \vspace{-.5cm}
%\centerline{{\small (a) Radial function}}
%\end{minipage}
%\hfill
%\begin{minipage}[b]{0.49\linewidth}
%\centering
%\includegraphics*[scale=0.22]{Fig4d_2.pdf}
%%  \vspace{-.5cm}
%\centerline{{\small (b) Partially radial function}}
%\end{minipage}
%\hfill
%\begin{minipage}[b]{0.49\linewidth}
%\centering
%\includegraphics*[scale=0.22]{Fig4d_3.pdf}
%%  \vspace{-.5cm}
%\centerline{{\small (c) Additive model}}
%\end{minipage}
%\hfill
%\begin{minipage}[b]{0.49\linewidth}
%\centering
%\includegraphics*[scale=0.22]{Fig4d_4.pdf}
%%  \vspace{-.5cm}
%\centerline{{\small (d) Sparse model}}
%\end{minipage}
%\hfill
%\caption{Examples for the smooth-structure assumption: (a)   3-dimensional radial function:  $f(x)=\sin\|x\|_2^2/\|x\|_2^2$;  (b)  3-dimensional partially radial model  $f(x)=\sin\|(x_1,x_2)\|_2^2/\|x_1,x_2\|_2^2+x_3$; (c)   3-dimensional additive model:  $f(x)=x_1^3+x_2^2+x_3$; (d) 3-dimensional sparse model $f(x)=\sin\|(x_1,x_2)\|_2^2/\|(x_1,x_2)\|_2^2,x_3=0$}
%\label{fig:feature-Structure}
%\end{figure}

\subsection{Depth selection for  extracting the group
structure} It was shown in \cite{Mhaskar1996} that for some fixed
activation function, i.e., analytic and non-polynomials, shallow
nets with $\left(^{\beta+d}_{\ d}\right)$ neurons can approximate any
polynomial defined on $\mathbb I^d$ of degree $\beta\in\mathbb N$ within an arbitrary
accuracy. However, if the polynomial is sparse, then shallow nets
fail to catch the sparseness information \cite{LinH2017} in the
sense that the same number of neurons is required to approximate sparse
  and non-sparse polynomials. However,
\cite{LinH2017,Schwab2018} found that deep nets  essentially
improve the performance of shallow nets by using the
``product-gate'' property of deep nets \cite{Yarotsky2017}. In particular, for deep
ReLU nets, the   following lemma was  proved in \cite[Proposition
3.3]{Schwab2018}.

\begin{lemma}\label{Lemma:poly nn 2}
For any $0<\varepsilon<1$ and $\ell\in\mathbb N$, there exists a
ReLU net $\tilde{\prod}$ with $\ell$ input units,  $\mathcal
O[(1+\log \ell)\log(\ell/\varepsilon)]$ depth and $\mathcal
O[(1+\log \ell)\log(\ell/\varepsilon)]$  free parameters
 such that for any $u_1,\dots,u_\ell$
satisfying $|u_k|\leq 1, k=1,\dots,\ell$, there holds
$$
       \left|\tilde{\prod}(u_1,\dots,u_\ell)-\prod_{k=1}^\ell u_k\right|\leq
       \varepsilon.
$$
\end{lemma}

Noting that each monomial  defined on $\mathbb I^d$ of degree at
most $\beta$ can be rewritten as $\beta$ products of elements in $[0,1]$, it
requires a deep net with $\mathcal O[(1+\log
\beta)\log(\beta/\varepsilon)]$ depth and   free parameters to
extract the monomial feature  according
to Lemma \ref{Lemma:poly nn 2}. Based on the ``product-gate-units'' (PGU), we can construct a deep net   such that for any $\mu$-sparse
polynomials, there are only $\mu+\mathcal O[(1+\log \beta)\log
(\beta/\varepsilon)]$ free parameters involved  to extract this structure
feature, which is much smaller than $\left(^{\beta+d}_{\ d}\right)$ provided $\mu$
is small and $\varepsilon$ is not extremely small.

Although the above interesting result illustrates the power of depth in
 extracting structure features, the depth of the
constructed deep ReLU net depends on the approximation accuracy,
making it be  practically difficult to get a deep net estimator, just as Figure \ref{fig:algorithm-issue} purports to show. In this paper,
we pursue a trade-off between  depth and number of free parameters in extracting
structure features by using an approach developed in a recent paper
\cite{Petersen2017}. The following ``product gate'' for deep ReLU nets
is our main tool, whose proof can be found in Appendix A.

\begin{lemma}\label{lemma:product gate 2}
Let $\theta>0$  and $\tilde{L}\in\mathbb N$ with $\tilde{L}>(2\theta)^{-1}$. For any $\ell\in\{2,3,\dots,\}$ and $\varepsilon\in
(0,1)$, there exists a deep ReLU net $\tilde{\times}_\ell$
with $2\ell \tilde{L}+8\ell$
layers and at most $c\ell^{\theta} \varepsilon^{-\theta}$ free
  parameters   bounded by $\ell^\gamma\varepsilon^{-\gamma}$
 such that
$$
       |u_1u_2\cdots u_\ell-\tilde{\times}_\ell(u_1,\dots,u_\ell)|\leq
       \varepsilon,\quad \forall u_1,\dots,u_\ell\in[-1,1],
$$
where $c$ and $\gamma$ are constants depending only on $\theta$ and $\tilde{L}$.
\end{lemma}

%\begin{figure}[t]
%    \centering
%  \includegraphics[width=6.5cm,height=4cm]{PGU-to-sparse.pdf}
%    \caption{Deep nets for sparse approximation}
%    \label{fig:PGU-to-sparse}
%\end{figure}

Comparing Lemma \ref{lemma:product gate 2} with Lemma \ref{Lemma:poly
nn 2}, as a  ``product-gate'',  we reduce  the
depth of deep ReLU nets on the price of adding the number of free parameters. The positive number $\theta$ performs as a balance exponent in the sense that small $\theta$ implies large depth but few free parameters, while large $\theta$ means small depth but a great number of free parameters.
For a fixed and $\varepsilon$-independent exponent $\theta$,
the depth of ReLU nets in Lemma \ref{lemma:product gate 2} is
independent of the accuracy $\varepsilon$, while the number of free parameters increases
from $\mathcal O[(1+\log \ell)\log (\ell/\varepsilon)]$ to
$\ell^{\ell\theta} \varepsilon^{-\ell\theta}$.
 Therefore, Lemma
\ref{lemma:product gate 2} exhibits   a trade-off between
depth and  parameters and removes the feature-depth correspondence.

 Denote by $\mathcal P_\beta^d$  the
set of algebraic polynomials defined on $\mathbb I^d$ with degree
at most $\beta$. For $B\geq 1$, define further $
     \mathcal P_{\beta,B}^d:=\left\{\sum_{|\alpha|\leq \beta}c_\alpha
     x^\alpha:|c_\alpha|\leq B\right\}
$ the  set of polynomials in $\mathcal P_\beta^d$ whose coefficients
are uniformly bounded by $B$, where
$\alpha=(\alpha_1,\dots,\alpha_d)\in\mathbb N_0^d$,
$|\alpha|=\alpha_1+\dots+\alpha_d$ and
$x^\alpha=(x^{(1)})^{\alpha_1}\cdots(x^{(d)})^{\alpha_d}$. Define
$\mathcal P_{\beta,B,\mu}^d$ the set of all $\mu$-sparse polynomials
in $\mathcal P_{\beta,B}^d$. So $P\in \mathcal P_{\beta,B,\mu}^d$ has at most
$\mu$ nonzero coefficients. The following theorem shows the
  performance of deep ReLU nets in extracting the sparse polynomial
  feature, whose proof will be given in Appendix A.

\begin{theorem}\label{Theorem:polynomial}
Let $\beta,\mu\in\mathbb N$,   $B,\theta>0$  and $\tilde{L}\in\mathbb N$ with $\tilde{L}>(2\theta)^{-1}$. For any
$0<\varepsilon<1$,
 there is a deep ReLU net structure
 with     $2\beta \tilde{L}+8\beta+1$ layers and
 at most
$\mu+c(\mu\beta B)^{\theta}
\varepsilon^{-\theta}$ nonzero parameters bounded by
$\max\{B,(\mu\beta B)^\gamma\varepsilon^{-\gamma}\}$,
 such that for each $P\in\mathcal P^d_{\beta,B,\mu}$ there exists a $h_P$ with the aforementioned structure satisfying
$$
       |P(x)-h_P(x)|\leq \varepsilon,\qquad\forall x\in \mathbb I^d,
$$
where $c$ and $\gamma$ are the constants  in Lemma
\ref{lemma:product gate 2}.
\end{theorem}

A similar result has been established in \cite{Schwab2018} for deep ReLU nets with depth $\mathcal O (\log \beta\log(\mu B\beta/\varepsilon))$ and  number of free parameters $\mu+ \mathcal O (\log \beta\log(\mu B\beta/\varepsilon))$.  Our result is different from \cite{Schwab2018} by introducing an exponent $\theta$ to balance the depth and number of free parameters. For a fixed $\theta$, the depth of deep ReLU nets studied in Theorem \ref{Theorem:polynomial} is independent of $\varepsilon$. Thus, Theorem \ref{Theorem:polynomial} shows a novel
relation between the depth and feature extraction for   sparse polynomial features as well as the group
structure features, by means of breaking through the exclusive  feature-depth correspondence in  \cite{Schwab2018}. Furthermore, if $\mu$ is not extremely small,  we can select a $\theta$ in Theorem \ref{Theorem:polynomial} such that  the capacity of deep nets in Theorem  \ref{Theorem:polynomial} is  smaller than that in \cite{Schwab2018} according to Lemma \ref{Lemma:covering number}. That is, Theorem \ref{Theorem:polynomial} provides a theoretical guidance on using  smaller capacity costs than \cite{Schwab2018} to get a same  accuracy in extracting the sparse feature.

\subsection{Deep nets for  extracting the
smoothness feature}
In \cite{Yarotsky2017}, Yarotsky succeeded in establishing a tight error estimate of approximating smooth functions by deep ReLU nets by
utilizing the ``product-gate'' property in Lemma \ref{Lemma:poly nn 2}. \cite[Theorem 1]{Yarotsky2017} showed that for any $f\in Lip^{(r,c_0)}$ with $r\in\mathbb N$, there is a deep ReLU net $h^\diamond_f$ with fixed structure,  $n$ free parameters and $\mathcal O(\log n)$ layers such that
\begin{equation}\label{Jackson21}
       \|f-h^\diamond_f\|_{L^\infty(\mathbb I^d)}\leq c'n^{-r/d}\log n,
\end{equation}
where $c'$ is a constant depending only on $c_0$, $d$, $r$ and $p\in[1,\infty)$. Comparing with  standard  results for linear approximants  such as   algebraic polynomials \cite{Pinkus1985}, there is an additional logarithmic {  term in (\ref{Jackson21})}.
This is due to the accuracy-dependent  depth   in Lemma \ref{Lemma:poly nn 2}.

This phenomenon was firstly noticed in \cite{Petersen2017}. After deriving
the ``product-gate'' property for deep ReLU nets with accuracy-independent depth,  \cite[Theorem 3.1]{Petersen2017} proved that there exists a deep ReLU net $h_f^*$
with fixed structure, $n$ free parameters layered on $(2+\lceil\log r\rceil)(11+r/d)$
hidden layers such that
\begin{equation}\label{Jackson3}
       \|f-h^*_f\|_{L^p(\mathbb I^d)}\leq c^*n^{-r/d},
\end{equation}
where $c^*$ is a constant depending only on $c_0$, $d$, $r$ and $p\in[1,{\infty})$.
It is obvious that (\ref{Jackson3}) improves {  (\ref{Jackson21})} by removing the logarithmic term. However, the analysis in \cite{Petersen2017} relies on
the localized approximation \cite{Chui1994}  of
deep nets  and thus, their result  holds only under the $L^p(\mathbb
I^d)$ norm with $1\leq p<\infty$. Noting that for  $f\in L^\infty(\mathbb I^d)$, $\|f\|_{L^p(\mathbb I^d)}\leq \|f\|_{L^\infty(\mathbb I^d)}$, (\ref{Jackson3})  does not match the optimal rate of uniform approximation by linear approximants.  In
the following theorem, we combine the approaches in \cite{Petersen2017} and
\cite{Yarotsky2017} to get a sharp error estimate of approximating smooth functions by deep ReLU nets  under the
$L^\infty(\mathbb I^d)$ metric.

%\begin{figure}[h]
%    \centering
%  \includegraphics[width=8cm,height=6cm]{structure-for-smooth.pdf}
%    \caption{Deep nets for approximating smooth functions}
%    \label{fig:PGU-for-smooth}
%\end{figure}

\begin{theorem}\label{Theorem:jackson}
Let $r=s+v$ with $s\in\mathbb N_0$ and $0<v\leq 1$, $c_0,\theta>0$  and $\tilde{L}\in\mathbb N$ with $\tilde{L}>(2\theta)^{-1}$.
For any   $\varepsilon\in(0,1)$,
  there exists a
deep ReLU net  structure  with
\begin{equation}\label{number of layer}
     \mathcal L(d,r,\tilde{L}):=2(d+s)\tilde{L}+8(d+s)+3
\end{equation}
layers and  at most $c(d+s)^{\theta} \varepsilon^{-(r+d)\theta/r}+(8d+5)\left(^{s+d}_{\ s}\right)\varepsilon^{-d/r}$  free
parameters bounded by $\max\{\tilde{B},3\varepsilon^{-1/r},(d+s)^\gamma\varepsilon^{-(r+d)\gamma/r}\}$, such that for any $f\in Lip^{(r,c_0)}$ there is a $h_f$ with the aforementioned structure satisfying
\begin{equation}\label{Jackson11}
       \|f-h_f\|_{L^\infty(\mathbb I^d)}\leq c_1\varepsilon,
\end{equation}
where $c_1$ is a constant  depending only on $c_0$, $d$ and $r$ and
$$
   \tilde{B}:=\max_{k_1+\dots+k_d\leq s}\max_{x\in\mathbb I^d}\left|\frac{1}{k_1!\dots
    k_d!}\frac{\partial^{k_1+\dots
    k_d}f(x)}{\partial^{k_1}x^{(1)}\dots\partial^{k_d}x^{(d)}}\right|.
$$
\end{theorem}

The proof of Theorem \ref{Theorem:jackson} will be presented in Appendix B.
Setting $\varepsilon=n^{-r/d}$, we get from Theorem \ref{Theorem:jackson} that there exists a deep net $h_f$  with at most $\mathcal O(n^{\max\{1,(r+d)\theta/d\}})$ free parameters and $ \mathcal L(d,r,\tilde{L})$ layers for  $\tilde{L}>(2\theta)^{-1}$  such that
\begin{equation}\label{Jackson2}
        \|f-h_f\|_{L^\infty(\mathbb I^d)}\leq c_1 n^{-r/d}.
\end{equation}
The  depth plays a crucial role in extracting the smooth features in the sense that to derive a similar  approximation accuracy as  linear approximants, $\theta$ should be not larger than $d/(r+d)$, implying  $\tilde{L}>(r+d)/(2d)$. However, when the depth $ \mathcal L(d,r,\tilde{L})$ with $\tilde{L}$ 
reaching this critical value, deep nets  with various depths are capable of extracting  smooth features. This  removes the feature-depth correspondence in extracting the smooth feature {  by making use of the structure of deep nets, since our constructed deep nets in the proof are sparse and share weights, which is different from the deep nets in the prominent work \cite{Yarotsky2017}}.
  Recalling Lemma \ref{Lemma:covering number}, for appropriately selected $\theta$, the capacity of deep nets in our construction is smaller than that of \cite{Yarotsky2017} by removing the logarithmic term caused by the accuracy-dependent layers.
Inequalities like (\ref{Jackson11}) have been
established for  shallow
nets with some sigmoid-type activation functions \cite{Mhaskar1996,Lin2014b}. However, different from Theorem
\ref{Theorem:jackson}, the magnitudes of weights in
\cite{Mhaskar1996} are so large that the capacity restriction (\ref{covering for shallow}) does not hold and the result in \cite{Lin2014b} suffers from the well known saturation phenomenon  in the sense that the approximation rate
cannot be improved any further  when the smoothness of the target
function goes beyond a specific level.
  It can be found in Theorem \ref{Theorem:jackson} that
deepening  the networks succeeds in  overcoming  these problems.
Although  \cite[Theorem 2]{Guo2019} declares that to extract the smoothness feature, deep nets perform not essentially better than shallow nets or linear approximant, our result in Theorem \ref{Theorem:jackson} yields that deep ReLU nets are at least  not worse than shallow nets.

\section{Depth Selection in  Exacting Composite Features}\label{Sec.Radial}

The previous section demonstrated  the role of depth in
extracting a single data feature. However, as shown in \cite{Bengio2013}, it is much more important to simultaneously extract  multiple features to feed the target-driven learning.  Extracting  composite features by deep nets, which is the purpose of this section, brings novel challenges in designing  deep nets, including the junction of deep nets with different utilities,   the balance of  accuracy   and  depth, and the depth-parameter trade-off.

To build up a network  to exact composite features, an intuitive approach is to stack deep nets by the a-priori information or  human experiences  in   a tandem  manner, just as  Figure \ref{Fig:flow}  implies.    The problem is, however, such a  brutal  stacking is practically inefficient, for both   the unavailability of the   a-priori information and  lacking of the prescribed accuracy for extracting a specific feature. More importantly, the stacking scheme requires much more free parameters and depths of deep nets to extract composite features, adding additional capacity costs according to Lemma \ref{Lemma:covering number}
%\begin{figure}[h]
%    \centering
%    \includegraphics[scale=0.4]{figPT.pdf}
%    \caption{Stacking strategy for composite features}
%    \label{fig:Stacking}
%\end{figure}

In this section, we provide some theoretical
guidance on selecting depth of deep nets to   extract composite
features by taking the depth-parameter trade-off into account. Without loss of generality, we are interested in extracting  features exhibited in the following assumption.
\begin{assumption}\label{Assumption:Features}
Let $r=s+v$ with $s\in\mathbb N_0$ and $v\in(0,1]$,  $D_1,\dots,D_{d^*},d^*\in\mathbb N$ with $d=D_1+\dots+D_{d^*}$, and
$\jmath,\mu\in \mathbb N$.
Assume that there is a function $g$ defined on $\mathbb I^{d^*}$ satisfying $g\in Lip^{(r,c_0)}$ such that (\ref{Structure111}) holds with
$P_{k,\jmath}\in \mathcal P_{\jmath,1/2,\mu}^{D_k}$ for $k=1,\dots,d^*$.
\end{assumption}

There are totally three types of features in Assumption \ref{Assumption:Features},   the smoothness feature of $g$ as well as  $f^*$, the group structure feature of $x$, and the sparsity feature of the structure polynomials $P_{k,\jmath}$, $k=1,\dots,d^*$.  An intuitive observation is that the depth and number of free parameters of deep nets to simultaneously extract these three features  should be larger than those to extract each single feature. However, as shown in the following theorem, it is not necessarily the case, provided the deep nets for different utilities are appropriately combined.
%
%\begin{figure}[t]
%    \centering
%  \includegraphics[width=6.5cm,height=4cm]{structure-for-composite.pdf}
%    \caption{Deep nets for extracting composite features}
%    \label{fig:Structure-for-composite-features}
%\end{figure}

\begin{theorem}\label{Theorem:jackson for trade-off}
Let $r=s+v$ with $s\in\mathbb N_0$ and $v\in(0,1]$, $d^*,\jmath,\mu\in\mathbb N$, $c_0,\theta>0$  and $\tilde{L}\in\mathbb N$ with $\tilde{L}>(2\theta)^{-1}$. For any   $0<\varepsilon<1/2$, there exists
  a deep ReLU net structure  with at most
\begin{equation}\label{layerforcompositi}
       \mathcal L^*(d^*,r,\tilde{L},\jmath)=\overbrace{\mathcal
       L(d^*,r,\tilde{L})}^{\mbox{smooth+group}}+\overbrace{2\jmath \tilde{L}+8\jmath+1}^{\mbox{group}}
\end{equation}
layers and at most
\begin{eqnarray}\label{neuronsforcompositi}
   &&\mathcal W(d^*,\varepsilon,\mu,\jmath,\theta):= \overbrace{(8d^*+5)\left(^{s+d^*}_{\ s}\right)\varepsilon^{-d^*/r}}^{{\mbox{smooth+group}}}\\
   &+&
   \overbrace{ \mu d^*}^{\mbox{sparse+group}}+\overbrace{c(d^*+s)^{\theta} \varepsilon^{-(r+d^*)\theta/r}
    +
   c(\mu\jmath)^{\theta}
    \varepsilon^{-\theta/\tau_r}}^{\mbox{depth-parameter  trade-off}} \nonumber
\end{eqnarray}
 free parameters bounded by
\begin{equation}\label{R}
    \max\{\tilde{B}_g,3\varepsilon^{-1/r},(d^*+s)^\gamma\varepsilon^{-(r+d^*)\gamma/r},(\mu\jmath)^\gamma\varepsilon^{-\gamma/\tau_r}\}
\end{equation}
  such that, for any $f^*$ satisfying
 Assumption \ref{Assumption:Features}, there is an $h_{f^*}$ possessing the aforementioned structure satisfying
$$
       \|f^*-h_{f^*}\|_{L^\infty(\mathbb I^d)} \leq c_2\varepsilon
$$
 where $\tau_r=\left\{\begin{array}{cc}
        1,& r\geq 1,\\
        v,& r<1,
 \end{array}\right.$ and
       $c_2,\tilde{B}_g$ are constants  depending only on $c_0,r,d^*$ and
       $g$.
\end{theorem}

The proof of Theorem \ref{Theorem:jackson for trade-off} will be given in Appendix C. Assumption \ref{Assumption:Features} implies  $f^*\in Lip^{(r,c_0)}$, which requires $\mathcal L(d,r,\tilde{L})$  layers to extract the smoothness feature according to Theorem \ref{Theorem:jackson}. However, with the help of the group structure feature, (\ref{layerforcompositi}) exhibits a reduction of layers from $\mathcal L(d^*,r,\tilde{L})$ to $\mathcal L(d,r,\tilde{L})$. To extract the group structure feature itself,   additional $2\jmath \tilde{L}+8\jmath+1$ layers are required. This shows that the classical tandem
stacking  is not necessary.  In particular, for some specific group  structure features, taking $d^*=1$ and $\jmath=1$ for example, it is easy to select some $\theta>0$ such that
$
         \mathcal L^*(d^*,r,\tilde{L},\jmath)\leq \mathcal L(d,r,\tilde{L}),
$
implying a waste of source of   the  tandem
stacking.

The number of free parameters, as exhibited in (\ref{neuronsforcompositi}), reflects the price to pay for extracting three composite features.  To yield an accuracy of  order $\varepsilon$, the group structure and smoothness feature require  at least $\varepsilon^{-d^*/r}$ free parameters. It should be mentioned that this number cannot be reduced further according to \cite[Theorem 2]{Guo2019} by noting in Assumption \ref{Assumption:Features} that $f^*$ corresponds a smooth function   defined on $\mathbb I^{d^*}$. The second term in (\ref{neuronsforcompositi}) reflects the difficulty in extracting the group structure feature. Without the sparseness assumption, it requires at least $\mathcal O\left(\sum_{k=1}^{d^*}\jmath^{D_k}\right)$ free parameters. If there is some $k$ such that $\jmath^{D_k}>\varepsilon^{-d^*/r}$, extracting the group structure feature  becomes the main difficulty in the learning process.
This imposes a strict restriction on $\jmath$ to maintain the optimality.
The sparsity assumption reduces this risk, allowing $\jmath$ to be very large. The rest two term in (\ref{neuronsforcompositi})
illustrates a depth-parameter trade-off in extracting composite features. In particular, to guarantee the optimal capability of feature extraction, $\theta$ must be smaller than the critical value $\theta_0:=\min\left\{\frac{d^*}{d^*+r},\frac{d^*\tau_r}{r}\right\}$. This implies a smallest depth in  (\ref{layerforcompositi}) by noting $\tilde{L}>1/(2\theta)$. In a word,
less parameters {  requires} smaller $\theta$, which results in larger $\tilde{L}$ and consequently larger $\mathcal L^*(d^*,r,\tilde{L},\jmath)$, while more parameters  require  larger $\theta$, and consequently smaller $\tilde{L}$ and depth.

As Theorem \ref{Theorem:jackson for trade-off} shows, the depth of network  is not unique to extract composite features, provided it is larger than a certain level. Furthermore, our results imply  two important advantages of deep nets in feature extraction. One is that, different from the classical   tandem tackling, deep nets succeed  in extracting composite features by embodying their interactions,  and thus reduce  the capacity costs. Such a reduction plays an important role in generalization, which will be analyzed in the next section. The other is the versatility of deep nets in extracting both single features and composite features in the sense that each feature corresponds to numerous depths, and vice versa.
To end this section, we present two corollaries for deep nets to extract composite features. The first one is the smoothness and radial features. Let $d^*=1$ and
$P_{1,\jmath}(x)=\frac1{\sqrt{d}}[(x^{(1)})^2+\dots+(x^{(d)})^2]$, then
$f^*$ is a radial function \cite{Chui2018a}.  Setting $\theta=\tau_r/(2+2r)$ and $\tilde{L}=2(r+1)/\tau_r$,  we have from Theorem \ref{Theorem:jackson for trade-off} with $\jmath=2$ and $\mu=1$  the following corollary directly.

\begin{corollary}\label{Corollary:radial}
There exists a deep ReLU
net structure with
$
     4(d+s+2)(r+1)/\tau_r +8(d+s) +20
$
layers and at most $
      c_3n
$ nonzero free parameters bounded by $c_4    n^{\max\{1,(r+1)\gamma,\gamma r/\tau_r\}}$ such that for any radial function $f^*\in Lip^{(r,c_0)}$ there is a deep net $h_{f^*}$ with the aforementioned structure satisfying
$$
        \|f^*-h_{f^*}\|_{L^\infty(\mathbb I^d)} \leq c_5n^{-r},
$$
$c_3,c_4,c_5$ are constants depending only on $c_0,r,d$ and $f^*$.
\end{corollary}

The
derived approximation rate is almost optimal according to \cite{Chui2018a} in the sense that the best approximation error for all deep nets satisfying the capacity restriction (\ref{covering1}) with $n$ parameters is of order  $(n/\log n)^{-r}$.  Our second corollary considers  using deep nets to simultaneously extract the partially radial and smooth features.
Let $d'\leq d$ and
$f^*(x)=f(\overbrace{x^{(1)},\dots,x^{(d')}}^{\mbox{radial}},x^{(d'+1)},\dots,x^{(d)})
=g(t_{d'},x^{(d'+1)},\dots,x^{(d)})$ with
$t_{d'}=(d')^{-1/2}((x^{(d_1)})^2)+\dots+(x^{(d')})^2)\in [0,1]$. Let $\theta=\frac{(d-d'+1)\tau_r}{2(d-d'+1+r)}$ and $L=\frac{2(d-d'+1+r)}{(d-d'+1)\tau_r}$. The following corollary is a direct consequence of Theorem
\ref{Theorem:jackson for trade-off} with $d^*=d-d'+1$, $\jmath=2$ and $\mu=1$.

\begin{corollary}\label{Corollary:partial radial}
 There exists a deep ReLU
net structure with
$$
     \frac{4(d-d'+1+r)(d-d'+3+s)}{(d-d'+1)\tau_r}+8(d-d'+1+s)+20
$$
layers and at most $
      c_6n
$  free parameters bounded by $c_{7}  n^{\max\{1,(r+d-d'+1)\gamma,\gamma r/\tau_r\}/(d-d'+1)}$ such that for any partially radial function $f^*\in Lip^{(r,c_0)}$ there is a deep net  $h_{f^*}$ with the aforementioned structure satisfying
$$
        \|f^*-h_{f^*}\|_{L^\infty(\mathbb I^d)} \leq c_{8}n^{-r/(d-d'+1)},
$$
where $c_6,c_{7},c_{8}$ are constants depending only on $c_0,r,d$ and $f^*$.
\end{corollary}

\section{ Generalization Capability of Deep Nets}
  This section  aims at   the generalization capability of deep ReLU nets. Our analysis  is carried out in the standard learning theory
framework \cite{Cucker2007} for regression.
In learning theory \cite{Cucker2007}, a sample
$D_m=\{(x_i,y_i)\}_{i=1}^m$  with $x_i\in\mathcal X=\mathbb I^d$ and $y_i\in\mathcal
Y\subseteq[-M,M]$ for some $M>0$  is  assumed to be drawn independently according to an
unknown Borel probability $\rho$ on ${\mathcal Z}={\mathcal X}\times
{\mathcal Y}$.
The generalization capability of an estimator $f$ is measured by the generalization error,  $\mathcal E(f):=\int_{\mathcal
Z}(f(x)-y)^2d\rho$, which quantifies the relation between the sample size $m$ and prediction accuracy.
The primary objective is to find an estimator based on $D_m$ of  the regression function
$f_\rho(x)=\int_{\mathcal Y} y d\rho(y|x)$ that minimizes the
generalization error, where $\rho(y|x)$ denotes the conditional
distribution at $x$ induced by $\rho$.

Let $\mathcal H_{n,L,\mathcal R}$ be defined by (\ref{Hypothesis space}).
We  consider
  generalization error estimates for the following empirical risk minimization (ERM) on deep nets:
\begin{equation}\label{ERM}
      f_{D,n,L}:=\arg\min_{f\in \mathcal
      H_{n,L,\mathcal R}}\frac1m\sum_{i=1}^m[f(x_i)-y_i]^2.
\end{equation}
Since $|y_i|\leq M$, it is natural to project the final output
$f_{D,n,L}$ to the interval $[-M, M]$ by the truncation operator
$\pi_M
f_{D,n,L}(x):=\mbox{sign}(f_{D,n,L}(x))\min\{|f_{D,n,L}(x)|,M\}.$

\begin{figure}[!t]
\begin{minipage}[b]{0.49\linewidth}
\centering
\includegraphics*[scale=0.32]{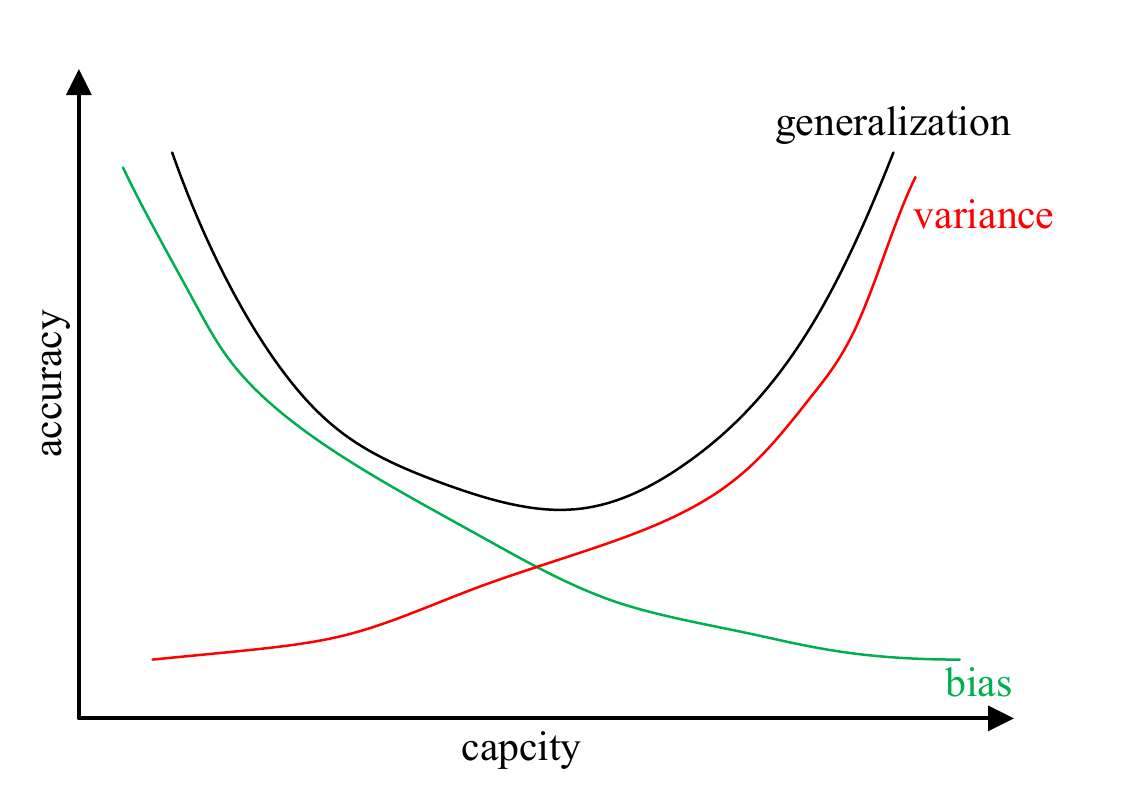}
%  \vspace{-.5cm}
\centerline{{\small (a) Bias-variance trade-off}}
\end{minipage}
\hfill
\begin{minipage}[b]{0.49\linewidth}
\centering
\includegraphics*[scale=0.24]{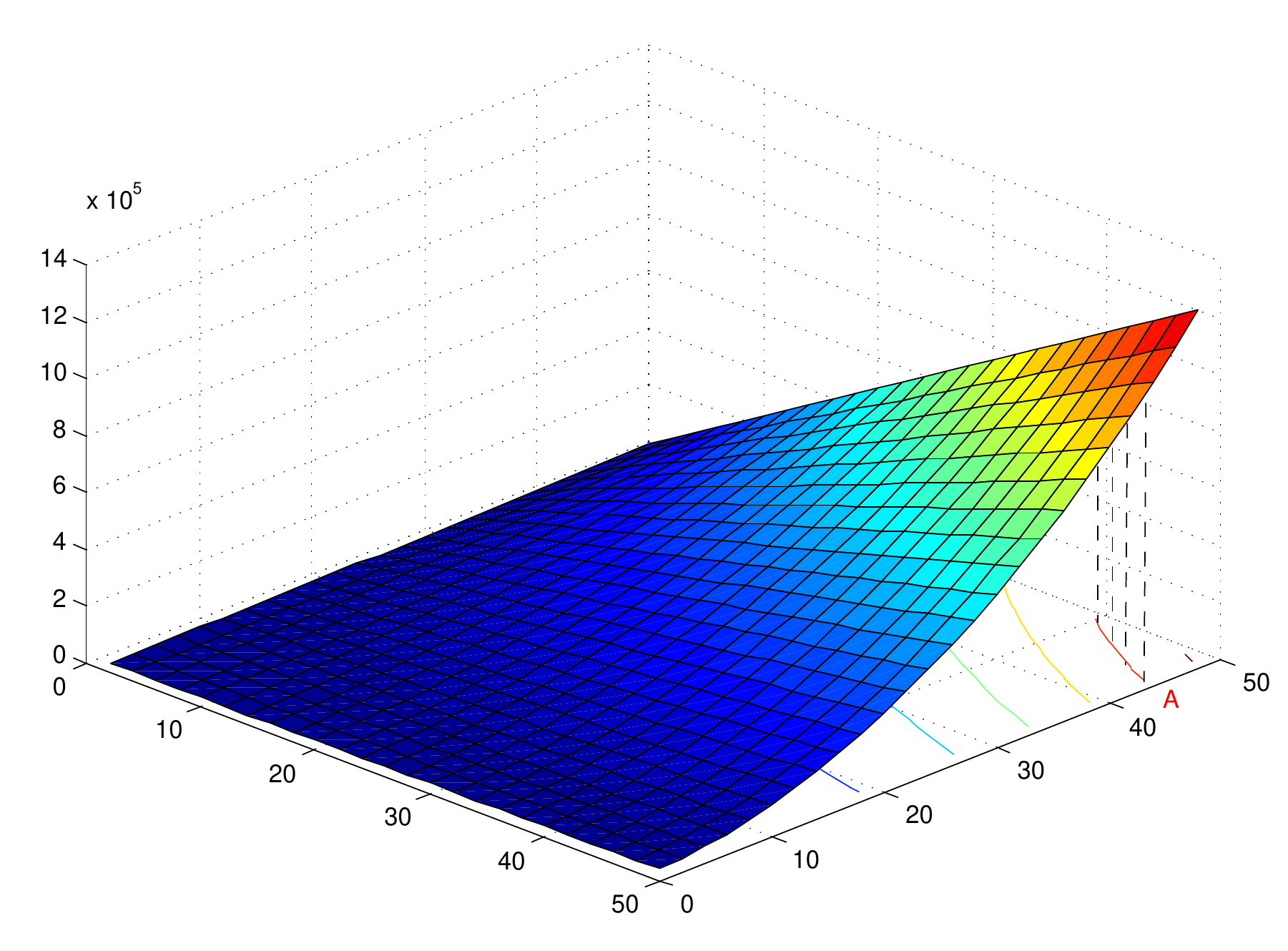}
%  \vspace{-.5cm}
\centerline{{\small (b) Depth-width trade-off}}
\end{minipage}
\hfill
\caption{Bias-variance trade-off for ERM on deep nets}
\label{fig:bias-variance}
\end{figure}

From  Theorems \ref{Theorem:polynomial}-\ref{Theorem:jackson for trade-off}, the accuracy of feature extraction decreases as the capacity of deep nets increases, resulting in  small bias for the ERM. However,  too large capacity makes ERM be sensitive to noise and  leads to  large variance. This is the well known bias-variance dilemma \cite[Chap.1]{Cucker2007}. The optimal generalization performance for ERM is obtained by balancing the bias and variance, just as Figure  \ref{fig:bias-variance} (a) purports to show. For ERM on deep nets, the problem is that the capacity depends on both   depth and  number of free parameters. As shown in Figure  \ref{fig:bias-variance} (b),  all $(L,n)$ pairs in the  curve ``A'' share the same  covering number bounds in Lemma \ref{Lemma:covering number} with $\varepsilon=0.01$.   In summary, there are two dilemmas to get a good generalization for ERM on deep nets: bias-variance dilemma in selecting the capacity and depth-parameter dilemma in controlling the bias.
The purpose of our study is not only to pursue the optimal generalization error for ERM on deep nets, but also to derive feasible candidates of $(L,n)$ pairs  to realize the optimality. The main result is the following theorem.

\begin{theorem}\label{Theorem: ERM}
Let $0<\delta<1$, $r=s+v$ with $s\in\mathbb N_0$ and $0<v\leq 1$, $\mu,\jmath,d,d^*\in\mathbb N$ and $f_{D,n,L^*}$ be defined by (\ref{ERM}) with  $L^*=\mathcal L^*(d^*,r,\tilde{L},\jmath)$, $\mathcal R$ be the value given in (\ref{R}), $ n=\left[C_1m^{\frac{d^*}{2r+d^*}}\right]$ and elements in $\mathcal H_{L,\mathcal R,n}$ possess the same structure as that in Theorem \ref{Theorem:jackson for trade-off}.    If   $f_\rho$ satisfies Assumption \ref{Assumption:Features}, $\tilde{L}>(2\theta)^{-1}$ with
\begin{equation}\label{restriction on theta}
   0< \theta\leq \theta_0:=\min\left\{\frac{d^*}{d^*+r},\frac{d^*\tau_r}{r}\right\},
\end{equation}
and
\begin{equation}\label{condition1th}
   \mu\jmath\leq n^{\frac{\tau_rd^*+\theta}{d^*\tau_r\theta}},\qquad \mbox{and}\quad \mu\leq n,
\end{equation}
then
\begin{equation}\label{learning rate}
      \mathcal E(\pi_Mf_{D,n,L^*})-\mathcal E(f_\rho)
      \leq
      C_2\tilde{L}^2m^{-\frac{2r}{2r+d^*}}\log m \log\frac3\delta
\end{equation}
holds  with confidence of at least $1-\delta$,
%Furthermore,
%\begin{eqnarray}\label{almost optimal learning rate}
%          &&C_3 m^{-\frac{2r}{2r+d^*}}
%            \leq
%         \sup_{\rho}E\left\{\mathcal E(\pi_Mf_{D,n,L^*})-\mathcal
%         E(f_\rho)\right\}\nonumber\\
%          &\leq&
%          C_4 \tilde{L}^2 m^{-\frac{2r}{2r+d^*}}\log m,
%\end{eqnarray}
%where the supremum runs over the set of all $f_\rho$ satisfying Assumption \ref{Assumption:Features} and
 where $C_1,C_2$  are constants  independent of $\delta$, $m$, $\tilde{L}$ or
  $n$.
\end{theorem}

The proof of Theorem \ref{Theorem: ERM} will be given in Appendix D.
 Condition (\ref{condition1th}) presents a restriction on the group structure and sparsity features in the sense that either $\mu$ or $\jmath$ should be relatively  small with respect to the size of data. It should be noted that the established learning rate cannot be essentially improved in the sense that for some special $P_{k,\jmath}$, the learning rate is optimal \cite[Theorem 3]{Chui2018a}.
We  emphasize that the learning rate in the above theorem is much
better than the optimal learning rate $m^{-\frac{2r}{2r+d}}$ for
learning $(r,c_0)$-smooth functions on $\mathbb I^d$
\cite{Gyorfi2002,Lin2018}.
 As pointed out in \cite{Chui2018a}, even restricting to learning a radial function,   to achieve a
learning rate similar to (\ref{learning rate}) with $d^*=1$, it requires at least $[m^{\frac{d-1}{2r+1}}]$ neurons to guarantee the generalization error of order
$ m^{-\frac{2r}{2r+1}} $.  This shows the outperformance of deep nets over shallow nets.

Theorem \ref{Theorem: ERM} also shows the almost optimal learning rates for deep nets in learning data with the group structure and smoothness features. The
relation $ n=\left[C_1m^{\frac{d^*}{2r+d^*}}\right]$ is obtained by  the bias-variance trade-off principle. It should be mentioned that there   is an additional $L^2$ in right-hand sides of (\ref{learning rate}), implying  advantages of small depth. However,  (\ref{restriction on theta}) implies that there is a critical depth, larger than which deep nets with suitable structures can achieve the almost optimal generalization error bounds. It also exhibits a depth-parameter trade-off in the sense that small $\theta$ results in small number of parameters. A different $\theta$  thus provides different $(L,n)$ pairs to    realize the optimal generalization performance of ERM on deep ReLU nets.

\section{Experimental Results}\label{Sec.Numerical}

In this section, we present both toy simulations and real data experiments to show the roles of depth   for ReLU nets in feature selection and prediction. All the numerical experiments are carried out in the Python-3.5.4 environment running on a workstation with a Pascal Titan X 12-GB GPU and 24-GB memory. Our implementation is derived from the publicly available Tensorflow-1.4.0 framework by using AdamOptimizer. {  Our codes are available at http://vision.sia.cn/our\%20team/Hanzhi-homepage/vision-ZhiHan\%28English\%29.html.}

\subsection{Experimental setting}
\begin{figure}
    \centering
    \includegraphics[width=1\linewidth]{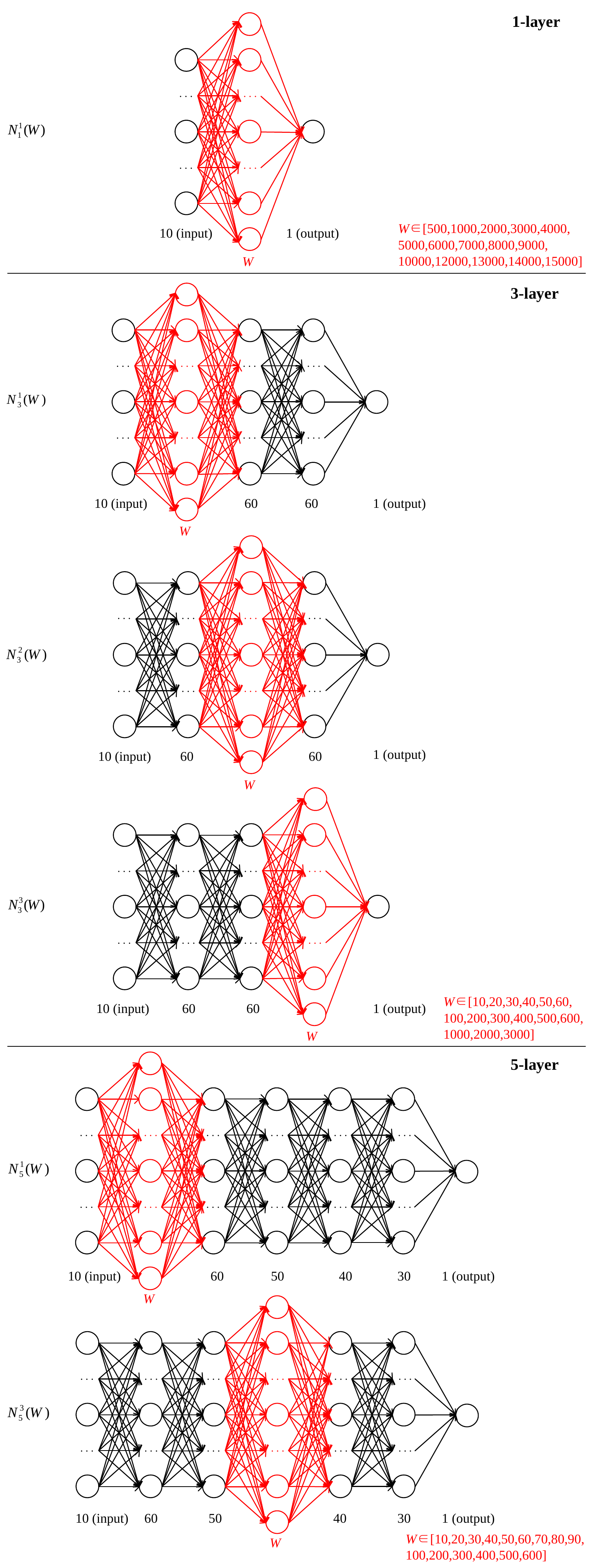}
    \caption{Network architectures of various depths and widths.}
    \label{fig:nn_design}
\end{figure}

The settings of simulations are described as follows.

{\bf Implementation and Evaluation}:  The are  five purposes in our experimental study. The first one is to verify the adaptivity of depths to the feature. The second one is to declare  the adaptivity of   features to the  depth. The third one aims at demonstrating the necessity of depth in feature extraction. The fourth one focuses on the necessity of depths in generalization. In our last experiment, we show the power of deep nets in some real applications.
In each simulation, we randomly generate  $m$ sample points $\{x_i\}_{i=1}^m$ on $\mathcal X\in\mathbb R^d$ according to the uniform distribution. Each $x_i$ corresponds to an output $y_i$ with either $y_i=f(x_i)$ (Sections 6.2, 6.3, 6.4) or $y_i=f(x_i)+\varepsilon_i$ (Section 6.5) with $\varepsilon_i$ {  some} Gaussian noise.
We repeat 10 times and record the average values of the following five quantities:

$\bullet$ Mean squared error (MSE): given an estimator $f_{D}$,   MSE, defined by $\frac1m\sum_{i=1}^m(f_D(x_i)-y_i)^2$,
measures the average squared difference between the estimated values and what is estimated.   It is a standard measurement to quantify the prediction performance of an estimator.

$\bullet$ Mean absolute error (MAE):   MAE,  defined by $\frac1m\sum_{i=1}^m|f_D(x_i)-y_i|$,  quantifies the fitting performance of $f_{D}$. It is another popular measurement, which is less sensitive to outliers than MSE, to quantify the prediction performance.

$\bullet$ Median absolute error (MdAE):  MdAE, defined by $m_{0.5}(\left| f_D(x_i)-m_{0.5}(y_i) \right|)$, is a robust measure of the variability of an estimator,
 where $m_{0.5}$ means a median. Thus, MdAE, together with MAE, shows the robustness of the estimator.

$\bullet$ R squared score (R$^2$S):   R$^2$S, defined by
$
   R^2S(y,f) = 1-\frac{{\sum_{i=1}^n(y_i-f_D(x_i))}^2}{{\sum_{i=1}^n(y_i-\bar{y})}^2}
$
with $\bar{y}=\frac1m\sum_{i=1}^m y_i$, is a statistical measurement that represents the proportion of the variance of an estimate by that of real outputs in a regression model.
It measures the fitness of the model.

$\bullet$ Explained variance score (EVS): EVS, defined by
$
1-\frac{{\sum_{i=1}^m(y_i-f_D(x_i))}^2}{\sum_{i=1}^m y_i^2},
$
measures the proportion to which a mathematical model accounts for the variation (dispersion) of a given data set.

All these measurements quantify the prediction performance of an estimator in terms of the prediction accuracy, sensitivity to outliers, robustness and fitness.

{\bf Structures of deep nets}: Generally speaking, there are four components in describing the structure of deep nets: depth, width in each layer,  sparsity in conjunction, and sharing weights. In our experiments, network width is equivalent to the number of neurons. If the sparsity in conjunction and  sharing weights are considered, there are too many structures even for a three-layer feed-forward network. Thus, we are only concerned with fully connected deep nets with different width and depth. In fact, we train over 200 networks of different depths and widths  in our simulations.  We use  $N_L^\ell(W)$ to represent  a network of $L$ layers and width $W$ in the $\ell$-th layer (marked as red as in Figure \ref{fig:nn_design}). For example, $N_3^2(100)$ and $N_3^2(200)$ are both 3-layer networks. The widths in layer-1 and layer-3 are fixed. The only difference is the widths in layer-2 are $100$ and $200$, respectively. In Figure \ref{fig:nn_design}, we present some examples for the structures adopted in our simulations. The details of structures will be explained in each simulation, if it is needed.

\subsection{Adaptivity of the Depth to features}

In this subsection, we study the  performance of deep nets in extracting the 10-dimensional ``square-feature'':
$$
 f(x)=\sum_{j=1}^{10}(x^{(j)})^2,
$$
where   $x=(x^{(1)},\dots,x^{(10)})$ is  i.i.d. generated according to the uniform distribution on  $[-100,100]^{10}$. The sizes of training dataset and testing dataset are $3000$ and $200$, respectively.
Our purpose is to show the adaptivity of   structures to the square feature, i.e., there are various   structures to extract the square feature.

\subsubsection{The necessity of depth}
\begin{figure}[h]
    \centering
    \includegraphics[width=0.7\linewidth]{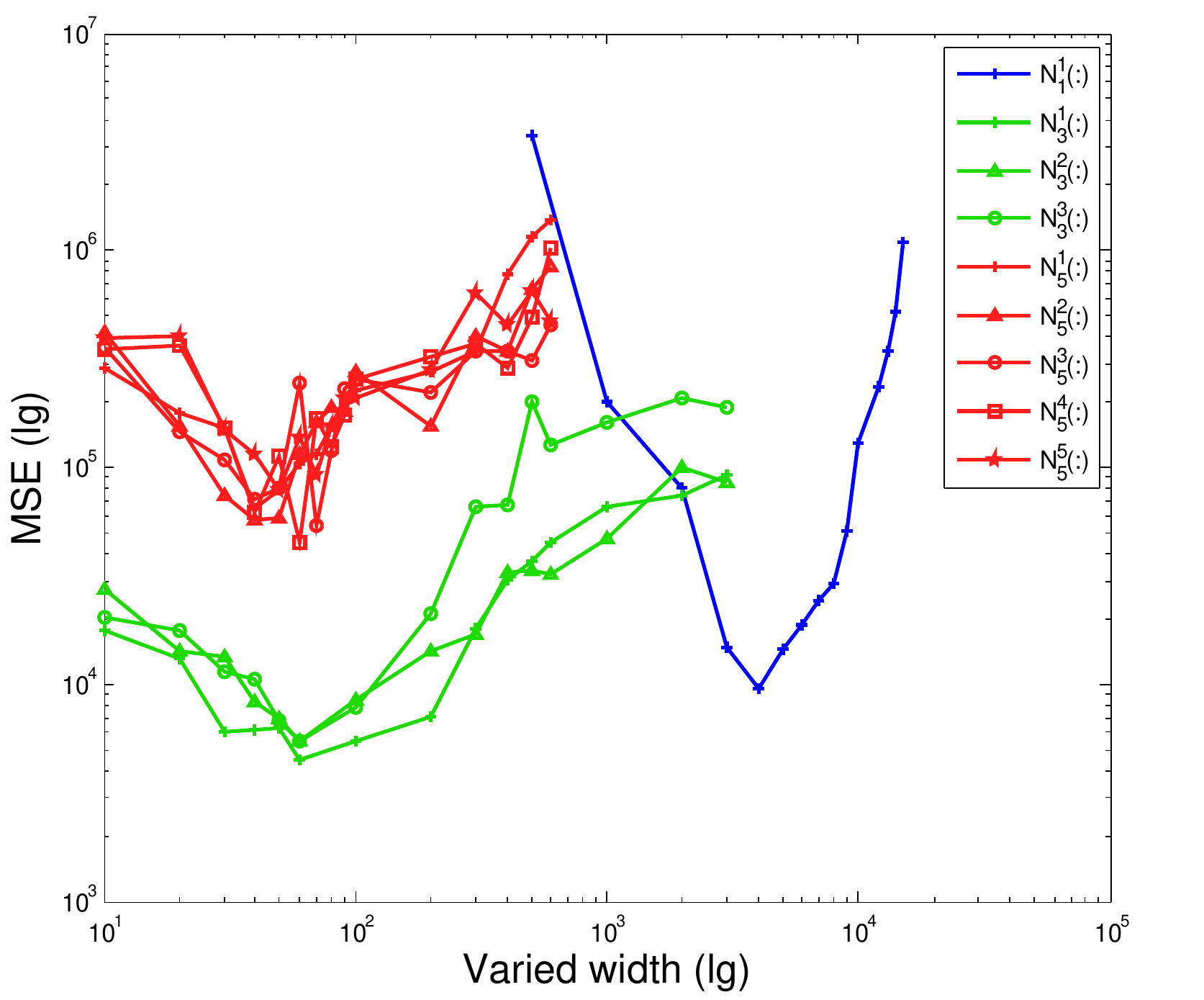}
    \caption{MSE curves of networks with various structures}
    \label{fig:d_and_w}
\end{figure}

For comparison, we train 135 networks of different depths and widths. The network architectures are illustrated in Figure \ref{fig:nn_design}. In particular, we  choose 3 different depths $L=\{1,3,5\}$ and select  15 different widths, which are
shown in different colors in Figure \ref{fig:d_and_w} and marked in different curves.

As shown in Figure \ref{fig:d_and_w}, all the curves show similar patterns, i.e., along with the increasement of width, the MSE decreases at the beginning and increases later. The difference is, for the deeper networks, it generally needs smaller width to reach the best performance. The average widths in the varied layers corresponding to the best performance networks of 1-layer, 3-layer and 5-layer are 4000, 60 and 52, respectively.

The outperformance of  3-layer over shallow nets in Figure \ref{fig:d_and_w} verifies the necessity of depth and show that deep nets can extract the square feature better than shallow nets with much fewer neurons. The superiority of  3-layer over 5-layer deep nets demonstrates that there exists an optimal depth in extracting some specific feature. Here the optimality means not only   the optimal accuracy, but also the solvability or convergence of the adopted AdamOptimizer algorithm, since its convergence issue   is questionable when the depth increases \cite{Goodfellow2016}. Thus, although Theorem \ref{Theorem:polynomial} proved that there are numerous depth-parameter pairs achieving the same accuracy, the convergence issue suggests to set the depth as small as possible.
%When the depth is fixed, depth 3 for example, Figure \ref{fig:d_and_w} says that there are numerous structures yielding similar accuracy.

\subsubsection{Role of the width for 3-layer deep nets}\label{Sec.Role-of-width}
Theorem \ref{Theorem:polynomial} presents a relation between  the depth and the number of free parameters in extracting the square feature. However, it does not give any guidance on the distribution of the width in each hidden layer.
In this experiment, we fix the total number of parameters of a 3-layer network at 8000 (slightly variation is allowed, and the range is $[8000, 8100]$). We manually change width of each layer, the numbers of parameters connecting Input and Layer-1 ($C_1$), connecting Layer-1 and Layer-2 ($C_2$), connecting Layer-2 and Layer-3 ($C_3$), and connecting Layer-3 and Output ($C_4$). Hence we generate a group of networks (20 in total) with different representative parameter distributions. The details of the networks are listed in Appendix E.

We  record the testing errors  in Figure \ref{fig:distributions}, where each spot represents one network and the coordinates $(p(C_1),p(C_2),p(C_3))$ are the percentage of the parameters occupied. As $p(C_1)+p(C_2)+p(C_3)+p(C_4)=1$, the 3-layer networks of various distributions can be uniquely positioned by this 3-dimensional coordinate system. The size of the spot represents the MSE of the corresponding network, i.e., smaller spot indicates smaller MSE. To be noted, the biggest MSE that can be reflected by the size of the spot is 10000. The networks with MSE larger than 10000 are represented by yellow spots, while the red spots mean the corresponding networks do not converge at all.

Figure \ref{fig:distributions} exhibits two phenomena for deep nets in feature extraction. The first one is the huge impact of the width distribution.
The pattern shown in this experiment is that more connections between Layer-1 and Layer-2 ($p(C_2)$) generally bring better results, while a large number of connections with Input or Output layers ($p(C_1)$ or $p(C_4)$) lead to  bad performances. This phenomenon indicates why a network is usually designed in a spindle shape. The other one is the adaptivity of the structure to the ``square-feature''. It can be found in Figure \ref{fig:distributions} that all green points perform similarly, which means that if the depth is suitable selected, then there is a large range of the width distributions such  that deep nets with such   distributions succeed in extracting the ``square-feature''.
\begin{figure}[h]
    \centering
    \includegraphics[width=0.7\linewidth]{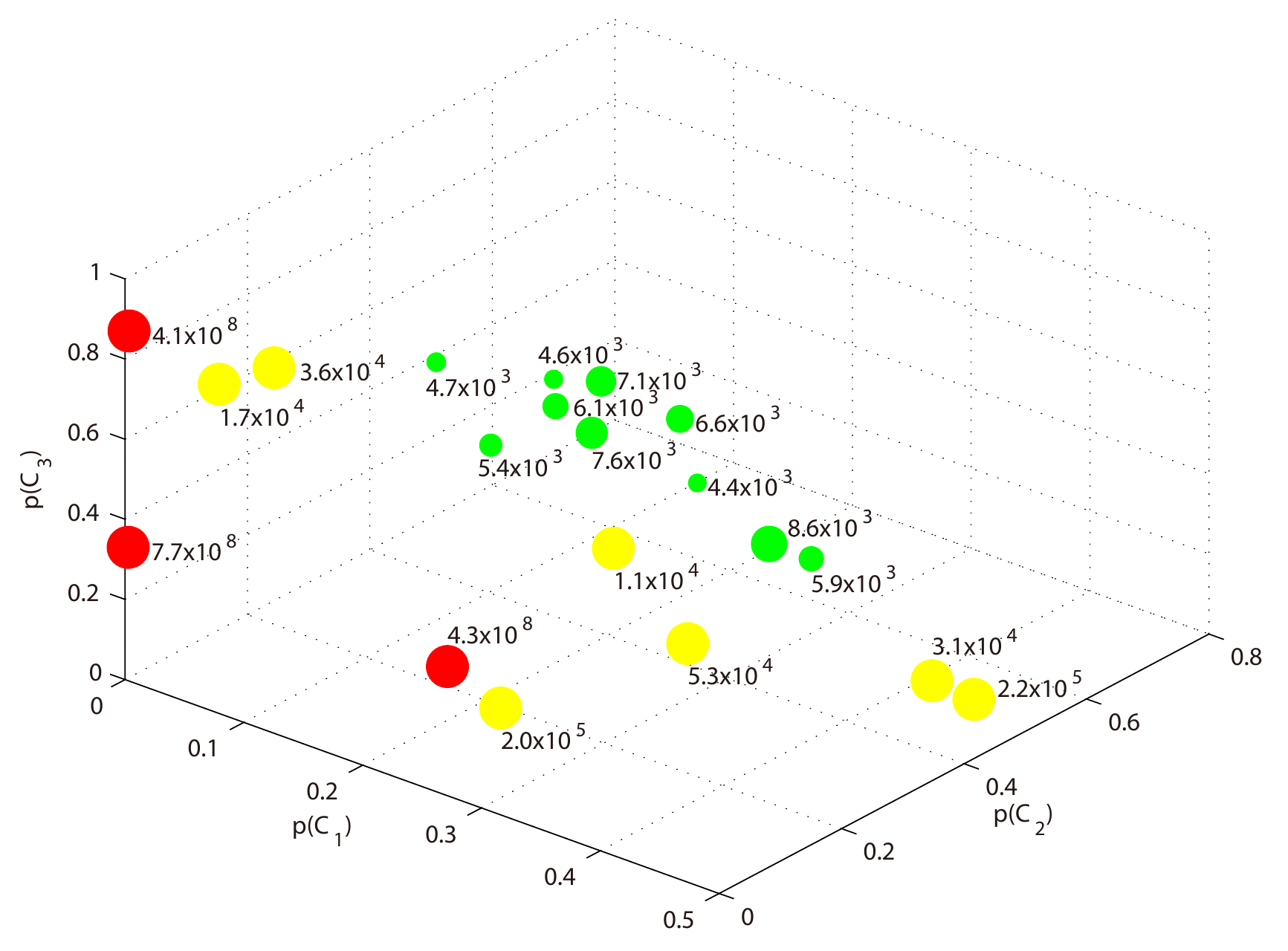}
    \caption{Networks with various width distributions.}
    \label{fig:distributions}
\end{figure}

\subsection{Adaptivity of  features to the depth}\label{Sec.Adaptivity-to-depth}

In Theorems \ref{Theorem:polynomial}-\ref{Theorem:jackson for trade-off}, it was proved that deep nets with fixed depth can extract different data features including the sparsity, group structure, and smoothness. In this simulation, we aim to verify this adaptivity of the feature to structures. We are interested in   partially radial features defined by
$$
 f_k(x)=\sum_{j=1}^{k}(x^{(j)})^2+\sum_{j=k+1}^{10}x^{(j)},
$$
where $x$ is generated by uniformly sampling from $[-100,100]^{10}$.

%\begin{table*}[hbt]
%\begin{center}
%\caption{Extracting partially radial feature with different groups.}
%\label{adaptivity}
%\begin{tabular}{c|c|c|c|c|c|c|c|c|c}
%\hline
%\multicolumn{2}{c|}{Evaluation metrics}&\multicolumn{1}{c|}{k=2}&\multicolumn{1}{c|}{k=3}&\multicolumn{1}{c|}{k=4}&\multicolumn{1}{c|}{k=5}&\multicolumn{1}{c|}{k=6}&\multicolumn{1}{c|}{k=7}&\multicolumn{1}{c|}{k=8}&\multicolumn{1}{c}{k=9}\\
%
%  \hline
%  \multirow{2}{*} {MAE}
%  &{1-layer}&$24.616$&$26.027$&$30.948$&$30.957$&$38.471$&$47.654$&$48.89$&$66.579$\\
%  &{3-layer}&$15.775$&$23.827$&$28.888$&$29.002$&$35.628$&$46.716$&$47.914$&$56.534$\\
%  \hline
%  \multirow{2}{*} {MSE}
%  &{1-layer}&$1357.934$&$1885.427$&$2345.819$&$2851.989$&$4216.136$&$5231.759$&$6304.671$&$9154.373$\\
%  &{3-layer}&$777.817$&$1682.958$&$2125.355$&$2555.937$&$3674.015$&$4884.519$&$6265.782$&$7098.902$\\
%  \hline
%  \multirow{2}{*} {MdAE}
%  &{1-layer}&$14.897$&$16.511$&$18.476$&$24.209$&$25.818$&$30.385$&$30.42$&$35.619$\\
%  &{3-layer}&$8.488$&$13.609$&$17.149$&$20.01$&$24.123$&$28.585$&$29.569$&$34.299$\\
%  \hline
%  \multirow{2}{*} {R$^2$S}
%  &{1-layer}&$1.0$&$1.0$&$1.0$&$1.0$&$1.0$&$1.0$&$1.0$&$1.0$\\
%  &{3-layer}&$1.0$&$1.0$&$1.0$&$1.0$&$1.0$&$1.0$&$1.0$&$1.0$\\
%  \hline
%  \multirow{2}{*} {EVS}
%  &{1-layer}&$1.0$&$1.0$&$1.0$&$1.0$&$1.0$&$1.0$&$1.0$&$1.0$\\
%  &{3-layer}&$1.0$&$1.0$&$1.0$&$1.0$&$1.0$&$1.0$&$1.0$&$1.0$\\
%  \hline
%\end{tabular}
%\end{center}
%\end{table*}

\begin{figure}[h]
    \centering
    \includegraphics[width=0.7\linewidth]{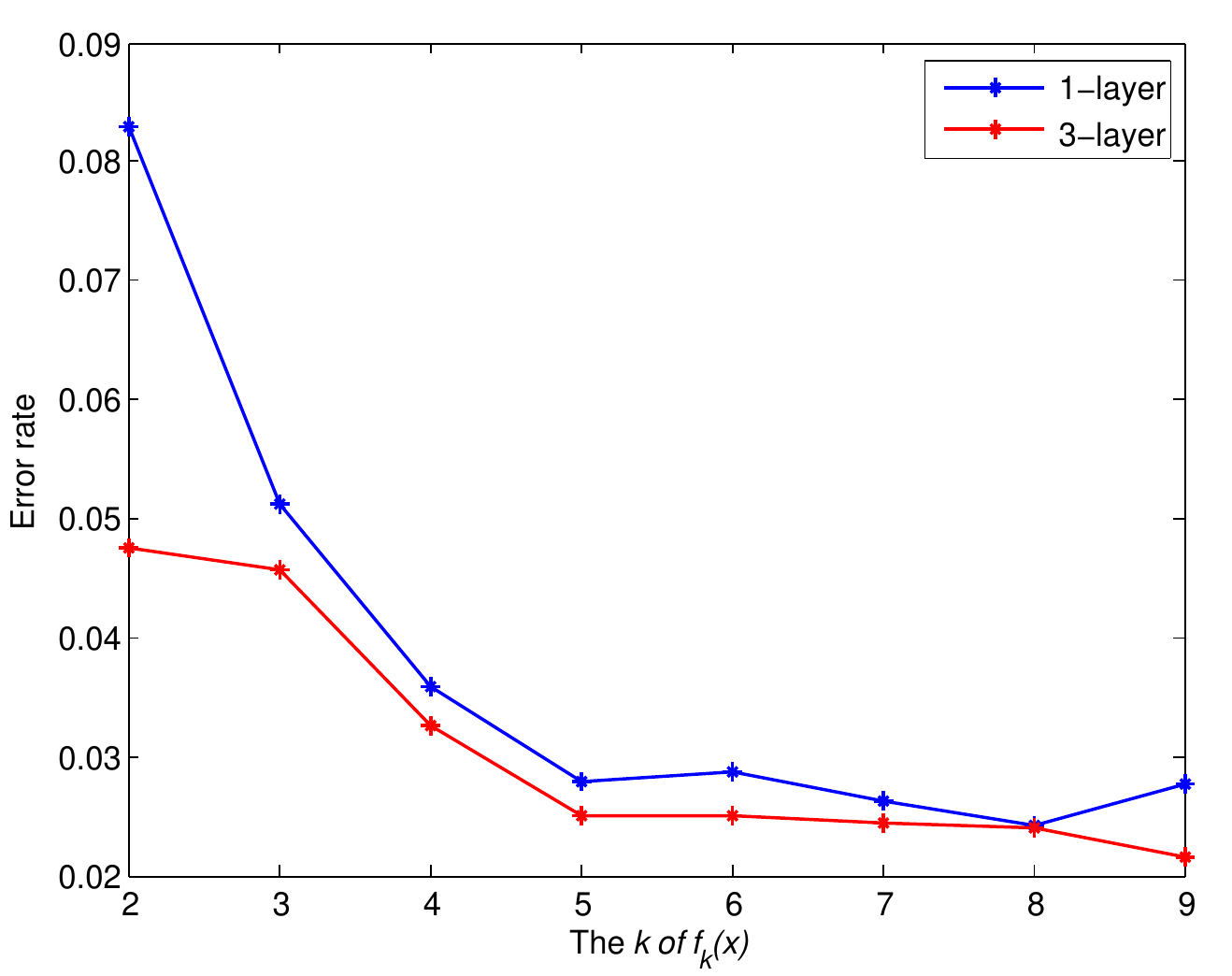}
    \caption{Adaptivity of the feature to structures.}
    \label{fig:adaptivity}
\end{figure}

%\begin{table*}[htb]
%\begin{center}
%\caption{Extracting the square-feature of different network depths.}\label{depth}
%\begin{tabular}{c c c c c c c}
%\hline
%  {Depth}&{1-layer}&{3-layer}&{5-layer}&{7-layer}&{9-layer}&{11-layer}\\
%  \hline
%  % after \\: \hline or \cline{col1-col2} \cline{col3-col4} ...
%  {MAE}&$48.989$&$44.006$&$145.568$&$176.055$&$427.975$&$1204.083$\\
%  {MSE}&$14837.713$&$4999.949$&$49579.938$&$86620.738$&$273505.566$&$3053474.62$\\
%  {MdAE}&$15.46$&$8.624$&$87.641$&$92.42$&$404.614$&$836.368$\\
%  {R$^2$S}&$1.0$&$1.0$&$1.0$&$1.0$&$1.0$&$1.0$\\
%  {EVS}&$1.0$&$1.0$&$1.0$&$1.0$&$1.0$&$1.0$\\
%  \hline
%\end{tabular}
%\end{center}
%\end{table*}

In the experiment, besides verifying the adaptivity of deep nets, we also compare the performances between deep nets and shallow nets to show the necessity of depth in extracting different data features.  As $k$ varies from $2$ to $9$, the structure of  deep nets (3-layer net) is fixed as $50-60-60$, while the widths in shallow nets is selected according to the test data directly to optimize their performance.

Figure \ref{fig:adaptivity} shows the result curves (the detailed numerical results can be found in Appendix F). It can be seen that a deep net with fixed structures performs robustly for dealing with different data features, and  always outperforms shallow nets. This demonstrates adaptivity of features to structures. Additionally, we are also aware  during the experiment that training shallow nets requires  more iterations.
 %Under the same training configuration, the training times of shallow network are about twice as many as that of deep net before convergence.

\begin{figure}[h]
    \centering
    \includegraphics[width=0.7\linewidth]{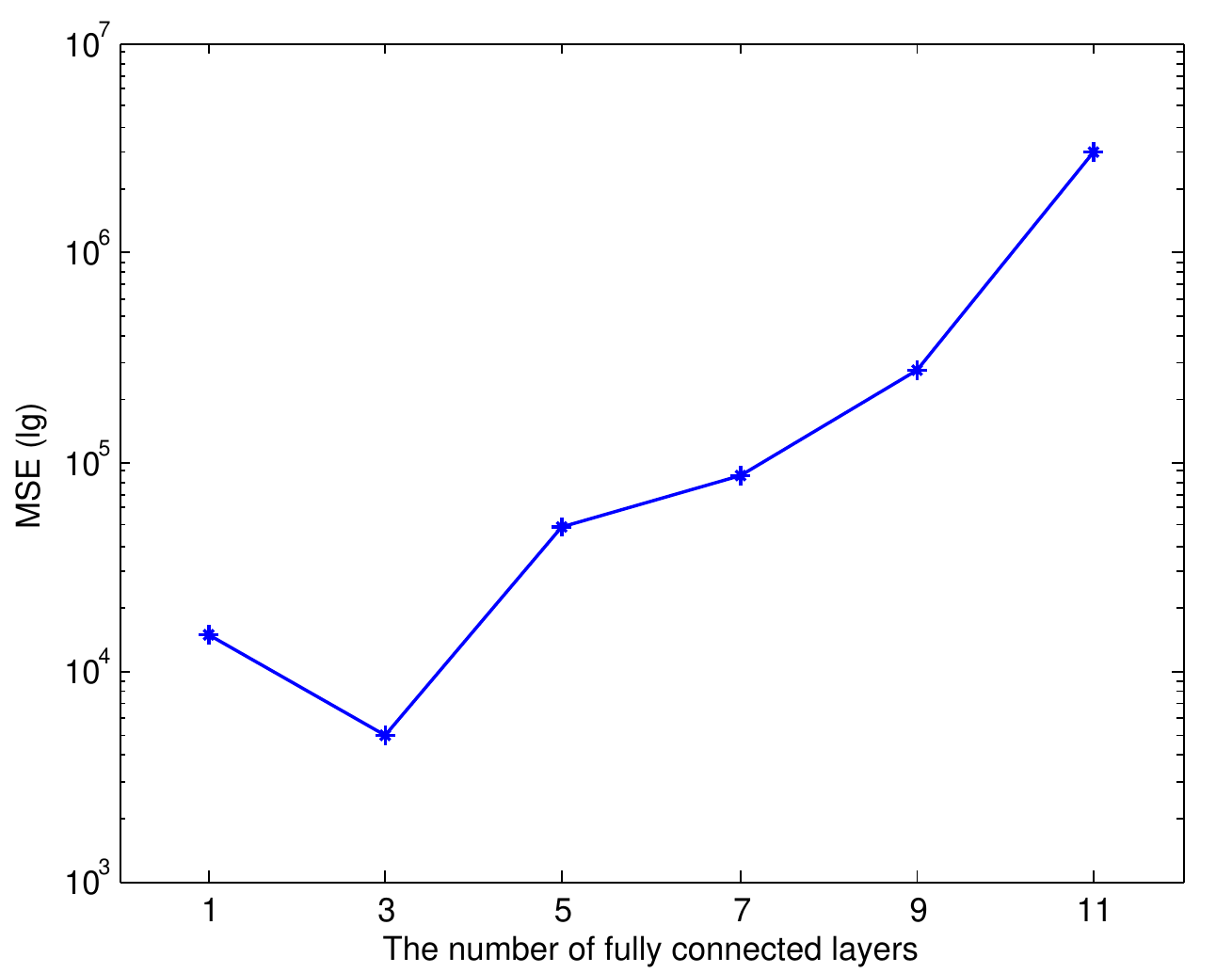}
    \caption{Best results from networks of different depths.}
    \label{fig:depth}
\end{figure}

\subsection{The role of depth in network} \label{Sec.Role-of-depth}
In this subsection, we study the role of depth in extracting the  10-dimensional ``square-feature''. The simulation setting is the same as that in Subsection 6.2.  The only difference is that we select more deep nets with different structures to perceive the impact of depth. There are six candidates for the depth, $1$, $3$, $5$, $7$, $9$ and $11$ and the width is chosen according to the test data directly from  much more candidates than those in Subsection 6.2. In particular,  the  number of neurons  in the widest layers of are $4000$, $60$, $30$, $30$, $9$ and $6$, respectively. The MES curve of simulation results is shown in  Figure \ref{fig:depth} (the detailed numerical results can be found in Appendix G).

It is shown that the depth plays a crucial role in improving the performance of neural networks in feature extraction.   We can see that a deep net performs better than a shallow net,  but a larger depth does not necessarily lead to better performance. For this simple case (single feature), deep nets with 3 layers are enough. Besides the   MSE curve in  Figure \ref{fig:depth}, Table \ref{depth} presented MSE, MAE, MdAE, R$^2$S and EVS results for the same purpose. All these measurements exhibit similar patterns to that of MSE in   Figure \ref{fig:depth} and verify both the necessity of depth in feature extraction and limitations of   deep nets with too many  hidden layers.

\subsection{Generalization capability verification}\label{Sec.Genralizaiton}
In this experiment, in order to test the generalization ability of networks, we train networks with noisy data in a more complex relationship.
The underlying relationship between the input signal $x=(x_1,x_2)$ and output is:
$$
      y=\sin\|x\|_2^2/\|x\|_2^2+\varepsilon,
$$
where $\varepsilon\sim\mathcal N(0,\sigma^2)$ is the Gaussian noise with the variance of $\sigma^2$. The training and test points are  generated by i.i.d. sampling 2000 and 200 points on $[-1,1]$ according to the uniform distribution, respectively, and the noise level is set $\sigma^2=0.1$. %The PSNR reaches $12.1225$ dB.

\begin{table}[htb]
\begin{center}
\caption{Network width candidates.}
\label{width_cand}
\begin{tabular}{c c c}
\hline
  {Depth}&{Range of width}&{Step length}\\
  \hline
  $1$&$[16,192]$&$16$\\
  $2$&$[4,32]$&$4$\\
  $3$&$[4,16]$&$4$\\
  $4$&$[4,16]$&$4$\\
  $5$&$[4,16]$&$4$\\
  $6$&$[4,16]$&$4$\\
  \hline
\end{tabular}
\end{center}
\end{table}

%\begin{table}[htb]
%\begin{center}
%\caption{Network width candidates.}
%\label{width_cand}
%\begin{tabular}{c c c c c c c}
%\hline
%  {Depth}&{1}&{2}&{3}&{4}&{5}&{6}\\
%  $Range of width$&$[16,192]$&$[4,32]$&$[4,16]$&$[4,16]$&$[4,16]$&$[4,16]$\\
%  $Step length$&16&$4$&$4$&$4$&$4$&$4$\\
%  \hline
%\end{tabular}
%\end{center}
%\end{table}

We compare the optimal MSE of deep nets of different depths. The optimal results are obtained by tuning two important parameters, the descent step (learning rate during network training) and the width  of each layer.
During the training process, the descent step changes dynamically as follow,
$$
R_d = R_0*D^{\lfloor(S_g/S_d)\rfloor},
$$
where $\lfloor\cdot\rfloor$ is the floor function, $R_0$ is the initial descent step and $R_d$ is the decayed descent step, $D$ is the decay rate, $S_g$ and $S_d$ are global step and decay step, respectively. Global step represents the current iteration number. Decay step controls the change frequency of descent step. For example, in this experiment, the decay step is   $1000$, and $D$ is $0.95$. The descent step decays to $95\%$ every 1000 iterations.
For choosing adequate $R_0$, we tried values of $0.0001$, $0.0005$ and $0.001$ on various networks of different depths. Empirically, we notice that a deeper network needs a smaller descent step. Therefore, in the experiment, the descent step is  $0.001$ for $1$, and $0.0005$ for   $2$, $3$, $4$, $5$ and $6$-layer networks.

The optimal widths of networks are chosen from a group of candidates, which are set empirically. Table \ref{width_cand} shows the details. For example, for a 3-layer network, the width candidates of each layer are $\{4,8,12,16\}$. As a result, there are $4^3$ networks for testing. To alleviate the test burden, in the experiment, we first fix  widths of non-middle layers at the medians of the corresponding ranges and test the middle layer width with all the candidates to elect the optimal one, then we tune other layers one by one by testing the candidates around the optimal width of the middle layer.

\begin{figure}[h]
\begin{minipage}[b]{0.49\linewidth}
\centering
\includegraphics*[scale=0.22]{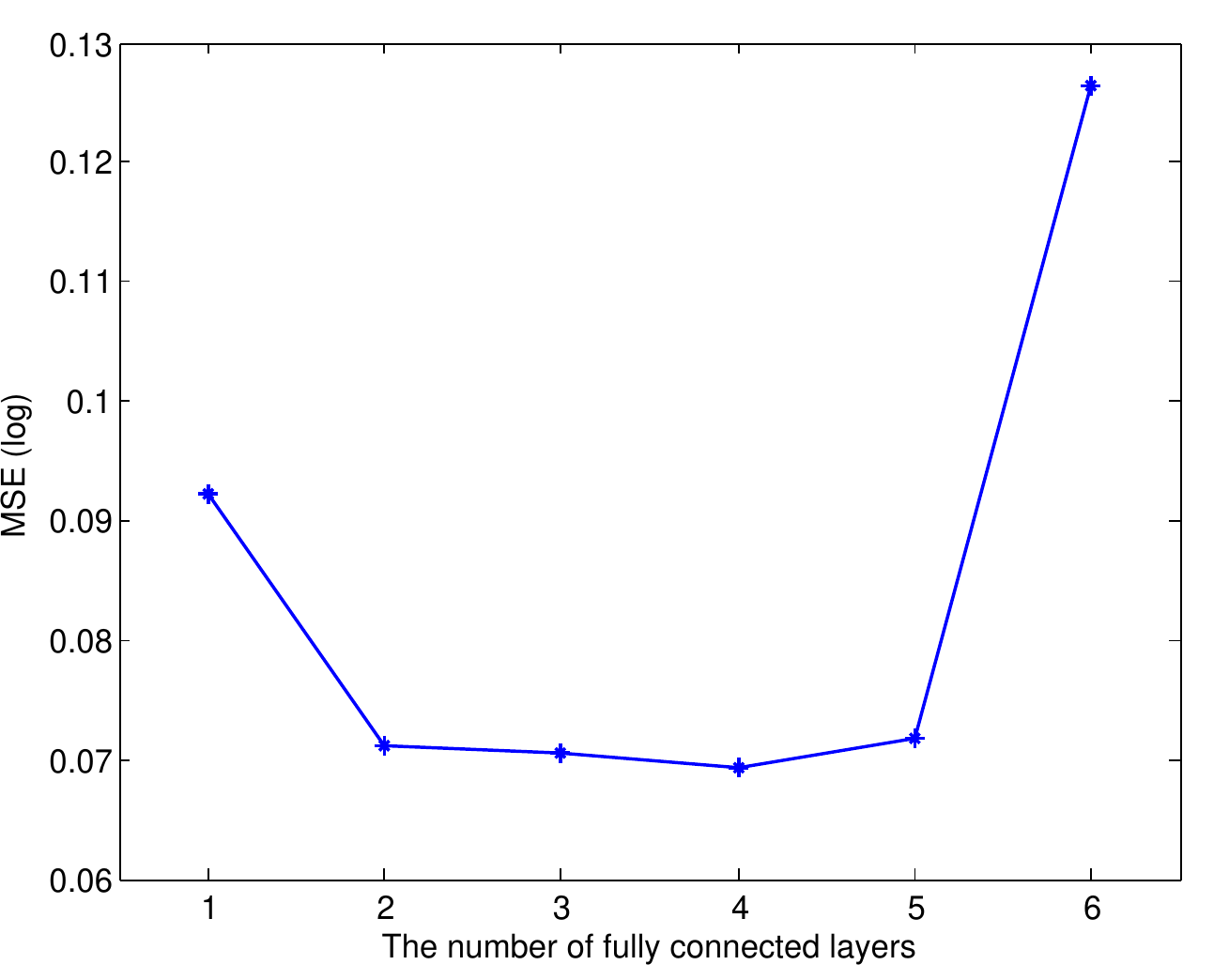}
\centerline{{\small (a) Accuracy and depth}}
\end{minipage}
\begin{minipage}[b]{0.49\linewidth}
\centering
\includegraphics*[scale=0.22]{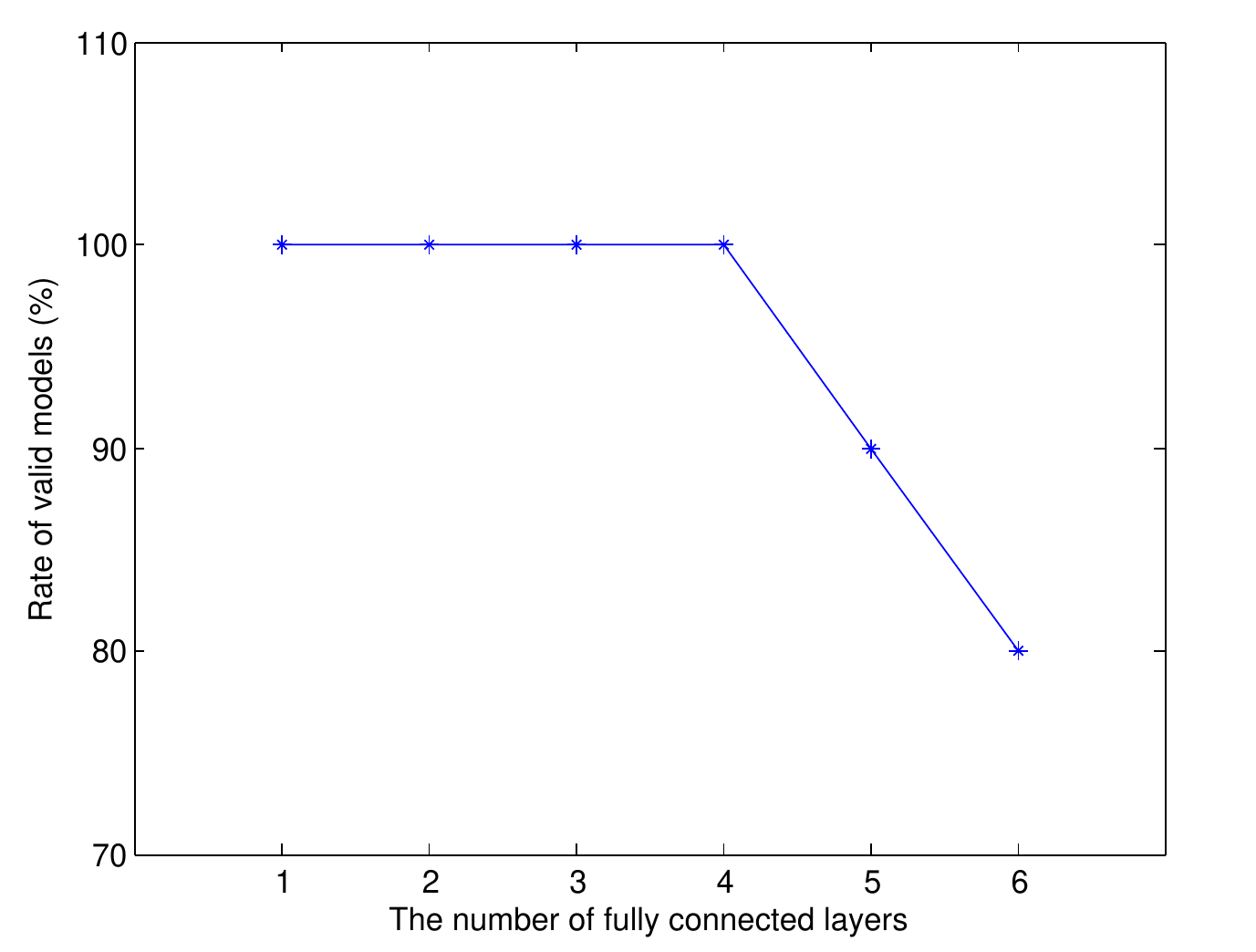}
%  \vspace{-.5cm}
\centerline{{\small (b)  {Valid model} and depth}}
\end{minipage}
\hfill
\caption{The generalization error result of deep nets.}
\label{fig:generalizaiton}
\end{figure}

In  Figure \ref{fig:generalizaiton}, we recorded the optimal MSE and the rate of  {valid model} of deep nets with different depths. Noting that the function $\sin\|x\|_2^2/\|x\|_2^2$ is smooth and radial, which are difficult for shallow nets to extract them simultaneously, according to the theoretical results in   \cite{Chui2018a}. In our simulation, we show that combining the feature extraction with target-driven learning in deep net is feasible. In fact, a deep net with four layers can significantly improve the performance of shallow nets.  Table \ref{genralizaiton} presents the regression result in terms of
MSE, MAE, MdAE, R$^2$S, and EVS, respectively and exhibits the same pattern as that of MSE in  Figure \ref{fig:generalizaiton}.

\begin{table*}[hbt]
\begin{center}
\caption{Noisy data training by networks of various depths.}\label{genralizaiton}
\begin{tabular}{c c c c c c c}
\hline
  {Depth}&{1-layer}&{2-layer}&{3-layer}&{4-layer}&{5-layer}&{6-layer}\\
  \hline
  {MAE}&$0.0765$&$0.0760$&$0.0741$&$0.0731$&$0.0755$&$0.0772$\\
  {MSE}&$0.0923$&$0.0712$&$0.0705$&$0.0694$&$0.0718$&$0.1265$\\
  {MdAE}&$0.0755$&$0.0745$&$0.0743$&$0.0708$&$0.0731$&$0.0770$\\
  {R$^2$S}&$-1.2481$&$-1.2040$&$-0.7588$&$-0.8710$&$-1.6787$&${-3.9073\times10^{11}}$\\
  {EVS}&$0.5909$&$0.5925$&$0.5943$&$0.5957$&$0.0674$&${-2.3854\times10^{11}}$\\
  \hline
\end{tabular}
\end{center}
\end{table*}

\begin{table*}[htb]
\begin{center}
\caption{The experiment results on synthetic seismic intensity dataset.}\label{syn_res}
\begin{tabular}{c c c c c c c}
\hline
  {Depth}&{1-layer}&{2-layer}&{3-layer}&{4-layer}&{5-layer}&{6-layer}\\
  \hline
  {MAE}&$0.176$&$0.125$&$0.054$&$0.036$&$0.061$&$0.25$\\
  {MSE}&$0.0578$&$0.0312$&$0.0049$&$0.0037$&$0.0074$&$0.1684$\\
  {MdAE}&$0.151$&$0.091$&$0.051$&$0.049$&$0.064$&$0.16$\\
  {R$^2$S}&$0.939$&$0.97$&$0.995$&$1.0$&$0.99$&$0.859$\\
  {EVS}&$0.939$&$0.97$&$0.995$&$1.0$&$0.99$&$0.859$\\
  \hline
\end{tabular}
\end{center}
\end{table*}

\subsection{Applications for the earthquake seismic intensity prediction}
For verifying our theoretical assertions on real applications, in this subsection, we do experiments on earthquake seismic intensity estimations.
Earthquake early warning (EEW) systems serve as the tools for coseismic risk reduction. One of the challenges in the development of EEW systems is the accuracy of seismic intensity estimation at the largest possible warning time. Seismic intensity is the intensity or severity of ground shaking at a given location. The level of seismic intensity depends heavily on the distance  between the observation site and the epicenter. It can be realized that the level of seismic intensity is a radial function by taking the epicenter as the origin.
In the experiments, we test on synthetic data and then deal with a real world dataset.

\subsubsection{Synthetic data experiment}
%Shaking intensity produced by an earthquake is localized, generally diminishing with distance from the earthquake's epicenter. The MMI is generally divided into 13 degrees. The lower degrees generally deal with the manner in which the earthquake is felt by people. The higher scales are based on observed structural damage[1]. It measures the effects of an earthquake at a given location.

\begin{figure}[h]
    \centering
    \includegraphics[width=0.7\linewidth]{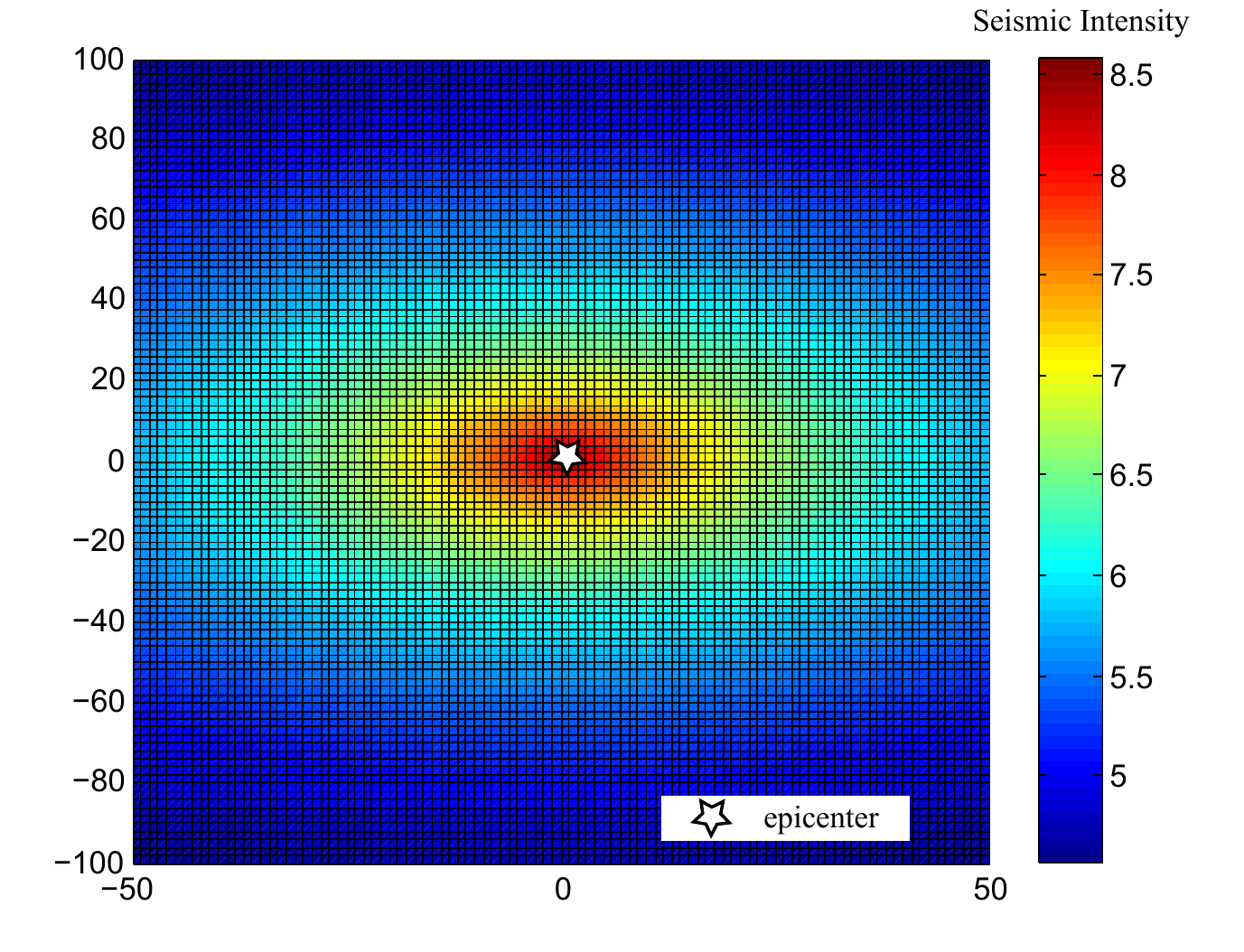}
    \caption{Dense synthetic MMI map generated by (\ref{MMI}).}
    \label{fig:syn_MMI}
\end{figure}

The Modified Mercalli intensity scale (MM or MMI), descended from Giuseppe Mercalli's Mercalli intensity scale of 1902, is the most used seismic intensity scale for measuring the intensity of shaking at a given location. It has been a common sense that seismic intensity is an expression of the amplitude, duration and frequency of ground motion. Thus, many attempts have been made to estimate MMI with the ground motion parameters \cite{Sokolov1998On}. Fourier amplitude spectrum (FAS) is one of the best features meeting the requirement based on which \cite{sokolov2002seismic} gives an estimation of MMI as
\begin{gather}\label{MMI}
\begin{aligned}
    &MMI=
     \exp\{1.2655+0.2089\mathcal M\\
     &-0.0011d-0.2451\log(d+2.1502\mathcal M)\},
\end{aligned}
\end{gather}
where $d$ is the estimated Joyner-Boore distance (in kilometers), and $\mathcal M$ is the moment magnitude. Figure \ref{fig:syn_MMI} shows a dense synthetic MMI map generated by (\ref{MMI}).

\begin{figure}[h]
    \centering
    \includegraphics[width=0.7\linewidth]{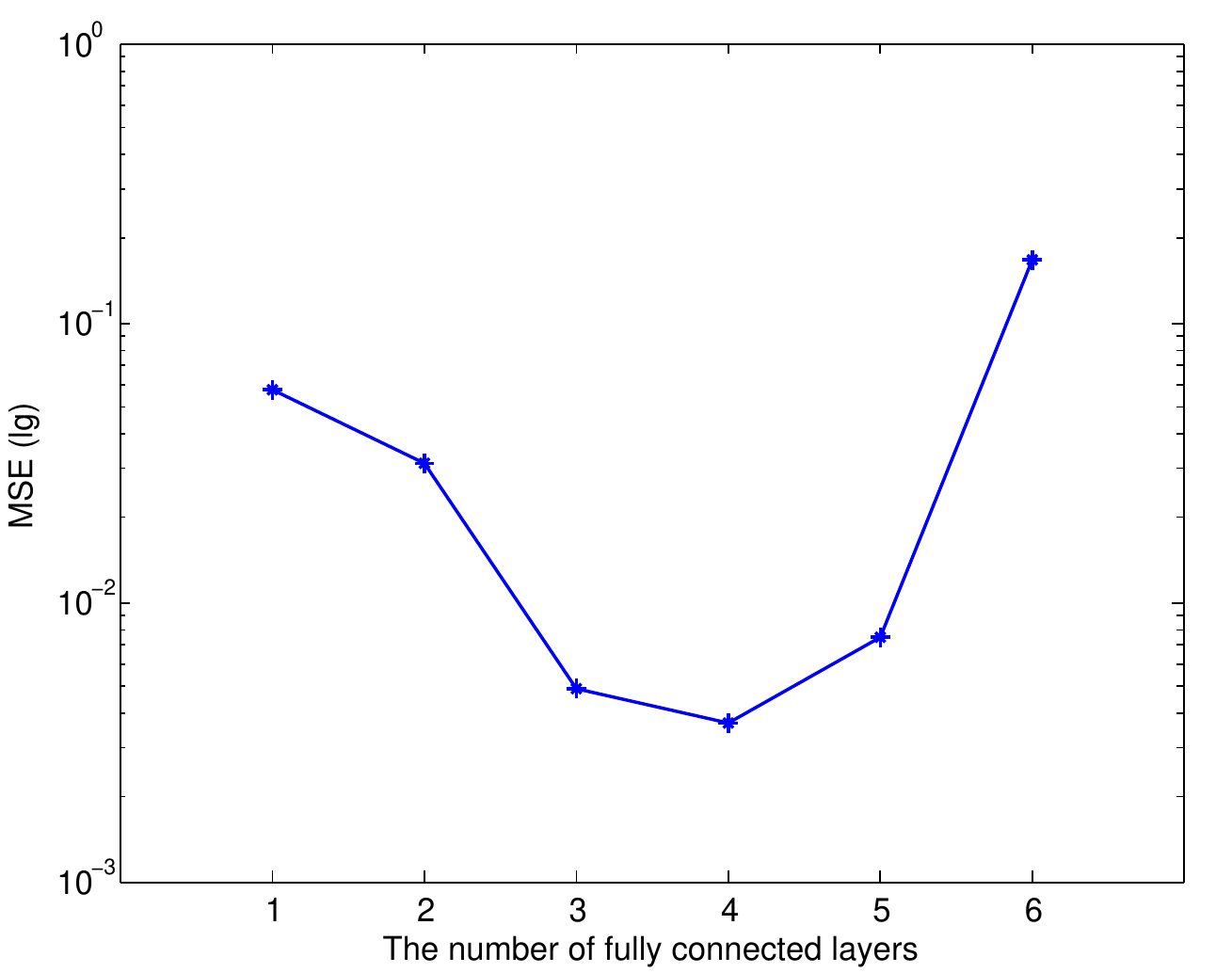}
    \caption{MSE curve of network during training on synthesis synthetic intensity dataset.}
    \label{fig:syn_res}
\end{figure}

For network testing, we generate 900 samples according to (\ref{MMI}) for training networks of 6 different depths. The testing results are reported in Figure \ref{fig:syn_res} and Table \ref{syn_res}. Similar to the  experiment in Subsection \ref{Sec.Genralizaiton}, networks perform well and the best result is also given by a 4-layer network.

\subsubsection{Real data experiment}
\begin{figure}[h]
    \centering
    \includegraphics[width=0.7\linewidth]{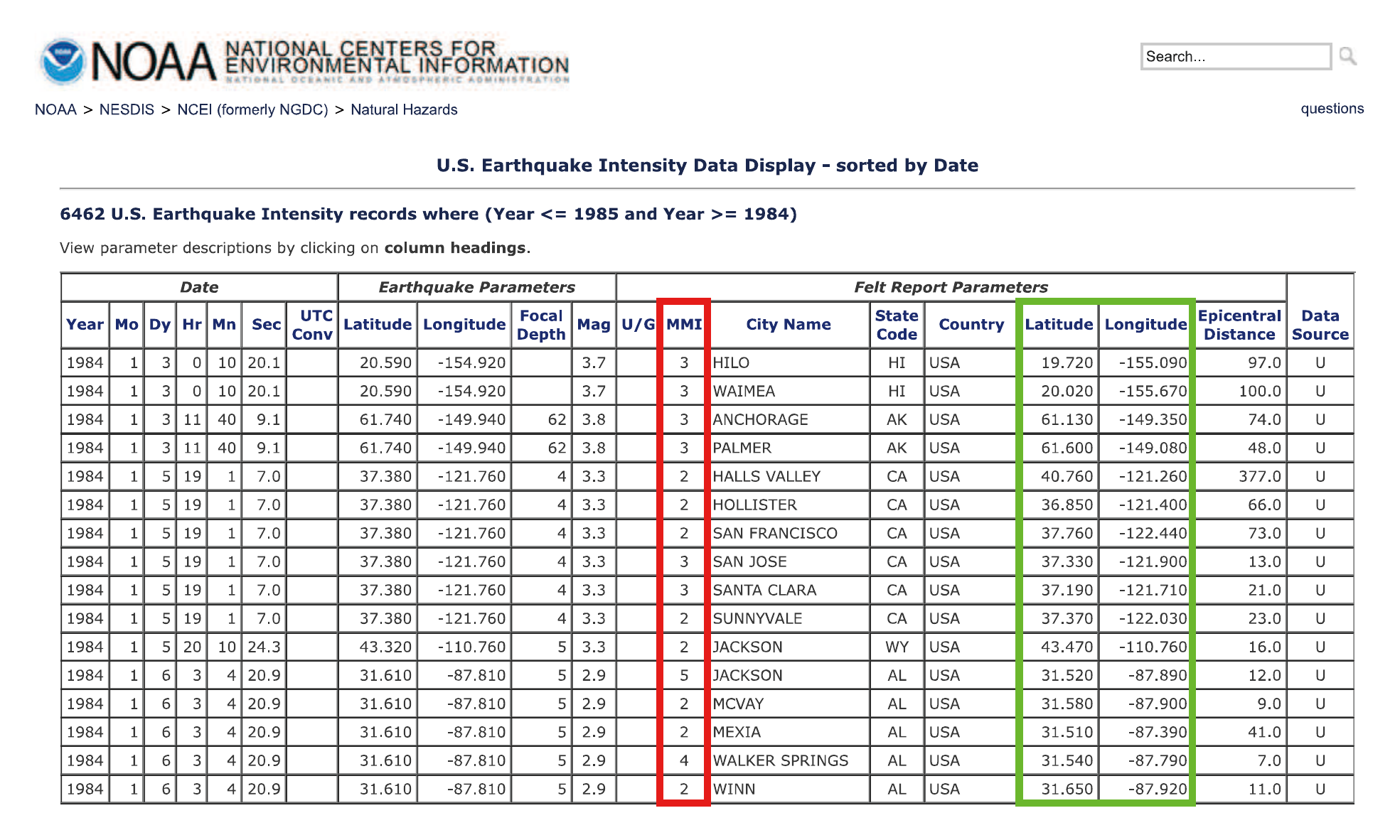}
    \caption{U.S. Earthquake Intensity Data }
    \label{fig:US_Eq}
\end{figure}

For the real data experiment, data are from the U.S. Earthquake Intensity Database\footnote{ https://www.ngdc.noaa.gov/seg/hazard /earthqk.shtml}, which collects damage and felt reports for over 23,000 U.S. earthquakes. The digital database contains information regarding epicentral coordinates, magnitudes, focal depths, names and coordinates of reporting cities (or localities), reported intensities, and the distance from the city (or locality) to the epicenter. Some samples of the data are shown in Figure \ref{fig:US_Eq}. The input of networks in this experiment are the latitude and longitude coordinates of the site where the earthquake occurred (green box), and the output is seismic intensity (red box). As the seismic intensity values in the database are integers in $\{2,\dots,6\}$ , we consider this task as a classification problem rather than a regression problem.

In order to have enough data for network training, we collect the most data impacted by the same epicenter from the dataset as the experiment data. There are total $608$ samples, in which, $500$ are used for training and the rest $108$ for testing. The parameter tuning strategy is similar to that of Section \ref{Sec.Genralizaiton}. Table \ref{US_Eq_Res} shows the comparison results between deep nets of different depths and traditional classification methods, i.e., support vector machine (SVM) and random forest (RF). It is  shown that a 5-layer deep net  gives the best performance, while   networks with other depths cannot compete with SVM.

\begin{table}[htb]
\begin{center}
\caption{Comparisons with traditional methods.}\label{US_Eq_Res}
\begin{tabular}{c|c|c}
  \hline
  \multicolumn{2}{c|}{Method}&{Recognition rate}\\
  \hline
  \multicolumn{2}{c|}{SVM}&{$62.96\%$}\\
  \multicolumn{2}{c|}{Random forest}&{$58.33\%$}\\
  \hline
  \multirow{4}{*}{Deep networks}
  &{1-layer}&{$57.41\%$}\\
  &{3-layer}&{$60.1\%$}\\
  &{5-layer}&{$\textbf{66.67\%}$}\\
  &{7-layer}&{$62.03\%$}\\
  \hline
\end{tabular}
\end{center}
\end{table}

\section{Conclusion}
In this paper, we  studied theoretical advantages of deep nets via considering the role of depth    in feature extraction and generalization. The main contributions are four folds. Firstly, under the same capacity costs (via covering numbers), we proved that deep nets are better than shallow nets in extracting the group structure features. Secondly, we proved that deep ReLU nets are one of the optimal tools in extracting the smoothness feature. Thirdly, we rigorously proved  the adaptivity of features to depths and vice verse, which was adopted to derive the optimal learning rate for implementing empirical risk minimization on deep nets. Finally, we conducted extensive numerical  experiments including   toy simulations and real data verifications to show the outperformance of deep nets in feature extraction and generalization. All these results presented reasonable explanations for the success of deep learning and provided solid guidance on using deep nets. {  In this paper, we only consider the depth selection in regression problems. It would be interesting and important to  develop similar conclusions for   classification. We will consider this topic and report   progress in our future study later.}

\section*{Acknowledgments}
The  research of Z. Han and S. Yu was partially supported by the National Natural Science
Foundation of China [Grant Nos. 61773367, 61821005], the Youth Innovation Promotion Association of the Chinese Academy of Sciences [Grant 2016183]. The
research of S.B. Lin was supported by the National Natural Science
Foundation of China [Grant No. 61876133], and the research of D.X. Zhou
was partially supported by the Research Grant Council of Hong Kong [Project No. CityU 11306617].

\newpage

\section*{Appendix A: Proofs of Theorem \ref{Theorem:polynomial}}

In Appendix A, we prove Lemma \ref{lemma:product gate 2} and Theorem
\ref{Theorem:polynomial}. Our main tool is  the following lemma,
which can be  found in \cite[Lemma A.3]{Petersen2017}.

\begin{lemma}\label{lemma:product gate}
Let $\theta>0$  and $\tilde{L}\in\mathbb N$ with $\tilde{L}>(2\theta)^{-1}$. For any  $\varepsilon\in(0,1)$, there exists a deep
ReLU net $\tilde{\times}_{2,\tilde{L},\varepsilon}$ with $2\tilde{L}+8$
  layers and at most $c\varepsilon^{-\theta}$ nonzero parameters
  which are bounded by $\varepsilon^{-\gamma}$
 such that
$$
       |uu'-\tilde{\times}_{2,\tilde{L},\varepsilon}(u,u')|\leq
       \varepsilon,\qquad\forall\ u,u'\in[-2,2],
$$
where $c,\gamma$ are   positive constants  depending only on $\tilde{L}$ and
$\theta$.
\end{lemma}
\begin{figure}[h]
    \centering
  \includegraphics[width=0.8\linewidth]{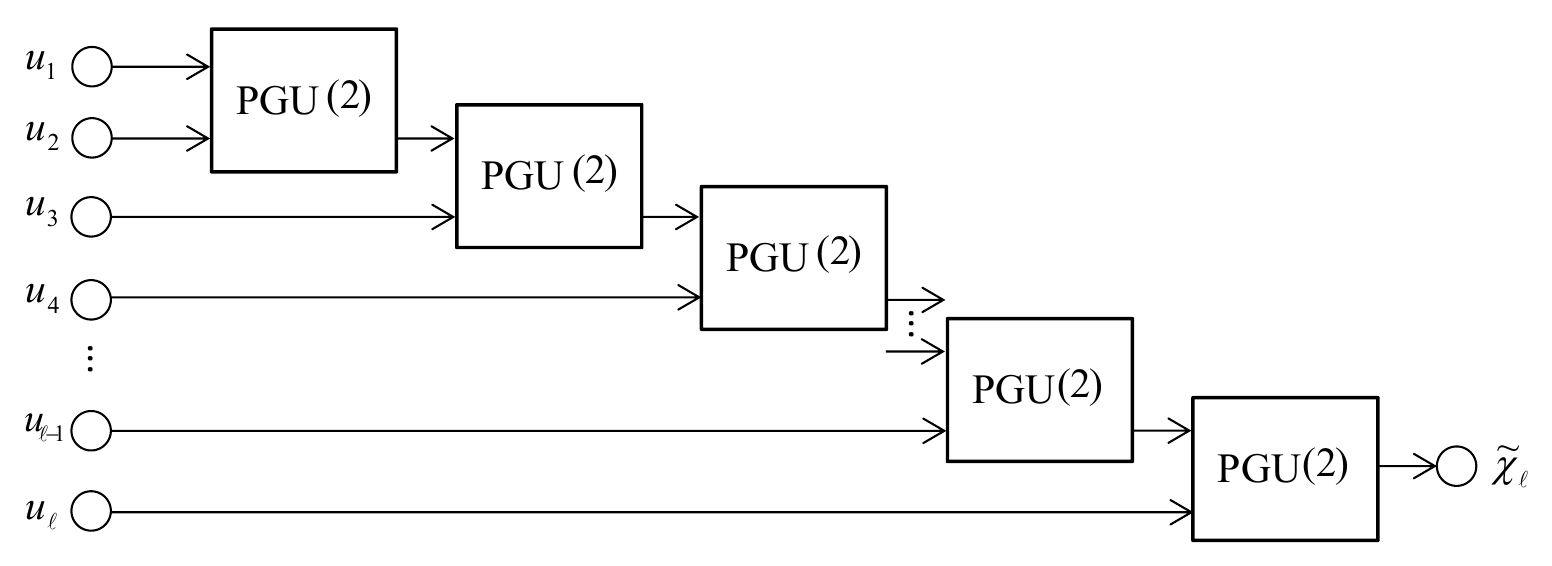}
    \caption{Network structure for PGU($\ell$)}
    \label{fig:PGU}
\end{figure}
Based on the above lemma, we can construct product gate units (PGUs) for multiple factors. In particular, there are numerous network structures for extending PGU with 2 factors to PGU with $\ell>2$ factors. We only prove that a network structure shown in Figure \ref{fig:PGU} is suitable for this purpose in  Lemma \ref{lemma:product gate 2}.

\begin{IEEEproof}[Proof of Lemma \ref{lemma:product gate 2}] For $j=2,\dots,\ell$ and $\varepsilon\in(0,1)$, define
\begin{eqnarray*}
     &&\tilde{\times}_{j,\tilde{L},\varepsilon/\ell}(u_1,\dots,u_j)\\
     &:= & \overbrace{\tilde{\times}_{2,\tilde{L},\varepsilon/\ell}(\tilde{\times}_{2,\tilde{L},\varepsilon/\ell}(
     \cdots\tilde{\times}_{2,\tilde{L},\varepsilon/\ell}}^{j-1}(u_1,u_2),\cdots,u_{u_{j-1}}),u_{j}).
\end{eqnarray*}
Then,
\begin{eqnarray*}
      && u_1 u_2\cdots u_\ell- \tilde{\times}_{\ell,\tilde{L},\varepsilon/\ell}(u_1,\dots,u_\ell)\\
      &=&
      u_1u_2\cdots u_\ell-\tilde{\times}_{2,\tilde{L},\varepsilon/\ell}(u_1,u_2)u_3\cdots
      u_\ell\\
      &+&
      \tilde{\times}_{2,\tilde{L},\varepsilon/\ell} (u_1,u_2)u_3\cdots
      u_\ell
      -
      \tilde{\times}_{3,\tilde{L},\varepsilon/\ell}(u_1,u_2,u_3)    u_4\cdots u_\ell\\
      &+&
      \dots\\
      &+&
       \tilde{\times}_{\ell-1,\tilde{L},\varepsilon/\ell}(u_1,\dots,u_{\ell-1})u_{\ell}
      -
      \tilde{\times}_{\ell,\tilde{L},\varepsilon/\ell}(u_1,\dots,u_{\ell}).
\end{eqnarray*}
But Lemma \ref{lemma:product gate} implies
$$
      |\tilde{\times}_{2,\tilde{L},\varepsilon/\ell}(u,u')|\leq |u| |u'|+\varepsilon/\ell \leq
      |u|+\varepsilon/\ell,\quad \forall\ u,u'\in[-1,1].
$$
This together with $0<\varepsilon<1$ and $u_i\in[-1,1]$ for
all $i=1,2,\dots,\ell$ yields
\begin{eqnarray*}
   && |\tilde{\times}_{j,\tilde{L},\varepsilon/\ell} (u_1,\dots,u_{j})|\\
   &\leq&
   |\tilde{\times}_{j-1,\tilde{L},\varepsilon/\ell}(u_1,\dots,u_{j-1})|+\varepsilon/\ell\\
   &\leq&\dots
   \leq
  | \tilde{\times}_{2,\tilde{L},\varepsilon/\ell}(u_1,u_2)|
   + (j-2)\varepsilon/\ell\\
   &\leq&
   1+(j-1)\varepsilon/\ell\leq 2,\qquad\forall\ j=2,\dots,\ell.
\end{eqnarray*}
Then, it follows from Lemma \ref{lemma:product gate} again that
\begin{eqnarray*}
     &&|u_1 u_2\cdots u_\ell-\tilde{\times}_{\ell,\tilde{L},\varepsilon/\ell}(u_1,\dots,u_\ell)|\\
     &\leq&
     |u_1u_2-\tilde{\times}_{2,\tilde{L},\varepsilon/\ell} (u_1,u_2)|\\
     &+&|\tilde{\times}_{2,\tilde{L},\varepsilon/\ell}(u_1,u_2)u_3
      -
      \tilde{\times}_{3,\tilde{L},\varepsilon/\ell}(u_1,\dots,u_3)|+\dots\\
      &+&
      |\tilde{\times}_{\ell-1,\tilde{L},\varepsilon/\ell}(u_1,\dots,u_{\ell-1})u_{\ell}
      -
      \tilde{\times}_{\ell,\tilde{L},\varepsilon/\ell}(u_1,\dots,u_{\ell})|\\
      &\leq&
      (\ell-1)\varepsilon/\ell<\varepsilon.
\end{eqnarray*}
Since  $\tilde{\times}_{2,\tilde{L},\varepsilon/\ell}$ is a deep net with $2\tilde{L}+8$
  layers and at most $c\ell^\theta\varepsilon^{-\theta}$ nonzero parameters  bounded $\ell^\gamma\varepsilon^{-\gamma}$ and the parameters in all PGU(2)  are the same,
 $ \tilde{\times}_{\ell,\tilde{L},\varepsilon/\ell}$ is a deep net with at most
  $(2\tilde{L}+8)\ell$ layers, and at most $c\ell^\theta\varepsilon^{-\theta}$
  non-zero parameters which are bounded by $\ell^\gamma\varepsilon^{-\gamma}$.
Setting $\tilde{\times}_\ell=\tilde{\times}_{\ell,\tilde{L},\varepsilon/\ell}$, we complete  the proof
of Lemma \ref{lemma:product gate 2}.
\end{IEEEproof}

We are now in a position to prove Theorem \ref{Theorem:polynomial}.
%
%\begin{figure}[h]
%    \centering
%    \includegraphics[width=0.7\linewidth]{Structure-Th1.pdf}
%    \caption{Structure for deep nets with product gate units (PGUs)}
%    \label{fig:Structure-th}
%\end{figure}

\begin{IEEEproof}[Proof of Theorem \ref{Theorem:polynomial}]
By Lemma \ref{lemma:product gate 2}, for any $\varepsilon\in(0,1),$ $\theta>0$ and $\tilde{L}$ with $\tilde{L}>(2\theta)^{-1}$
there exists a deep ReLU net $\tilde{\times}_\beta$ with   at most
  $(2\tilde{L}+8)\beta$ layers, and at most $c(\mu B\beta)^\theta\varepsilon^{-\theta}$
  non-zero parameters which are bounded by $(\mu B\beta)^\gamma\varepsilon^{-\gamma}$ such that
\begin{equation}\label{PGU-in-proof}
       |u_1u_2\cdots u_{\beta}-\tilde{\times}_\beta(u_1,\dots,u_\beta)|
       \leq\frac{\varepsilon}{\mu B},\quad \forall u_1,\dots,u_\beta\in[-1,1].
\end{equation}
Then, for any monomial
$$
      x^\alpha=\overbrace{x^{(1)} \cdots x^{(1)}}^{\alpha_1}
      \cdots\overbrace{x^{(d)} \cdots  x^{(d)}}^{\alpha_d}  \cdot \overbrace{1\cdots1}^{\beta-|\alpha|}
$$
  satisfying
$|\alpha|\leq \beta$, it follows from (\ref{PGU-in-proof}) that
$$
     |x^\alpha-\tilde{\times}_\beta(\overbrace{x^{(1)},\dots,x^{(1)}}^{\alpha_1},\dots,
     \overbrace{x^{(d)},\dots,x^{(d)}}^{\alpha_d},\overbrace{1,\dots,1}^{\beta-|\alpha|})|
       \leq\frac{\varepsilon}{\mu B}.
$$
Rewrite
   $P\in\mathcal P_{\beta,B,\mu}^d$ as
$P=\sum_{\alpha\in\Lambda_\mu}c_\alpha
x^\alpha$ with $|c_\alpha|\leq B$, where $\Lambda_\mu\subseteq\{\alpha:|\alpha|\leq\beta\}$ is a set of at most $\mu$ elements.  Define
\begin{equation}\label{a.111}
    h_P(x)=\sum_{\alpha\in\Lambda_\mu}
    c_\alpha\tilde{\times}_\beta(\overbrace{x^{(1)},\dots,x^{(1)}}^{\alpha_1},\dots,
     \overbrace{x^{(d)},\dots,x^{(d)}}^{\alpha_d},\overbrace{1,\dots,1}^{\beta-|\alpha|}).
\end{equation}
Then,
$$
      |h_P(x)-P(x)|\leq  \sum_{\alpha\in\Lambda_\mu}|c_\alpha|\frac{\varepsilon}{\mu B}\leq \varepsilon.
$$
Since the parameters in PGU($\beta$) are shared to be same, $\mu$ different $\alpha\in \Lambda_L$ activate  $\mu$ PGU($\beta$),   there are totally $\mu+c(\mu B\beta)^\theta\varepsilon^{-\theta}$ free parameters in $h_P$, which are distributed  in $(2\tilde{L}+8)\beta+1$ and bounded by $(\mu B\beta)^\gamma\varepsilon^{-\gamma}$. This completes  the
proof of Theorem \ref{Theorem:polynomial}.
\end{IEEEproof}

\section*{Appendix B: Proof of Theorem \ref{Theorem:jackson}}

 We
need a construction of deep nets  motivated by \cite{Yarotsky2017}.
For $t\in\mathbb R$, define
\begin{equation}\label{def.psi}
   \psi(t)=\sigma(t+2)-\sigma(t+1)-\sigma(t-1)+\sigma(t-2).
\end{equation}
Then,
\begin{equation}\label{property psi}
 \psi(t)=\left\{\begin{array}{cc} 1,&|t|\leq 1,\\
                                  0,&|t|\geq 2,\\
                                  2-|t|,&1<|t|<2.
                                  \end{array}\right.
\end{equation}
Furthermore, for $N\in\mathbb N$  and $\mathbf
j=(j_1,\dots,j_d)\in\{0,1,\dots,N\}^d,$ define
\begin{equation}\label{def.Phi}
    \phi_{{\bf
    j},N}(x)=\prod_{k=1}^d\psi\left(3N\left(x^{(k)}-\frac{j_k}N\right)\right).
\end{equation}
Direct computation yields
\begin{equation}\label{property Phi 1}
     \sum_{{\bf j}\in\{0,1,\dots,N\}^d}\phi_{{\bf j},N}(x)=1,\qquad\forall\ x\in\mathbb I^d
\end{equation}
and
\begin{equation}\label{property Phi 2}
     \mbox{supp}\phi_{{\bf j},N}\subseteq\{x:|x^{(k)}-j_k/N|<1/N,\forall k\},
\end{equation}
where  $\mbox{supp}f$ denotes the support of the function $f$. Before presenting the proof of Theorem \ref{Theorem:jackson}, we introduce two technical   lemmas.
The first one can be found in \cite[Lemma 1]{Kohler2014}.
\begin{lemma}\label{Lemma:Taylor polynomials}
Let $r=s+v$ with $s\in\mathbb N_0$ and $0<v\leq 1$. If $f\in
Lip^{(r,c_0)}$,  $x_0\in\mathbb R^d$  and $p_{s,x_0,f}$ is the
Taylor polynomial of $f$ with degree $s$ at  around $x_0$,
%i.e.,
%\begin{eqnarray}\label{Taylor polynomial}
%    p_{s,x_0,f}(x)&=&\sum_{k_1+\cdots+k_d\leq s}\frac{1}{k_1!\dots
%    k_d!}\frac{\partial^{k_1+\dots
%    k_d}f(x_0)}{\partial^{k_1}x^{(1)}\dots\partial^{k_d}x^{(d)}}\nonumber\\
%    &&(x^{(1)}-x_0^{(1)})^{k_1}
%    \cdots(x^{(d)}-x_0^{(d)})^{k_d},
%\end{eqnarray}
then
\begin{equation}\label{Taylor approximation}
   |f(x)-p_{s,x_0,f}(x)|\leq \tilde{c}_1\|x-x_0\|_2^r,\qquad\forall\ x\in\mathbb
   I^d,
\end{equation}
where $\tilde{c}_1$ is a constant depending only on $s$, $c_0$ and $d$.
\end{lemma}

The second one focuses on
approximating functions in $Lip^{(r,c_0)}$ by products of Taylor
polynomials  and ReLU nets.
\begin{lemma}\label{Lemma: app by product}
For   $f\in Lip^{(r,c_0)}$ with $r=s+v$, $N\in\mathbb N$ and ${\bf
j}/N=(j_1/N,\dots,j_d/N)$, define
\begin{equation}\label{first construction}
      f_1(x)=\sum_{\mathbf
      j\in\{0,1,\dots,N\}^d}\phi_{{\mathbf j},N}(x)p_{s,{\bf j}/N,f}(x),
\end{equation}
then
\begin{equation}\label{first app}
   |f(x)-f_1(x)|\leq \tilde{c}_2N^{-r},\qquad\forall x\in\mathbb I^d,
\end{equation}
where $\tilde{c}_2$ is a constant depending only on $r$,  $c_0$ and $d$.
\end{lemma}

\begin{IEEEproof} Due to (\ref{property Phi 1}),  (\ref{property Phi 2}), $0\leq\phi_{\bf j,N}(x)\leq 1$ and Lemma \ref{Lemma:Taylor
polynomials}, we have
\begin{eqnarray*}
      &&|f(x)-f_1(x)|\\
      &=&\left|f(x)-\sum_{\mathbf
      j\in\{0,\dots,N\}^d}\phi_{{\mathbf j},N}(x)p_{s,{\bf
      j}/N,f}(x)\right|\\
      &\leq&
      \sum_{\mathbf
      j\in \{0,\dots,N\}^d}\phi_{{\mathbf j},N}(x)|f(x)-p_{s,{\bf
      j}/N,f}(x)|\\
      &\leq&
      \sum_{\mathbf j: |x^{(k)}- j_k/N|<1/N,\forall k}|f(x)-p_{s,{\bf
      j}/N,f}(x)|\\
      &\leq&
      2^d\max_{\mathbf j: |x^{(k)}-\mathbf j_k/N|<1/N,\forall k}|f(x)-p_{s,{\bf
      j}/N,f}(x)|\\
      &\leq&
      2^d\tilde{c}_1\max_{\mathbf j: |x^{(k)}-\mathbf j_k/N|<1/N,\forall
      k}\|x-\mathbf j/N\|_2^r\\
      &\leq&
      \tilde{c}_12^d\sqrt{d^r}N^{-r}.
\end{eqnarray*}
This proves Lemma \ref{Lemma: app by product} with
$\tilde{c}_2=\tilde{c}_12^d\sqrt{d^r}$.
\end{IEEEproof}

 With these helps, we   prove Theorem
\ref{Theorem:jackson} as follows.

\begin{IEEEproof}[Proof of Theorem \ref{Theorem:jackson}]
Rewrite
\begin{equation}\label{f1-rewrite}
   f_1=\sum_{{\bf j}\in\{0,1,\dots,N\}^d}\sum_{\alpha:|\alpha|\leq s}a_{{\bf j},\alpha}\phi_{{\bf j},N}(x)x^\alpha
\end{equation}
with
$
    |a_{{\bf j},\alpha}|\leq  \tilde{B}.
$
  Then for arbitrarily
 fixed $\alpha$ with $|\alpha|\leq s$ and ${\bf j}\in \{0,1,\dots,N\}^d$, we can rewrite
\begin{eqnarray*}
   &&\phi_{{\bf j},N}(x)x^\alpha\\
   &=&
   \overbrace{x^{(1)}\cdots x^{(1)}}^{\alpha_1}\cdots\overbrace{x^{(d)}\cdots x^{(d)}}^{\alpha_d}\cdot \overbrace{1\cdots1}^{s-|\alpha|}\prod_{k=1}^d\psi_k(x) ,
\end{eqnarray*}
where
\begin{equation}\label{def-psi-k}
    \psi_k(x)=
    \psi\left(3N\left(x^{(k)}-\frac{j_k}N\right)\right).
\end{equation}
But Lemma \ref{lemma:product gate 2} shows that for any $0<\nu<1$ and $u_1,\dots,u_{s+d}\in[-1,1]$, there exists a deep net $\tilde{\times}_{d+s}$  with $2(d+s)\tilde{L}+8(d+s)$
layers and at most $c(d+s)^{\theta} \nu^{-\theta}$ free
  parameters which are bounded by $(d+s)^\gamma\nu^{-\gamma}$
 such that
\begin{equation}\label{theore.p.pgu}
       |u_1u_2\cdots u_{d+s}-\tilde{\times}_{d+s}(u_1,\dots,u_{d+s})|\leq
       \nu.
\end{equation}
Define
\begin{eqnarray}\label{deep-net-for-smooth}
   h_f(x)&:=&
   \sum_{{\bf j}\in\{0,1,\dots,N\}^d}\sum_{\alpha:|\alpha|\leq s}a_{{\bf j},\alpha}
   \tilde{\times}_{d+s}\big(\psi_1(x),\dots,\psi_d(x),\nonumber\\
   &&
   \overbrace{x^{(1)},\cdots x^{(1)}}^{\alpha_1},\dots,\overbrace{x^{(d)},\dots x^{(d)}}^{\alpha_d}, \overbrace{1,\dots,1}^{s-|\alpha|}\big).
\end{eqnarray}
Note  that $h_f$ possesses $(N+1)^d\left(^{s+d}_{\ s}\right)$  PGU($d+s$), $\tilde{\times}_{d+s}$, with the same parameters and   also involves $(N+1)^d\left(^{s+d}_{\ s}\right)$ $\Psi_k(\cdot)$. But (\ref{def.psi}) and (\ref{def-psi-k}) show that there are totally $8d+4$  free parameters for  each $\Psi_k(\cdot), k=1,\dots,d$. Therefore, there are totally
$2(s+d)\tilde{L}+8(s+d)+3$
layers and at most $c(d+s)^{\theta} \nu^{-\theta}+(8d+5)(N+1)^d\left(^{s+d}_{\ s}\right)$ free
  parameters which are bounded by $\max\{\tilde{B},3N,(d+s)^\gamma\nu^{-\gamma}\}$ in $h_f$.
Furthermore, (\ref{deep-net-for-smooth}), (\ref{theore.p.pgu}) and (\ref{f1-rewrite}) yield that for arbitrary $x\in\mathbb I^d$,
\begin{eqnarray*}
  |f_1(x)-h_f(x)|
  &\leq&
   \sum_{{\bf j}\in\{0,1,\dots,N\}^d}\sum_{\alpha:|\alpha|\leq s}|a_{{\bf j},\alpha}|\varepsilon\\
   &\leq&
   (N+1)^d\left(^{s+d}_{\ s}\right)\tilde{B}\nu.
\end{eqnarray*}
Recalling Lemma \ref{Lemma: app by product}, we have
$$
    |f(x)-h_f(x)|\leq
   \tilde{c}_2N^{-r}+(N+1)^d\left(^{s+d}_{\ s}\right)\tilde{B}\nu.
$$
 Setting $N+1=\left[\nu^{-1/(d+r)}\right]$, we obtain
$$
    |f(x)-h_f(x)|\leq
   \tilde{c}_3\nu^{r/(r+d)},
$$
where $\tilde{c}_3:=\tilde{c}_2+2^d\left(^{s+d}_{\ s}\right)\tilde{B}.$
Denote $\varepsilon=\nu^{r/(r+d)}$. Then, for any $\varepsilon\in(0,1)$,   $h_f$ is a deep net with $2(d+s)\tilde{L}+8(d+s)+3$
layers and at most $c(d+s)^{\theta} \varepsilon^{-(r+d)\theta/r}+(8d+5)\left(^{s+d}_{\ s}\right)\varepsilon^{-d/r}$ free
parameters which are bounded by $\max\{\tilde{B},3\varepsilon^{-1/r},(d+s)^\gamma\varepsilon^{-(r+d)\gamma/r}\}$ satisfying
$$
    |f(x)-h_f(x)|\leq
   \tilde{c}_3\varepsilon,
$$
This completes the proof of Theorem
\ref{Theorem:jackson}.
\end{IEEEproof}

\section*{Appendix C: Proof of Theorem \ref{Theorem:jackson for
trade-off}}

\begin{IEEEproof}[Proof of Theorem \ref{Theorem:jackson for
trade-off}]
  For any $0<\nu_1<1/2$, $\theta>0$,  $\tilde{L}\in\mathbb N$ with
$\tilde{L}>(2\theta)^{-1}$ and $k=1,\dots,d^*$, it follows from Theorem \ref{Theorem:polynomial} that
 there is  deep ReLU net
$h_{P_{k,\jmath}}$  with   at most  $2\jmath \tilde{L}+8\jmath+1$ layers and
 at most
$\mu+c(\mu\jmath)^{\theta}
\nu_1^{-\theta}$ nonzero parameters which are  bounded by
$ (\mu\jmath)^\gamma\nu_1^{-\gamma}$
 such that
\begin{equation}\label{add1}
       |P_{k,\jmath}(x)-h_{P_{k,\jmath}}(x)|\leq \nu_1,
       \qquad\forall x\in \mathbb I^{D_k}, k=1,\dots,d^*.
\end{equation}
Since $|P_{k,\jmath}(x)|\leq 1/2$ for $x\in\mathbb I^d$, we have from (\ref{add1}) and $0<\nu_1\leq 1/2$ that $|h_{P_{k,\jmath}}(x)|\leq 1$.
  Define
$$
      h_{d^*}(x)=
      (h_{P_{1,\jmath}}(x),h_{P_{2,\jmath}}(x),\dots,h_{P_{d^*,\jmath}}(x)).
$$
Let $g$ be the $d^*$-dimensional function in Assumption \ref{Assumption:Features}, i.e.,
$$
     f^*(x)= g(P_{1,\jmath}(x),\dots,P_{d^*,\jmath}(x)).
$$
For arbitrary $0<\nu_2<1$, Theorem \ref{Theorem:jackson} shows that there is deep net  $h_g$ with
 $
     \mathcal L(d^*,r,\tilde{L}) $
layers and at most $c(d^*+s)^{\theta} \nu_2^{-(r+d^*)\theta/r}+(8d^*+5)\left(^{s+d^*}_{\ s}\right)\nu_2^{-d^*/r}$  free
parameters which are bounded by $\max\{\tilde{B}_g,3\nu_2^{-1/r},(d^*+s)^\gamma\nu_2^{-(r+d^*)\gamma/r}\}$
  such that
\begin{equation}\label{Jackson1}
       \|g-h_g\|_{L^\infty(\mathbb I^{d^*})}\leq c_1\nu_2,
\end{equation}
where
$$
   \tilde{B}_g:=\max_{k_1+\dots+k_{d^*}\leq \jmath}\max_{x\in\mathbb I^d}\left|\frac{1}{k_1!\dots
    k_{d^*}!}\frac{\partial^{k_1+\dots
    k_{d^*}}f(x)}{\partial^{k_1}x^{(1)}\dots\partial^{k_d}x^{({d^*})}}\right|
$$
  Define
\begin{equation}\label{a.222}
      h_{f^*}(x)= h_g(h_{d^*}(x)).
\end{equation}
For arbitrary $x\in\mathbb I^d$, there holds
\begin{eqnarray*}
       &&|f^*(x)-h_{f^*}(x)|\\
       &=&
       |g(P_{1,\jmath}(x),\dots,P_{d^*,\jmath}(x))
       -h_g(h_{d^*}(x))|\\
       &\leq&
       |g(P_{1,\jmath}(x),\dots,P_{d^*,\jmath}(x))-g(h_{d^*}(x))|\\
       &+&
       | g(h_{d^*}(x))
       -h_g(h_{d^*}(x))|.
\end{eqnarray*}
Due to the smoothness of $g$, we have from (\ref{add1}) and (\ref{lip}) that
\begin{eqnarray*}
   &&|g(P_{1,\jmath}(x),\dots,P_{d^*,\jmath}(x))-g(h_{d^*}(x))|\\
       &\leq&
       \tilde{c}_4 \max_{1\leq k\leq d^*}
       |P_{k,\jmath}(x)-h_{P_{k,\jmath}}(x)|^{\tau_r}
        \leq
        \tilde{c}_4\nu_1^{\tau_r},
\end{eqnarray*}
where $\tilde{c}_4>0$ is a constant depending only on $c_0$, $d^*$ and $g$.
But (\ref{Jackson1})  yields
\begin{eqnarray*}
    | g(h_{d^*}(x))
       -h_g(h_{d^*}(x))|
       \leq
       c_1\nu_2, \qquad \forall\ x\in\mathbb I^d.
\end{eqnarray*}
Thus, we have
$$
     |f^*(x)-h_{f^*}(x)|\leq   c_1\nu_2+\tilde{c}_4\nu_1^{\tau_r}.
$$
Set $\nu_2=\nu_1^{\tau_r}= \varepsilon$. Then for any $0<\varepsilon\leq 1/2$,   we get
$$
        |f^*(x)-h_{f^*}(x)|\leq \tilde{c}_5\varepsilon,
$$
where $\tilde{c}_5$ is a constant depending only on $c_0,d,d^*,r,s$ and $g$.
Under this circumstance, $h_{f^*}$ is a deep net with
$$
        \mathcal L^*(d^*,r,\tilde{L},\jmath)=\mathcal
       L(d^*,r,\tilde{L})+ 2\jmath \tilde{L}+8\jmath+1
$$
layers
 and
 at most
 $$
   c(d^*+s)^{\theta} \varepsilon^{-\frac{(r+d^*)\theta}{r}}+(8d^*+5)\left(^{s+d^*}_{\ s}\right)\varepsilon^{-\frac{d^*}{r}} +d^*\mu+c(\mu\jmath)^{\theta}
    \varepsilon^{-\frac{\theta}{\tau_r}}
$$
free parameters which are bounded by
$$
    \max\{\tilde{B}_g,3\varepsilon^{-1/r},(d^*+s)^\gamma\varepsilon^{-(r+d^*)\gamma/r},(\mu\jmath)^\gamma\varepsilon^{-\gamma/\tau_r}\}.
$$
This completes the proof of Theorem \ref{Theorem:jackson for
trade-off}.
\end{IEEEproof}

\section*{Appendix D: Proof of Theorem \ref{Theorem: ERM}}

In this section only, $C_1',C_2',\dots$ denote constants independent of $L$, $n$, $\varepsilon$, $\delta$, $\mu$ or $\jmath$.
To prove Theorem \ref{Theorem: ERM}, we need the following
well-known oracle inequality in learning theory, the proof of which
can be found in \cite{Chui2018a}.

\begin{lemma}\label{Lemma:oracle}
Let $\rho_X$ be the marginal distribution of  $\rho$ on $\mathcal X$
and $(L^2_{\rho_{_X}}, \|\cdot\|_\rho)$ denote the Hilbert space of
square-integrable functions on $\mathcal X$ with respect to
$\rho_X$.  Set $\mathcal
E_{D}(f):=\frac1m\sum_{i=1}^m(f(x_i)-y_i)^2$ and define
\begin{equation}\label{ERM!!!!!!}
        f_{D,\mathcal H}=\arg\min_{f\in\mathcal H}
        \mathcal E_{D}(f),
\end{equation}
where $\mathcal H$ is a
set of functions defined on $\mathcal X$.
Suppose  further that there exist  $n', \mathcal U>0$, such that
\begin{equation}\label{oracle condition}
    \log \mathcal N(\varepsilon,\mathcal H,L^\infty(\mathbb I^d))\leq
     n'\log \frac{\mathcal U}{\varepsilon},\qquad\forall  \varepsilon>0.
\end{equation}
Then for any $h\in\mathcal H$,
\begin{eqnarray*}
      &&  Prob\{\|\pi_Mf_{ D,\mathcal
      H}-f_\rho\|_\rho^2>\varepsilon+2\|h-f_\rho\|_\rho^2\}\\
      &\leq&
      \exp\left\{n'\log\frac{16\mathcal UM}{\varepsilon}-\frac{3m\varepsilon}{512M^2}\right\}\\
      &+&
      \exp\left\{\frac{-3m\varepsilon^2}{16(3M+\|h\|_{L_\infty(\mathcal X)})^2\left(
    6\|h-f_\rho\|_\rho^2+
    \varepsilon\right)}\right\}.
\end{eqnarray*}
\end{lemma}

 Now we apply
Lemma \ref{Lemma:oracle}, Lemma \ref{Lemma:covering number}, and
Theorem \ref{Theorem:jackson for trade-off} to prove   Theorem \ref{Theorem: ERM}.

\begin{table*}[htb]
\begin{center}
\caption{Network architecture and corresponding MSE.}\label{distributions}
\begin{tabular}{cccccccc}
\hline
%\multicolumn{3}{c}{Number of nodes}&\multicolumn{4}{c}{\multirow{2}{*}{Parameter percentage of each layer}}&\multirow{2}{1cm}{MSE}\\
\multicolumn{3}{c}{Widths}&\multicolumn{4}{c}{Parameter percentage (\%)}&\multirow{2}{1cm}{MSE}\\
  {Layer-1}&{Layer-2}&{Layer-3}&$p(C_1)$&$p(C_2)$&$p(C_3)$&$p(C_4)$\\
  \hline
  $150$&$30$&$60$&$18.62$&$57.34$&$22.56$&$1.48$&$4410.946$\\
  $50$&$90$&$30$&$6.45$&$57.52$&$35.27$&$0.76$&$4588.056$\\
  $60$&$60$&$60$&$5.72$&$39.74$&$53.11$&$1.43$&$4690.85$\\
  $120$&$20$&$200$&$14.85$&$30.92$&$49.33$&$4.91$&$5402.052$\\
  $220$&$20$&$50$&$27.80$&$58.11$&$12.83$&$1.26$&$5936.061$\\
  $80$&$50$&$60$&$10.07$&$50.75$&$37.94$&$1.24$&$6147.745$\\
  $80$&$70$&$20$&$10.13$&$71$&$18.38$&$0.49$&$6573.265$\\
  $50$&$110$&$20$&$6.06$&$65.99$&$27.47$&$0.48$&$7120.719$\\
  $100$&$40$&$70$&$13.45$&$50.12$&$35.41$&$1.02$&$7551.107$\\
  $210$&$20$&$70$&$26.11$&$54.58$&$17.57$&$1.73$&$8574.581$\\
  $210$&$10$&$300$&$26.24$&$28.87$&$37.44$&$7.45$&$11103.629$\\
  $40$&$10$&$600$&$5.1$&$5.47$&$74.75$&$14.68$&$16730.241$\\
  $350$&$10$&$60$&$43.39$&$47.59$&$7.54$&$1.48$&$31249.375$\\
  $50$&$20$&$300$&$6.34$&$12.01$&$74.88$&$6.77$&$35842.366$\\
  $300$&$5$&$500$&$36.2$&$21.65$&$30.13$&$12.03$&$52774.265$\\
  $420$&$1$&$1000$&$26.24$&$10.45$&$12.45$&$50.86$&$203840.239$\\
  $380$&$10$&$10$&$44.92$&$51.48$&$1.35$&$2.25$&$217662.506$\\
  $1$&$15$&$500$&$0.25$&$0.19$&$87.14$&$12.42$&$413626478.3$\\
  $70$&$1$&$2400$&$26.24$&$1.74$&$29.86$&$33.16$&$434036281.1$\\
  $1$&$1$&$2670$&$0.25$&$0.02$&$33.25$&$66.48$&$766736147.9$\\
  \hline
\end{tabular}
\end{center}
\end{table*}

\begin{table*}[hbt]
\begin{center}
\caption{Extracting partially radial feature with different groups.}
\label{adaptivity}
\begin{tabular}{c|c|c|c|c|c|c|c|c|c}
\hline
\multicolumn{2}{c|}{Evaluation metrics}&\multicolumn{1}{c|}{k=2}&\multicolumn{1}{c|}{k=3}&\multicolumn{1}{c|}{k=4}&\multicolumn{1}{c|}{k=5}&\multicolumn{1}{c|}{k=6}&\multicolumn{1}{c|}{k=7}&\multicolumn{1}{c|}{k=8}&\multicolumn{1}{c}{k=9}\\

  \hline
  \multirow{2}{*} {MAE}
  &{1-layer}&$24.616$&$26.027$&$30.948$&$30.957$&$38.471$&$47.654$&$48.89$&$66.579$\\
  &{3-layer}&$15.775$&$23.827$&$28.888$&$29.002$&$35.628$&$46.716$&$47.914$&$56.534$\\
  \hline
  \multirow{2}{*} {MSE}
  &{1-layer}&$1357.934$&$1885.427$&$2345.819$&$2851.989$&$4216.136$&$5231.759$&$6304.671$&$9154.373$\\
  &{3-layer}&$777.817$&$1682.958$&$2125.355$&$2555.937$&$3674.015$&$4884.519$&$6265.782$&$7098.902$\\
  \hline
  \multirow{2}{*} {MdAE}
  &{1-layer}&$14.897$&$16.511$&$18.476$&$24.209$&$25.818$&$30.385$&$30.42$&$35.619$\\
  &{3-layer}&$8.488$&$13.609$&$17.149$&$20.01$&$24.123$&$28.585$&$29.569$&$34.299$\\
  \hline
  \multirow{2}{*} {R$^2$S}
  &{1-layer}&$1.0$&$1.0$&$1.0$&$1.0$&$1.0$&$1.0$&$1.0$&$1.0$\\
  &{3-layer}&$1.0$&$1.0$&$1.0$&$1.0$&$1.0$&$1.0$&$1.0$&$1.0$\\
  \hline
  \multirow{2}{*} {EVS}
  &{1-layer}&$1.0$&$1.0$&$1.0$&$1.0$&$1.0$&$1.0$&$1.0$&$1.0$\\
  &{3-layer}&$1.0$&$1.0$&$1.0$&$1.0$&$1.0$&$1.0$&$1.0$&$1.0$\\
  \hline
\end{tabular}
\end{center}
\end{table*}

\begin{table*}[htb]
\begin{center}
\caption{Extracting the square-feature of different network depths.}\label{depth}
\begin{tabular}{c c c c c c c}
\hline
  {Depth}&{1-layer}&{3-layer}&{5-layer}&{7-layer}&{9-layer}&{11-layer}\\
  \hline
  {MAE}&$48.989$&$44.006$&$145.568$&$176.055$&$427.975$&$1204.083$\\
  {MSE}&$14837.713$&$4999.949$&$49579.938$&$86620.738$&$273505.566$&$3053474.62$\\
  {MdAE}&$15.46$&$8.624$&$87.641$&$92.42$&$404.614$&$836.368$\\
  {R$^2$S}&$1.0$&$1.0$&$1.0$&$1.0$&$1.0$&$1.0$\\
  {EVS}&$1.0$&$1.0$&$1.0$&$1.0$&$1.0$&$1.0$\\
  \hline
\end{tabular}
\end{center}
\end{table*}

\begin{IEEEproof}[Proof of Theorem \ref{Theorem: ERM}]
Due to (\ref{restriction on theta}) and (\ref{condition1th}), it follows from Theorem \ref{Theorem:jackson for trade-off} with $\varepsilon=n^{-r/d^*}$ that there exists an
 $h_\rho\in \mathcal H_{L,n,\mathcal R}$
  with  $L=\mathcal L^*(d^*,r,\tilde{L},\jmath)$, $\mathcal R$ given in (\ref{R}), and at most
 $C_1'n$ free parameters such that
$$
      \|f_\rho-h_{\rho}\|^2_\rho\leq \|f_\rho-h_{\rho}\|^2_{L^\infty(\mathbb I^d)} \leq c_2^2n^{-2r/d^*}.
$$
Noting further  $|y|\leq M$ almost surely as well as  $\|f_\rho\|_{L_\infty(\mathbb I^d)}\leq M$, we get
$$
       \|h_{\rho}\|_{L^\infty(\mathbb I^d)} \leq c_2+M=:C_2'.
$$
But Lemma \ref{Lemma:covering number}  shows
\begin{eqnarray*}
   \log\mathcal N\left( \varepsilon,\mathcal H_{n,L,\mathcal
       R},L^\infty(\mathbb I^d)\right)
       \leq
       C_3'L^2n\log\frac{n}{\varepsilon},
\end{eqnarray*}
where we used (\ref{condition1th}), (\ref{deep-net-for-smooth}), (\ref{a.111}), (\ref{a.222}), Figure \ref{fig:PGU} and  the fact that the largest width of deep nets in   PGU(2) in get an $\varepsilon$ approximation of the product of two real number \cite[Lemma A.3]{Petersen2017} is  smaller than $\tilde{C}_{pgu2} \varepsilon^{-\theta}$
to derive
\begin{eqnarray*}
      D_{\max}
      &\leq& \max\left\{\left(^{\jmath+d}_{\ d}\right)\jmath  c \tilde{C}_{pgu2} \jmath^\theta n^{\theta r/d^*},\right.\\
      &&\left. C_1'n\left(^{s+d^*}_{\ d^*}\right)(s+4d^*+ c \tilde{C}_{pgu2} s^\theta n^{\theta r/d^*} )\right\}\\
      &\leq&
      C_4' n^{C_4'},
\end{eqnarray*}
and $\tilde{C}_{pgu2}$ is an absolute constant.
Plugging the above three estimates into Lemma \ref{Lemma:oracle} with $n'=C_5'L^2n$, $\mathcal U=n$, we have
\begin{eqnarray*}
      &&  Prob\{\|\pi_Mf_{ D,n,L}-f_\rho\|_\rho^2>\varepsilon+2\|h_\rho-f_\rho\|_\rho^2\}\\
      &\leq&
      \exp\left\{C_5'L^2n\log\frac{16Mn}{\varepsilon}-\frac{3m\varepsilon}{512M^2}\right\}\\
      &+&
      \exp\left\{\frac{-3m\varepsilon^2}{16(3M+C_2')^2\left(
    6 c_5^2n^{-2r/d^*}+
    \varepsilon\right)}\right\}.
\end{eqnarray*}
 For
 \begin{equation}\label{coditionto-varep}
 \varepsilon\geq C_3'n^{-2r/d^*}L^2\log n,
  \end{equation}
  we then get
 \begin{eqnarray*}
      &&  Prob\{\|\pi_Mf_{ D,n,L}-f_\rho\|_\rho^2>3\varepsilon\}\\
      &\leq&
      \exp\left\{C_6'(nL^2\log n-m\varepsilon)\right\}
      +\exp\{-C_7'm\varepsilon\}.
 \end{eqnarray*}
Since
 $
    n=\left[C_{1}m^{\frac{d^*}{2r+d^*}}\right],
$
we obtain
\begin{eqnarray*}
     &&Prob\{\|\pi_Mf_{ D,n,L}-f_\rho\|_\rho^2>3\varepsilon\}
     \leq
     2\exp\{-C_{8}'m\varepsilon\}\\
     &\leq&
     2\exp\{-C_{9}'m^\frac{2r}{2r+d^*}\varepsilon/\log n\}.
\end{eqnarray*}
 Let
$$
   \varepsilon= C_{10}'m^{-\frac{2r}{2r+d^*}}L^2\log n\log\frac3\delta
$$
with $C_{10}'C_9'>1$   such that (\ref{coditionto-varep}) holds. Then,
$$
      Prob\{\|\pi_Mf_{ D,n,L}-f_\rho\|_\rho^2>3\varepsilon\}
     \leq \delta.
$$
  Thus,   with confidence of at least
$1-\delta$, there holds
\begin{eqnarray*}
       \|\pi_Mf_{
     D,n,L}-f_\rho\|_\rho^2
     \leq
      C_{11}'L^2m^{-\frac{2r}{2r+1}}\log m \log\frac3\delta.
\end{eqnarray*}
This proves (\ref{learning rate})  by noting the well-known relation
\begin{equation}\label{equality}
        \mathcal E(f)-\mathcal E(f_\rho)=\|f-f_\rho\|_\rho^2
\end{equation}
%The upper bound of (\ref{almost optimal learning rate}) can be derived from
%(\ref{learning rate}) and the standard expectation formula
%$$
%      E[\xi]=\int_0^\infty{ Prob}[\xi>t]dt
%$$
%directly. To prove the lower bound of (\ref{almost optimal learning
%rate}), we note that since $x_1,\dots, x_m$ are
%independent random variables,
%so are $(P_{1,\jmath}(x_i), \dots, P_{d^*,\jmath}(x_i))_{i=1}^m$ whose coefficients are 0 for after $\mu$ monomials for a fixed arrangement of monomials in the polynomials.    Thus, the data set $((P_{1,\jmath}(x_i), \dots, P_{d^*,\jmath}(x_i)),y_i)_{i=1}^m$ are
%independently drawn according to some distribution $\rho'$ defined
%on $\mathbb I^{d^*}\times [-M,M]$.  From \cite[Theorem 3.2]{Gyorfi2002},
%there exists some $\rho_0'$ with the regression function $g_\rho\in
%Lip^{(r,c_0)}$ defined on $\mathbb I^{d^*}$, such that the learning rates of all
%estimates based on $m$ data are not smaller than
%$C_{12}'m^{-\frac{2r}{2r+d^*}}$.  Setting $f_\rho(x) =g_\rho(P_{1,\jmath}(x), \dots, P_{d^*,\jmath}(x))$, we   conclude that the lower bound of (\ref{almost
%optimal learning rate}) holds.
and completes the proof of Theorem
\ref{Theorem: ERM}.
\end{IEEEproof}

\section*{Appendix E: Detailed Numerical Results of Experiment \ref{Sec.Role-of-width}}

The detailed numerical results are shown in Table \ref{distributions}.

\section*{Appendix F: Detailed Numerical Results of Experiment \ref{Sec.Adaptivity-to-depth}}

The detailed numerical results are shown in Table \ref{adaptivity}.

\section*{Appendix G: Detailed Numerical Results of Experiment \ref{Sec.Role-of-depth}}

The detailed numerical results are shown in Table \ref{depth}.


\begin{thebibliography}{99}


\bibitem{Allen-Zhu2018}
Z. Allen-Zhu, Y. Li, and Z. Song. A convergence theory for deep learning via over-parameterization. arXiv preprint arXiv:1811.03962, 2018.

%
%\bibitem{Adeli2009}
% H. Adeli, A. Panakkat, A probabilistic neural network for earthquake
%magnitude prediction, Neural Networks,  22 (2009), 1018-1024.
%
\bibitem{Barron1993}
A. Barron. Universal approximation bounds for superpositions of a sigmoidal function. IEEE Trans. Inform. Theory, 1993, 39(3): 930-945.

\bibitem{Bengio2013}
Y. Bengio, A. Courville, and P. Vincent. Representation learning: A
review and new perspectives. IEEE Trans. Pattern Anal. Mach. Intel.,
3, 1798-1828, 2013.

\bibitem{Bianchini2014}
M. Bianchini and F. Scarselli.  On the complexity of neural network
classifiers: a comparison between shallow and deep architectures.
IEEE. Trans. Neural Netw. \& Learn. Sys.,  25: 1553-1565,
2014.

\bibitem{Bishop}
C. M. Bishop. Pattern Recognition and Machine Learning. Springer,
2006.


\bibitem{Bruna2013}
J. Bruna and S. Mallat. Invariant scattering convolution networks. IEEE Trans.   Pattern Anal. Mach. Intel.,  35: 1872-1886, 2013.

 \bibitem{Chui1994}
C. K. Chui, X. Li, and H. N. Mhaskar. Neural networks for lozalized
approximation. Math. Comput., 63: 607-623, 1994.



\bibitem{Chui2018}
C. K. Chui, S. B. Lin, and D. X. Zhou. Construction of neural networks
for realization of localized deep
  learning.  Front. Appl. Math. Statis., 4: 14, 2018.

\bibitem{Chui2018a}
C. K. Chui, S. B. Lin, and D. X. Zhou, Deep neural networks for rotation-invariance approximation and learning. Anal. Appl., 17: 737-772, 2019.

\bibitem{Cucker2007}
F. Cucker and D. X. Zhou. Learning Theory: An Approximation Theory
Viewpoint,  Cambridge University Press, Cambridge, 2007.

\bibitem{Cybenko1989}
G. Cybenko. Approximation by superpositions of sigmoidal function.
Math. Control Signals Syst., 2: 303-314, 1989.


\bibitem{Delalleau2011}
O. Delalleau and Y. Bengio.  Shallow vs. deep sum-product
networks.  NIPs, 666-674, 2011.

\bibitem{Donoho1993}
D. L. Donoho. Unconditional bases are optimal bases for data compression and for statistical estimation.
Appl. Comput. Harmonic. Anal., 1(1):100-115, 1993.


\bibitem{Eldan2015}
R. Eldan and O. Shamir. The power of depth for feedforward neural
networks. arXiv preprint arXiv:1512.03965, 2015.


\bibitem{Evgeniou2000}
T. Evgeniou, M. Pontil,  and  T. Poggio.  Regularization networks and
support vector machines.  Adv. Comput. Math.,   13: 1-50, 2000.





\bibitem{Goodfellow2016}
I. Goodfellow, Y. Bengio, and A. Courville. {Deep Learning}.  MIT Press,
2016.

\bibitem{Guo2017}
Z. C. Guo, D. H. Xiang, X. Guo, and D. X. Zhou. Thresholded spectral algorithms for sparse approximations. Anal. Appl. 15: 433--455, 2017.

\bibitem{Guo2019}
Z. C. Guo, L. Shi, and S. B. Lin. Realizing data features by deep nets, IEEE Trans. Neural Netw. Learn. Syst., DOI: 10.1109/TNNLS.2019.2951788.

\bibitem{Gyorfi2002}
L. Gy\"{o}rfy, M. Kohler, A. Krzyzak and H. Walk. A
Distribution-Free Theory of Nonparametric Regression. Springer,
Berlin, 2002.
%
%
%\bibitem{Guliyev2018}
%N. Guliyev, V. Ismailov . Approximation capability of two hidden
%layer feedforward neural networks with fixed weights.
%Neurocomputing,   316: 262-269, 2018.
%
%
%
\bibitem{Hagan1996}
M. Hagan, M. Beale, and H. Demuth.  Neural Network Design. PWS
Publishing Company, Boston, 1996.
%
\bibitem{Hastie2001}
T. Hastie, R. Tibshirani, and J. Friedman. The Elements of
Statistical Learning: Data mining, Inference and Prediction.
Springer, New York, 2001.
%
%
%
%\bibitem{Hanin2017}
%B. Hanin. Universal function approximation by deep neural nets with
%bounded width and relu activations. arXiv preprint arXiv:1708.02691,
%2017.
%
\bibitem{Harvey2017}
N. Harvey, C. Liaw, A. Mehrabian. Nearly-tight VC-dimension bounds
for piecewise linear neural networks. Conference on Learning Theory.
2017: 1064-1068.


\bibitem{Hinton2006}
G. E. Hinton, S. Osindero, and Y. W. Teh. A fast learning algorithm for
deep belief netws. Neural Comput., 18:, 1527-1554, 2006.

%\bibitem{Hinton2012}
%G. Hinton, L. Deng, D. Yu, G. E. Dahl, A. R. Mohamed, N. Jaitly, A.
%Senior, V. Vanhoucke, P. Nguyen, T. N. Sainath, B. Kingsbury, Deep
%neural networks for acoustic modeling in speech recognition: The
%shared views of four research groups. IEEE Signal Process. Mag., 29
%(2012), 82-97.

\bibitem{Imaizumi2018}
M. Imaizumi and K. Fukumizu. Deep Neural Networks Learn Non-Smooth
Functions Effectively. arXiv preprint arXiv:1802.04474, 2018.

%\bibitem{Ismailov2014}
%V. E. Ismailov, On the approximation by neural networks with bounded
%number of neurons in hidden layers, J. Math. Anal. Appl., 417
%(2014), 963-969.
%
\bibitem{Kohler2014}
M. Kohler. Optimal global rates of convergence for noiseless
regression estimation problems with adaptively chosen design. J.
Mult. Anal., 132: 197-208, 2014.

\bibitem{Kohler2017}
M. Kohler and A. Krzyzak. Nonparametric regression based on
hierarchical interaction models. IEEE Trans. Inform. Theory, 63:
1620-1630, 2017.

%
%\bibitem{Konovalov2008}
%V. N. Konovalov, D. Leviatan, V. E. Maiorov, Approximation by
%polynomials and ridge functions of classes of s-monotone radial
%functions, J. Approx. Theory,  152 (2008), 20-51.
%
%\bibitem{Konovalov2009}
%V. N. Konovalov, D.  Leviatan, V. E. Maiorov, Approximation of
%Sobolev classes by polynomials and ridge functions, J. Approx.
%Theory,   159 (2009), 97-108.
%
%
%
%\bibitem{Kurkova2017}
%V. K\r{u}rkov\'{a}, M. Sanguineti, Probabilistic lower bounds for
%approximation by shallow perceptron networks, Neural Networks, 91
%(2017), 34-41.
%
%
%\bibitem{Krizhevsky2012}
%A. Krizhevsky, I. Sutskever, G. E. Hinton, Imagenet classification
%with deep convolutional neural networks, NIPS, 2097-1105, 2012.

%
%
%\bibitem{Lecun2015}
%Y. LeCun, Y. Bengio, G. Hinton, Deep learning,  Nature,  521(7553)
%(2015), 436-444.

\bibitem{Leshno1993}
M. Leshno, V. Y. Lin, A. Pinks, and S. Schocken. Multilayer feedforward
networks with a nonpolynomial activation function can approximat any
function. Neural Networks, 6: 861-867, 1993.


\bibitem{LinH2017}
H. W. Lin, M. Tegmark and D. Rolnick.  Why does deep and cheap
learning works so well?. J. Stat. Phys.,  168: 1223-1247,
2017.


\bibitem{Lin2014b}
S.  Lin,  Y.  Rong and  Z.  Xu.  Multivariate Jackson-type
inequality for a new type neural network approximation. Appl.
Math. Model.,   38: 6031-6037, 2014.

%
%\bibitem{Lin2017}
%S. B. Lin, J.Zeng, and X. Chang. Learning rates for classification with
%Gaussian kernels. Neural Comput.,   29(12): 3353-3380, 2017.

\bibitem{Lin2017a}
S. B. Lin. Limitations of shallow nets approximation. Neural
Networks,   94: 96-102, 2017.


\bibitem{Lin2018}
S. B. Lin. Generalization and expressivity for deep nets. IEEE
Trans. Neural Netw. Learn. Syst., In Press.

\bibitem{Lin2018CA}
S. B. Lin and D. X. Zhou. Distributed kernel-based gradient descent
algorithms. Constr. Approx., 47: 249-276, 2018.

\bibitem{Lin2019}
S. Lin, J. Zeng, and X. Zhang. Constructive neural network learning. IEEE Trans. Cyber., 49: 221 - 232, 2019.

\bibitem{LinT2008}
T. Lin and H. Zha. Riemannian manifold learning. IEEE Trans. Pattern
Anal. Mach. Intel.,   30: 796-809, 2008.

% \bibitem{Maiorov1999a}
%V. Maiorov, On best approximation by ridge functions, J. Approx.
%Theory, 99 (1999), 68-94.

\bibitem{Maiorov1999b}
V. Maiorov  and A. Pinkus. Lower bounds for approximation by MLP neural
networks. Neurocomputing,  25: 81-91, 1999.


%
%\bibitem{Maiorov2000}
%V. Maiorov, R. Meir, On the near optimality of the stochastic
%approximation of smooth functions by neural networks, Adv. Comput.
%Math.,   13 (2000), 79-103.
%
%
%
%
%\bibitem{McCane2017}
%B. McCane, L. Szymanski, Deep radial kernel networks: approximating
%radially symmetric functions with deep networks, arXiv preprint
%arXiv:1703.03470, 2017.





\bibitem{Mhaskar1996}
H. N. Mhaskar. Neural networks for optimal approximation of smooth
and analytic functions. Neural Comput., 8: 164-177, 1996.

\bibitem{Mhaskar2016a}
H. N. Mhaskar and T. Poggio. Deep vs. shallow networks: An
approximation theory perspective. Anal. Appl.,  14: 829-848, 2016.


\bibitem{Montufar2013}
G. Mont\'{u}far, R. pascanu, K. Cho, and Y. Bengio.  On the number
of linear regions of deep nerual networks.  Nips, 2014:  2924-2932.



\bibitem{Petersen2017}
P. Petersen and F. Voigtlaender. Optimal aproximation of piecewise
smooth functions using deep ReLU neural networks. Neural Networks,
108: 296-330, 2018.

\bibitem{Pinkus1985}
A. Pinkus. $n$-Widths in Approximation Theory.
  Springer-Velag, Berlin Heidelberg, 1985.

  \bibitem{Pinkus1999}
A. Pinkus. Approximation theory of the MLP model in neural networks.
Acta Numerica, 8: 143-195, 1999.

%\bibitem{Poggio2017}
%T. Poggio, H. Mhaskar, L. Rosasco, B. Miranda, Q. Liao, Why and when
%can deep-but not shallow-networks avoid the curse of dimensionality:
%A review, Intern. J.   Auto. Comput.,  DOI:
%10.1007/s11633-017-1054-2, 2017.



\bibitem{Safran2016}
I. Safran and O. Shamir. Depth-width tradeoffs in approximating natural
functions with neural networks. arXiv reprint arXiv:1610.09887v2,
2016.

\bibitem{Satriano2011}
C. Satriano, Y. M. Wu, A. Zollo, and H. Kanamori. Earthquake early
warning: Concepts, methods and physical grounds. Soil Dynamics
Earth. Engineer., 31: 106-118, 2011.

\bibitem{Schwab2018}
C. Schwab and J. Zech. Deep learning in high dimension: Neural network
expression rates for generalized polynomial chaos expansions in UQ.
Anal. Appl.,  17: 19-55, 2019.

\bibitem{Shaham2015}
U. Shaham,  A. Cloninger, and R. R. Coifman. Provable approximation
properties for deep neural networks. Appl. Comput. Harmonic Anal.,  44: 537-557, 2018.
%
%\bibitem{Shi2011}
%L. Shi, Y. L. Feng and  D. X. Zhou.  ``Concentration estimates for
%learning with $l_1$-regularizer and data dependent hypothesis
%spaces'',  Appl. Comput. Harmon. Anal.,   vol. 31, pp.  286-302,
%2011.

 \bibitem{Sokolov1998On}
V. Y. Sokolov and Y. K. Chernov. On the correlation of seismic intensity with Fourier amplitude spectra. Earthquake Spectra., 14: 679-694, 1998.


\bibitem{sokolov2002seismic}
V. Y. Sokolov. Seismic intensity and Fourier acceleration spectra: revised relationship. Earthquake Spectra., 18: 161-187, 2002.

\bibitem{Steinwart2008}
I. Steinwart, and A. Christmann.
Support Vector Machines.
 Springer, New York, 2008.


\bibitem{Vikraman2016}
K. Vikraman. A deep neural network to identify foreshocks in real
time. arXiv preprint arXiv:1611.08655, 2016.

%\bibitem{Xiang2009}
%D. H. Xiang and D. X. Zhou.
%Classification with Gaussians and convex loss.
% J. Mach. Learn. Res., 10: 1447-1468, 2009.


\bibitem{Yarotsky2017}
D. Yarotsky. Error bounds for aproximations with deep ReLU networks.
Neural Networks, 94: 103-114, 2017.

\bibitem{Ying2017}
Y. Ying and D. X. Zhou. Unregularized online learning algorithms with general loss functions. Appl. Comput.
Harmonic Anal., 42: 224-244, 2017.





\bibitem{Zhou2003}
 D. X. Zhou. Capacity of reproducing kernel spaces in learning
 theory.
IEEE Trans. Inform. Theory,  49: 1743-1752, 2003.

\bibitem{Zhou2006}
{D. X. Zhou and K. Jetter.  Approximation with polynomial kernels and
SVM classifiers. Adv. Comput. Math., 25: 323-344, 2006.}


\bibitem{Zhou2018}
D. X. Zhou. Deep distributed convolutional neural networks:
Universality. Anal. Appl.,  16: 895-919, 2018.

\bibitem{Zhou2018a}
D. X. Zhou. Universality of Deep Convolutional Neural Networks.
  Appl. Comput. Harmonic. Anal., 48: 784-794, 2020
%
%
\bibitem{Zhou2020b}
D. X. Zhou. Theory of deep convolutional neural networks: Downsampling. Neural Netw., 124: 319-327, 2020.

%\bibitem{Trifunac1979Preliminary}
%M. D. TRIFUNAC, Preliminary empirical model for scaling fourier amplitude spectra of strong ground acceleration in terms of modified mercalli intensity and recording site conditions, Earthquake Engineering \& Structural Dynamics., 7(1979), 63-74.
%
%
%\bibitem{Trifunac1989Scaling}
%M. D. TRIFUNAC, Scaling strong motion fourier spectra by modified mercalli intensity , local soil and local geologic site conditions, Doboku Gakkai Ronbunshu., 410 (1989), 217-224.




%\bibitem{Chernov1999Correlation}
%Yu. K. Chernov, V. Yu. Sokolov, Correlation of seismic intensity with Fourier acceleration spectra, Physics \& Chemistry of the Earth Part A Solid Earth \& Geodesy., 24 (1999), 523-528.


%
%
%
%
%
%
%
%

\end{thebibliography}
\end{document}